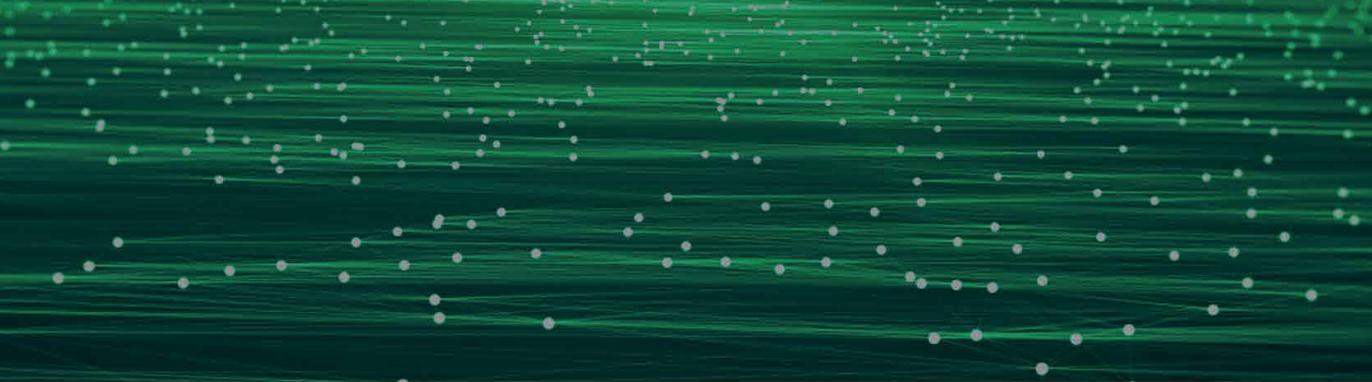

IntechOpen Series
Artificial Intelligence, Volume 7

# Machine Learning
## Algorithms, Models and Applications

*Edited by Jaydip Sen*

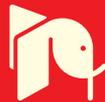

# Machine Learning - Algorithms, Models and Applications

*Edited by Jaydip Sen*

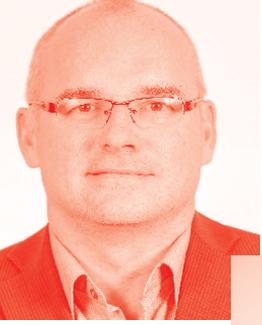
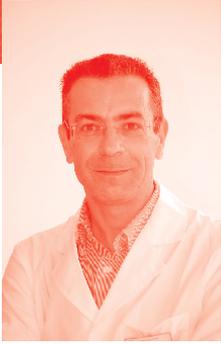
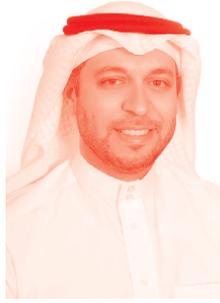
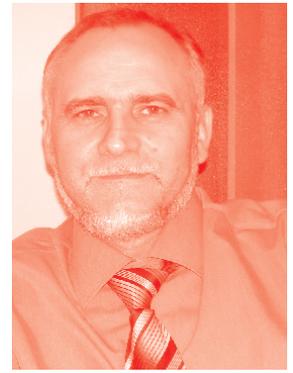
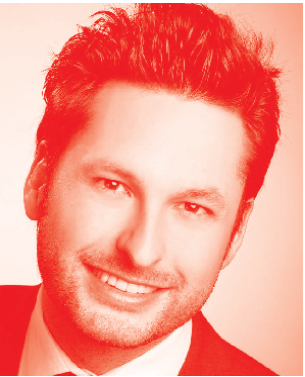
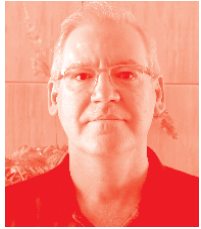
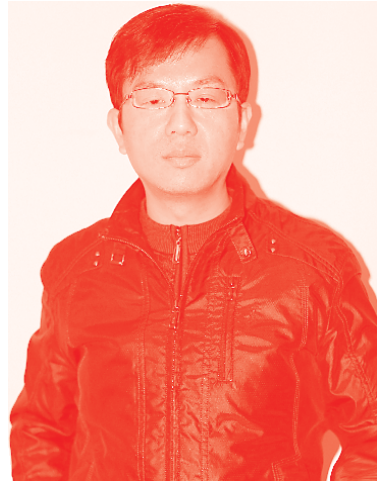

IntechOpen

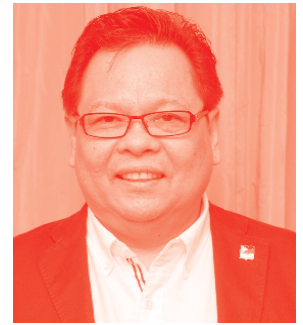
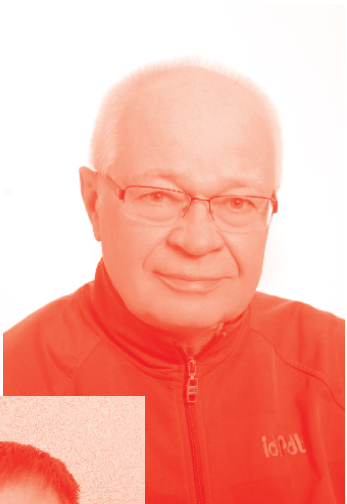
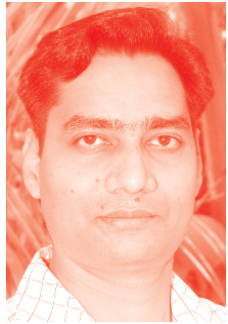
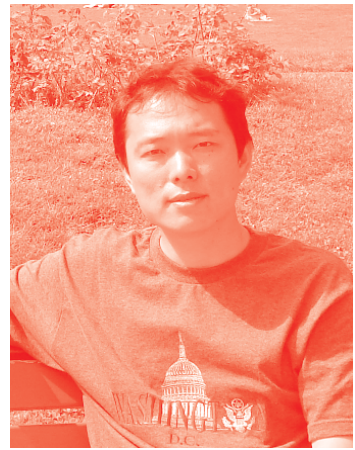
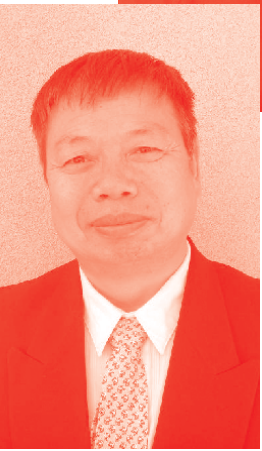
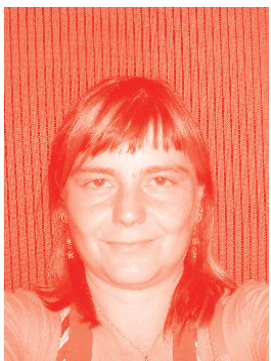

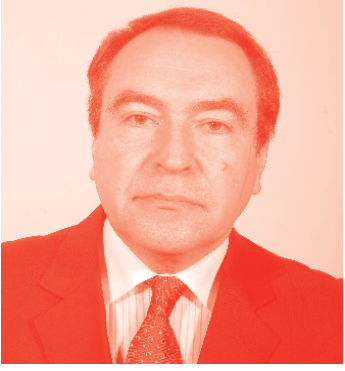
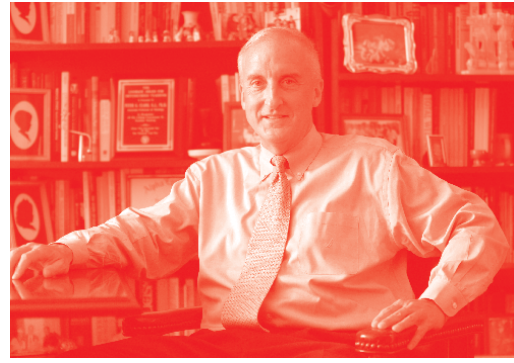
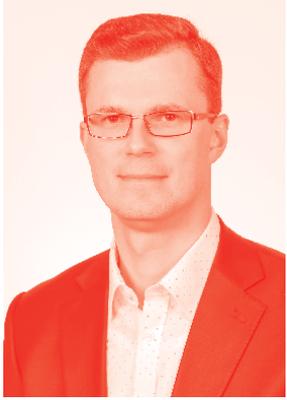
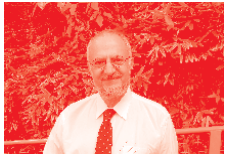
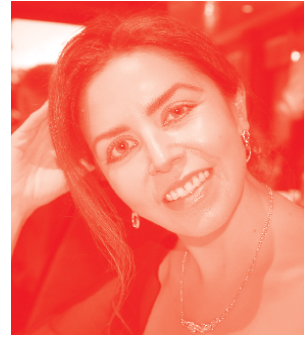

*Supporting open minds since 2005*

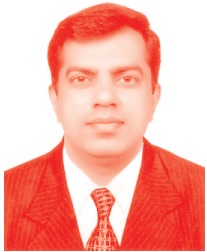
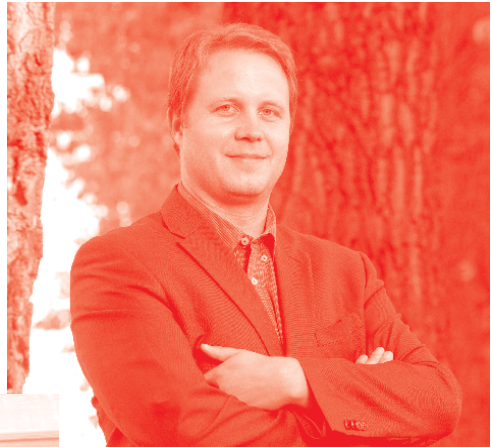
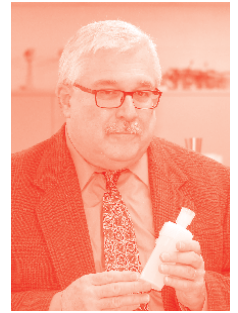
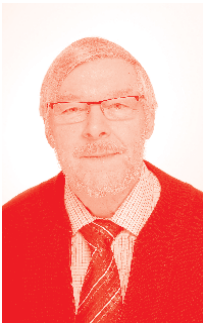
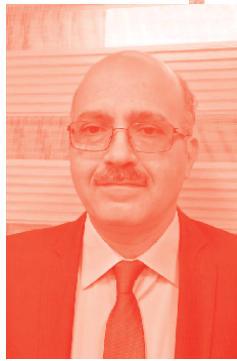
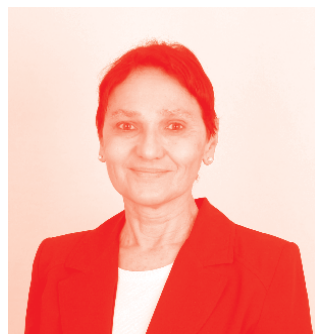
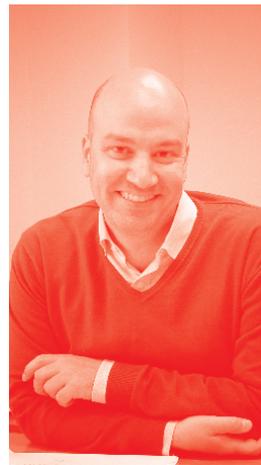

Notice
Statements and opinions expressed in the chapters are these of the individual contributors and not necessarily those of the editors or publisher. No responsibility is accepted for the accuracy of information contained in the published chapters. The publisher assumes no responsibility for any damage or injury to persons or property arising out of the use of any materials, instructions, methods or ideas contained in the book.

# We are IntechOpen,
# the world's leading publisher of Open Access books
# Built by scientists, for scientists

## 5,600+
Open access books available

## 138,000+
International authors and editors

## 170M+
Downloads

## 156
Countries delivered to

Our authors are among the
## Top 1%
most cited scientists

## 12.2%
Contributors from top 500 universities

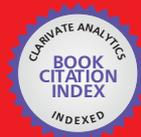

## Interested in publishing with us?
## Contact book.department@intechopen.com

Numbers displayed above are based on latest data collected.
For more information visit www.intechopen.com

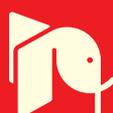

IntechOpen Book Series

# Artificial Intelligence

Volume 7

## Aims and Scope of the Series

Artificial Intelligence (AI) is a rapidly developing multidisciplinary research area that aims to solve increasingly complex problems. In today's highly integrated world, AI promises to become a robust and powerful means for obtaining solutions to previously unsolvable problems. This Series is intended for researchers and students alike interested in this fascinating field and its many applications.

# Meet the Series Editor

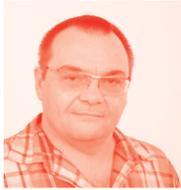 Andries Engelbrecht received the Masters and Ph.D. degrees in Computer Science from the University of Stellenbosch, South Africa, in 1994 and 1999 respectively. He is currently appointed as the Voigt Chair in Data Science in the Department of Industrial Engineering, with a joint appointment as Professor in the Computer Science Division, Stellenbosch University. Prior to his appointment at Stellenbosch University, he has been at the University of Pretoria, Department of Computer Science (1998-2018), where he was appointed as South Africa Research Chair in Artifical Intelligence (2007-2018), the head of the Department of Computer Science (2008-2017), and Director of the Institute for Big Data and Data Science (2017-2018). In addition to a number of research articles, he has written two books, Computational Intelligence: An Introduction and Fundamentals of Computational Swarm Intelligence.

# Meet the Volume Editor

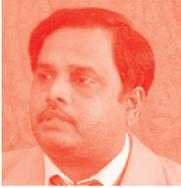 Jaydip Sen is associated with Praxis Business School, Kolkata, India, as a professor in the Department of Data Science. His research areas include security and privacy issues in computing and communication, intrusion detection systems, machine learning, deep learning, and artificial intelligence in the financial domain. He has more than 200 publications in reputed international journals, refereed conference proceedings, and 18 book chapters in books published by internationally renowned publishing houses, such as Springer, CRC press, IGI Global, etc. Currently, he is serving on the editorial board of the prestigious journal Frontiers in Communications and Networks and in the technical program committees of a number of high-ranked international conferences organized by the IEEE, USA, and the ACM, USA. He has been listed among the top 2% of scientists in the world for both the years 2020 and 2021 as of August 2021.

# Contents

# Preface

Machine learning (ML) is the ability of a system to automatically acquire, integrate, and then develop knowledge from large-scale data, and then expand the acquired knowledge autonomously by discovering new information, without being specifically programmed to do so. In short, the ML algorithms can find application in the following: (1) a deeper understanding of the cyber event that generated the data under study, (2) capturing the understating of the event in the form of a model, (3) predicting the future values that will be generated by the event based on the constructed model, and (4) proactively detect any anomalous behavior of the phenomenon so that appropriate corrective actions can be taken beforehand. ML is an evolutionary area, and with recent innovations in technology, especially with the development of smarter algorithms and advances in hardware and storage systems, it has become possible to perform a large number of tasks more efficiently and precisely, which were not even imaginable a couple of decades before. Over the past few years, deep learning (DL), a specialized subset of machine learning that involves more complex architectures, algorithms, and models for solving complex problems and predicting future outcomes of complex events, has also evolved.

Recent times are witnessing rapid development in machine learning algorithm systems, especially in reinforcement learning, natural language processing, computer and robot vision, image processing, speech, and emotional processing and understanding. There are numerous applications of machine learning that have emerged or are evolving at present in several business domains, such as medicines and healthcare, finance and investment, sales and marketing, operations and supply chain, human resources, media and entertainment, and so on.

Of late, the applied ML systems in the industry are exhibiting some prominent trends. These trends will utilize the power of ML and artificial intelligence (AI) systems even further to reap benefits in business and society, in general. Some of these trends are as follows: (1) less volume of code and faster implementation of ML systems; (2) increasing use of light-weight systems suitable for working on the resource-constrained internet of things (IoT) devices; (3) automatic generation of codes for building ML models; (4) designing novel processes for robust management of the development of ML systems for increased reliability and efficiency; (5) more wide-spread adoption of deep-learning solutions into products of all domains and applications; (6) increased use of generative adversarial networks (GAN)-based models for various image processing applications including image searching, image enhancement, etc.; (7) more prominence of unsupervised learning-based systems that require less or no human intervention for their operations; (8) use of reinforcement learning-based systems; and finally, (9) evolution of few-shots, if not zero-shot learning-based systems.

In tune with the increasing importance and relevance of ML models, algorithms, and their applications and with the emergence of more innovative uses-cased of DL- and AI-based systems, the current volume presents a few innovative research works and their applications in real-world, such as stock trading, medical and healthcare systems, and software automation. The chapters in the book illustrate how ML and DL algorithms and models are designed, optimized, and applied for achieving higher precision and efficiency in business and other processes in real-world scenarios.

The book presents six chapters that highlight different architectures, models, algorithms, and applications of machine learning, deep learning, and artificial intelligence. The subject matter discussed in the chapters of the book illustrates the complexities involved in the design, training, validation, testing, and deployment of machine learning and deep learning models in real-world applications.

In the introductory chapter entitled *Introductory Chapter: Machine Learning in Finance-Emerging Trends and Challenges*, Jaydip Sen, Rajdeep Sen, and Abhishek Dutta present a landscape of the emerging trends and challenges in banks and other financial institutions that machine learning-based models will face in the near future and how the organizations should transform those challenges into opportunities for their business.

In Chapter 2, *Design and Analysis of Robust Deep Learning Models for Stock Price Prediction*, Jaydip Sen and Sidra Mehtab present a set of deep learning-based regression models for precise and robust prediction of future prices of a stock listed in the diversified sector in the National Stock Exchange (NSE) of India. The models are built on convolutional neural network (CNN) and long short-term memory (LSTM) network architectures, and they are designed to handle high-frequency stock price data. The performances of the models are compared based on their execution time and prediction accuracies.

In Chapter 3, *Articulated Human Pose Estimation Using Greedy Approach*, Pooja Kherwa, Sonali Singh, Saheel Ahmed, Pranay Berry, and Sahil Khurana propose a method of bottom-up parsing for human pose estimation that localizes anatomical key points of humans. The authors have also presented results on the performance of their proposed mechanism.

In Chapter 4, *Ensemble Machine Learning Algorithms for Prediction and Classification of Medical Images*, Racheal S. Akinbo and Oladunni A. Daramola discuss how ensemble machine learning models can be applied for classifying medical images, such as ultrasound reports, X-rays, mammography, fluoroscopy, computer-aided tomography, magnetic resonance image, magnetic resonance angiography, nuclear medicine, and positron emission tomography.

In Chapter 5, *Delivering Precision Medicine to Patients with Spinal Cord Disorders; Insights into Applications of Bioinformatics and Machine Learning from Studies of Degenerative Cervical Myelopathy*, Kalum J. Ost, David W. Anderson and David W. Cadotte present an approach towards designing a machine learning-based clinical diagnosis system that supports the diagnostic prediction of DCM severity. The proposed system is intended to serve as a support system to the patients rather than a recommendation system. Extensive discussion has been made on the system design, data collection process, data preprocessing, model building, and performance results of the system.

In Chapter 6, *Enhancing Program Management with Predictive Analytics Algorithms (PAAs)*, Bongs Lainjo identifies various practical aspects of software development with a particular focus on predictive model design and deployment. The author also discusses how predictive analytics algorithms can be used to predict future events in different domains of applications, for example, healthcare, manufacturing, education, sports, and agriculture.

I hope that the volume will be very useful for advanced graduate and doctoral students, researchers, practicing data scientists and data engineers, professionals, and consultants working on the broad areas of machine learning, deep learning, artificial intelligence, and other allied areas in computer science. It will also be of use to faculty members and scientists in graduate schools, universities, and research labs engaged in teaching research in machine learning and its applications. Since the volume is not an introductory treatise on the subject, the readers are expected to have basic knowledge of the fundamental principles and theoretical backgrounds of machine learning models and algorithms to understand their advanced applications discussed in the book.

I express my sincere thanks to all authors of the chapters in the volume for their valuable contributions. Without their cooperation and eagerness to contribute, this project would never have been successful. All the authors have been extremely cooperative and punctual during the submission, editing, and publication process of the book. I express our heartfelt thanks to Ms. Marica Novakovic, author service manager of IntechOpen Publishers, for her support, encouragement, patience, and cooperation during the project and her wonderful coordination of all activities involved in it. My sincere thanks also go to Ms. Anja Filipovic, commissioning editor of IntechOpen Publishers, for reposing faith in me and delegating me with the critical responsibility of editorship of such a prestigious academic volume. I will be certainly failing in my duty if I do not acknowledge the encouragement, motivation, and assistance I received from Ms. Sidra Mehtab, my assistant editor and the co-author of one of the chapters in the book. I sincerely thank the valuable contributions received from the coauthors, of the introductory chapter of the book, Mr. Rajdeep Sen and Mr. Abhishek Dutta. The members of my family have always been the sources of my inspiration and motivation for such scholastic work. I dedicate this volume to my beloved sister, Ms. Nabanita Sen, who unfortunately left us on 27 September 2021, due to the deadly disease of cancer. My sister has been always the pillar of strength for me, and this volume also would not have been a success without her constant support and motivation, despite her suffering due to her terminal illness. Last but not the least, I gratefully acknowledge the support and motivation I received from my wife Ms. Nalanda Sen, my daughter Ms. Ritabrata Sen, my mother Ms. Krishna Sen, my brother Mr. Rajdeep Sen, and my sister-in-law, Ms. Toby Kar. Without their support, motivation, and inspiration, the publication of this volume would not have been possible.

**Jaydip Sen**
Professor,
Department of Data Science,
Praxis Business School,
Kolkata, India

# Introductory Chapter: Machine Learning in Finance-Emerging Trends and Challenges

*Jaydip Sen, Rajdeep Sen and Abhishek Dutta*

## 1. Introduction

The paradigm of machine learning and artificial intelligence has pervaded our everyday life in such a way that it is no longer an area for esoteric academics and scientists putting their effort to solve a challenging research problem. The evolution is quite natural rather than accidental. With the exponential growth in processing speed and with the emergence of smarter algorithms for solving complex and challenging problems, organizations have found it possible to harness a humongous volume of data in realizing solutions that have far-reaching business values.

Financial services, banking, and insurance remain one of the most significant sectors that has a very high potential in reaping the benefits of machine learning and artificial intelligence with the availability of rich data, innovative algorithms, and novel methods in its various applications. While the organizations have only skimmed the surface of the rapidly evolving areas such as deep neural networks and reinforcement learning, the possibility of applying these techniques in many applications vastly remains unexplored. Organizations are leveraging the benefits of innovative applications of machine learning in applications like customer segmentation for target marketing of their newly launched products, designing optimal portfolio strategies, detection, and prevention of money laundering and other illegal activities in the financial markets, smarter and effective risk management is credit, adherence to the regulatory frameworks in finance, accounts, and other operations, and so on. However, the full capability of machine learning and artificial intelligence still remains unexplored and unexploited. Leveraging such capabilities will be critical for organizations to achieve and maintain a long-term competitive edge.

While one of the major reasons for the slow adoption of AI/ML models and methods in financial applications is that the algorithms are not well known and there is an inevitable trust deficit in deploying them in critical and privacy-sensitive applications, the so-called "black-box" nature of such models and frameworks that analyzes their internal operations in producing outputs and their validations also impede faster acceptance and deployment of such models in real-world applications.

This introductory chapter highlights some of the challenges and barriers that organizations in the financial services sector at the present encounter in adopting machine learning and artificial intelligence-based models and applications in their day-to-day operations.

The rest of the chapter is organized as follows. Section 2 presents some emerging applications of machine learning in the financial domain. Section 3 highlights emerging computing paradigms in finance. Some important modeling paradigms in the era of machine learning and artificial intelligence are discussed in Section 4. Section 5 discusses some new challenges and barriers currently faced by the financial modelers. Some emerging risks, new choices, and modern practices in financial modeling are presented in Section 6. Finally, Section 7 concludes the chapter.

## 2. Emerging application of machine learning in finance

With the increasing availability and declining cost for complex models executing on high-power computing devices exploiting the unlimited capacity of data storage, the financial industry is geared up to exploit the benefits of machine learning to leverage a competitive business edge. While some of the use cases have already found their applications in the real world, others will need to overcome some existing business and operational challenges before they are deployed. Some of the applications are mentioned below.

**Risk Modeling:** One of the major applications of AI/ML models and algorithms is in the extensive domain of risk modeling and management [1]. While on one hand, the risk modeling credit and market is a critical application of machine learning, on the other hand, a non-finance application such as operational risk management, compliance, and fraud management is also quite important. The majority of the classification approaches and modeling techniques in machine learning such as binary logistic regression, multinomial logistic regression, linear and quadratic discriminant analysis, and decision trees, etc., are the foundational building blocks of applied modeling in the real world. However, in data science applications, the availability of data and its richness play a pivotal role. Hence, in data-rich applications such as credit risk modeling and scoring, designing mortgage schemes, the AI/ML models have already made substantial inroads in comparison to scenarios such as low default credit portfolios for well-known parties that lack the availability of data. Fraud analytics remains another intensive area of AI/ML applications in the non-financial domain.

**Portfolio Management:** The portfolios are designed based on the recommendations of smart algorithms that optimize various parameters with return and risk being the two most important ones [2]. Using the information provided by the users such as their ages of retirement, amount of investment, etc., and other associate details such as their current ages, current assets at hand, the algorithm allocate the invested amount into diverse asset classes to optimize the return and the risk associated with the portfolio. Once an initial allocation is made, the algorithm continuously monitors the market environment and changes the allocation so that the portfolio remains at its optimized level always. These AI-enabled portfolio managers, known as the Robo-advisors are increasingly being used in real-world portfolio design due to their superior adaptability and optimization skill to their human counterparts.

**Algorithmic Trading:** Algorithmic trading exploits the use of algorithms to carry out stock trading in an autonomous manner with minimal human intervention [3]. Invented in 1970, algorithmic trading deploys automated pre-programmed stock trading instructions that can be executed very fast and at a very short time interval (i.e., at a very high frequency) to optimize trading returns and objectives. Machine learning and artificial intelligence have pushed algorithmic trading into a new dimension where not only advanced trading strategies can be made very fast but also deep insights can be made into the stock price and overall market movements. While most hedge funds and financial organizations do not make their trading strategies public, it is well known that machine learning, of late, is playing an increasingly important role in calibrating high-frequency, high-volume trading decisions in real-time for critical applications.

**Fraud detection and analysis:** Fraud detection and analysis is one of the most critical machine learning applications in the finance industry [4]. There is an increased level of security and privacy risk associated with sensitive information both from the organization and personal front due to ubiquitous availability of connectivity, high computational power in devices, an increased amount of data stored and communicated online. These issues have changed the way online fraud analysis and detection are being made. While detection in earlier days used to depend on matching a large set of complex rules, the recent approaches are largely based on the execution of learning algorithms that adapts to new security threats making the detection process more robust while being agile.

**Loan and Insurance Underwritings:** Loans, credit, and insurance underwriting is also an area where the large-scale deployment of machine learning models can be made by financial institutions for achieving competitive advantage [5]. At large banks and insurance firms, the availability of historical data of consumers, financial lending/borrowing information, insurance outcomes, and default-related information in paying debts, can be leveraged in training robust machine learning models. The learned patterns and trends can be exploited by the learning algorithms for lending and underwriting risks in the future to minimize future defaults. The deployment of such models can lead to a paradigm shift in business efficiency and profit. However, at present, there is a limited utilization of such models in the industry as their deployments are largely confined within large financial institutions.

**Financial Chatbots:** Automation in the finance industry is also an outcome of the deployment of machine learning and artificial intelligence. Accessing the relevant data, machine learning models can yield an insightful analysis of the underlying patterns inside them that helps in making effective decisions in the future. In many cases, these models may provide recommended actions for the future so that the business decision can be made in the most efficient and optimum way. AI-based systems in financial applications also can minimize their errors learning fast from their past actions that also reduce wastages of precious resources including time. AI chatbots provide an effective way of interaction with the customers while automating many routine tasks in a financial institution [6].

**Risk Management:** Machine learning techniques are revolutionizing the way corporates handle the risks associated with their operations. Risk management examples are diverse and infinite ranging from deciding about the amount a bank should lend a customer or how to improve compliance to a process or the way risk associated with a model can be minimized [7].

**Asset price prediction:** Predicting future asset prices is one of the most challenging tasks that finance professionals have to do frequently. The price of an asset is affected by numerous factors driven by the market including speculative activities. In the classical approach, an asset price is determined by analyzing historical financial reports and past market performances. With rich data available, of late, machine learning-based models have started playing significant roles in predicting future asset prices in a robust and precise manner [8].

**Derivative pricing:** In the traditional approach, derivative pricing models are designed on numerous assumptions which do not hold good in the real world. These assumptions attempt to establish an empirical relationship among several predictor variables such as the strike price, maturity time, and option type, and the target variable which is the price of the derivative. Machine learning models have gotten rid of the requirement of such assumptions as they attempt to fit in the best possible function between the predictors and the target by minimizing the error. Accuracy and the minimal time needed for the deployment of the models in real-world use cases make machine learning the most impactful paradigm in the task of derivative pricing [9].

**Money Laundering:** As per a report published by the United Nations, it is estimated that 2–5% of the world's aggregated GDP is accounted for the amount of money being laundered worldwide. Machine learning models can find applications in detecting money laundering activities with minimal false-positive cases [10].

## 3. Emerging paradigms in computing in finance

Several new capabilities and approaches and frameworks in machine learning, data science, and artificial intelligence have become available to the modelers and engineers for all disciplines including finance professionals and researchers. Some of them are as follows.

**Virtual Agents:** The machine learning paradigm will witness the increasing deployment of agents in various tasks. These agents have the capability of performing complex data mining tasks through a large set of policy rules, defined procedures and regulations, and provide automated responses to queries.

**Cognitive Robotics:** The robots in the cognitive domain have the power of automating several tasks which are currently done by humans. This automation comes with an additional level of sophistication, speed, and precision in performing the tasks.

**Text Analytics:** The applications of sophisticated algorithms, frameworks, and models of natural language processing in analyzing voluminous and complex financial contracts and documents help processing and decision making faster and more accurately with minimal associated risks.

**Video Analytics:** Advancements in the fields of computer vision, image processing, speech processing, and speech recognition together with the exponential growth in hardware capabilities have led to very promising progress in compliance, audit, model validation in various financial applications including automated generation and presentation of financial reports.

## 4. Emerging trends in modeling techniques

With the increasing proliferation of machine learning models in innovative applications in the financial industry, some computing and modeling paradigms will find more adoption. Some of them are as follows.

**Sparsity-aware learning:** Sparsity-aware learning has evolved as an alternative model regularization approach to address several problems that are usually encountered in machine learning [11]. Considerable effort has been spent in designing such frameworks in an iterative manner for solving estimation tasks of model parameters avoiding overfit.

Iteratively designing schemes such frameworks in solving estimation tasks of model parameters avoiding overfit.

Sparsity-aware learning systems are well-suited in financial modeling applications leading to extremely robust and accurate models for various applications in finance.

**Reproducing Kernel Hilbert Spaces:** Reproducing Kernel Hilbert Spaces (RKHS) is essentially a Hilbert space function that evaluates a continuous function in the linear space [12]. These functions find important applications in statistical learning as every functional representation in RKHS represents minimization of an empirical function embodying the associated risk, and the representation is made as a linear combination of the data points in the training set transformed by the kernel function. Accordingly, RKHS has a very high potential in risk modeling and evaluation in finance.

**Monte Carlo simulation:** This method of modeling provides the modeler with a large range of possible outcomes and probabilities that they will occur for any choice of action that is taken. It is used in a diverse set of domains like finance, energy, project management, and monitoring, research and development, and insurance. It performs risk analysis by designing models of possible results by substituting a range of values – a probability distribution – for any factor that has inherent uncertainty. The ability in handling uncertainty makes this approach particularly popular in modern-day modeling in finance [13].

**Graph theory:** Multivariate financial data pose a very complex challenge in processing and visualization in addition to being difficult in modeling. Graph theory provides the modeler with a very elegant, efficient, and easily interpretable method of handling multivariate financial data [14].

**Particle filtering:** It is a method of modeling nonlinear and non-Gaussian systems with a very high level of precision. Its ability to handle multi-modal data makes it one of the most effective and popular modeling techniques in many fields including finance [15]. Stated in simple words, particle filtering is a technique for identifying the distribution of a population that has a minimum variance by identifying a set of random samples traversing through all the states to obtain a probability density function that best fits into the original distribution and then substituting the integral operation on the function by the mean of the sample.

**Parameter learning and convex paths:** While optimization methods have been proved to be very effective in training large-scale deep neural networks involving millions of parameters, the regularization of these methods has become of paramount importance for proper training of such networks [16]. Accordingly, intensive work has been also carried out in estimating the biases associated with the optimum value of the objective function arrived at by the algorithms. The estimation of such biases provides the modeler with an idea about the degree of inaccuracy in the models for critical applications including financial modeling.

**Deep learning and reinforcement learning:** The application of machine learning in finance has largely been manifested in the form of models built on deep neural network architecture and smarter algorithms for the optimization and training of such networks. Reinforcement learning-based models have made the automation of such models a reality. A vast gamut of applications, such as algo trading, capital asset pricing, stock price prediction, portfolio management can be very effectively designed and executed using deep learning and reinforcement learning frameworks [17–26].

## 5. New challenges in financial modeling

Despite the numerous opportunities and advantages that machine learning and artificial intelligence-based applications are likely to provide to the financial sector, there will be some initial challenges and barriers too. This section highlights some of the challenges as follows.

**Data challenges:** While the availability of data in finance is quite plenty, the time series data in finance (e.g., stock prices) are quite small in size for data-hungry machine learning and deep learning models. Models built on limited time series data are naturally less trained and improperly designed. The result is a sub-optimal performance of the models. Another problem in finance is that financial data cannot be synthesized unlike images in the fields of computer vision and image processing. Since finance data cannot be synthesized, one has to wait for financial data to be produced in the real world before using them in model training and validation. The third challenge with financial data is the high level of noise associated with high-frequency trading data. Since high-frequency data in finance are invariably associated with a high level of noise, the machine learning models trained on such noisy data are intrinsically imprecise. Data evolution in finance poses the fourth challenge. Unlike data in most other fields, where the data features do not change with time, the features in financial data in harmony with financial markets evolve and change with time. This implies that financial variables will not carry the same meaning and significance over a long period, say one decade. The changes in the semantic meaning and significance of financial variables make it particularly difficult for the machine learning model to derive a consistent explanation and learning over a reasonably long period [27].

**Black-box nature of the models:** Machine learning and artificial intelligence-based models are black-box in nature [28]. In these models, while the outputs from the model are available and most of the time, they are easily interpretable, the biggest shortcoming is their lack of power of explanation of the output. In many critical applications in finance, mere outputs are not sufficient, and strong logical support for explaining the output is mandatory to instill sufficient confidence in the minds of the decision-makers. In absence of such explainable characteristics of the machine learning models, it will always remain a difficult job for the modelers to advocate the suitability of such models in critical business use cases.

**Validation challenges of the models:** Due to their higher complexity and opaqueness in operation, the machine learning models pose some significant challenges to risk management and validation [29]. While the regulators demand the machine learning models to comply with the SR 11–7 and OCC 2011–2012 standards of risk management, the optimum execution of the models may not be possible if all those guidelines are to be strictly adhered to. Model risk is an event that occurs when a model is designed following its intended objective but it introduces errors while in execution yielding inaccurate results from the perspective of its design and business use case. Another manifestation of model risk can happen when a model is built and deployed inaccurately or with faults without proper knowledge about its limitations and shortcoming.

**Challenges in model testing and outcome analysis:** The performance of a model and its accuracy in testing are evaluated by outcome analysis [29]. Since the neural network model has a natural tendency to overfit or underfit the data based on the training, it is imperative on the part of the model evaluation team to address the bias-variance trade-offs in the training and validation process. Since the traditional k cross-validation procedure used in backtesting of predictive models does not work effectively for machine learning model validation, the machine learning model validators should take particular care in carrying out normalization and feature selection before model training and validation. Validation loss and accuracy should be strictly monitored and analyzed to ensure that no overfitting or underfitting is present in the model before it is sent to the production phase. Neural network models are also difficult to evaluate on their sensitivity analysis as these models lack explainability. Since establishing a functional relationship between the explanatory variables and the target variable in a neural network model is a difficult task unlike statistical models, sensitivity analysis between the inputs and the output may involve a computationally involved challenge while the results of the analysis may be quite complex.

**Challenges with models designed by vendors:** As per the requirements specified in SR 11–7 and OCC 2011–2012 standards, models supplied by vendors are required to adhere to the same rigor as internally developed models [29]. However, in many practical situations, due to proprietary restrictions testing of vendors supplied models becomes a challenge. For vendor-supplied models, banks and financial institutions will have to mostly rely on softer forms of validation. The softer form of validation may include periodic review of model performance and conceptual soundness, stringent assessment of model customization, review of the development process, and applicability of the model in the portfolio of operations of the bank.

## 6. Emerging risks, new choices, and modern practices

In all domains including finance, the major cause that contributes to the risk in a machine learning model is the complexity associated with the model. The machine learning algorithms are intrinsically very complex as they work on voluminous and possibly unstructured data such as texts, images, and speech. As a consequence, training of such algorithms demands sophisticated computing infrastructure and a high level of knowledge and skill on the part of the modelers. However, countering such complexities with overly sophisticated models can be very counterproductive. Instead of adopting a complex approach, the banks and financial institutions should understand the risks associated with their business and operations and manage them with simple model-validation approaches. The issues that will dominate the risk management landscape for machine learning models in the financial industry are – (i) interpretability of the models, (ii) model bias, (iii) extensive feature engineering, (iv) importance of model hyperparameters, (v) production readiness of models, (vi) dynamic calibration of models, and (vii) explainable artificial intelligence. These issues are briefly discussed below.

**Model interpretability:** Since all machine learning models are essentially black boxes, their interpretability requirements have to be determined by the banks and financial institutions based on the risk associated with their use cases [28]. While some use cases may demand a highly granular and transparent interpretation of the model's working, the requirements for some use cases may not be so stringent. For example, in credit granting by banks to its customers, there may be a clear explanation required to be given by a machine learning model in cases where such credits are refused by the banks. However, another use case that involves minimal risk to the bank's operations such as recommendation of a product sent to the mobile app installed on a customer's hand-held device may not demand any understanding of the reason why the model has made such recommendation.

The model validation process should make sure that models comply with their intended objectives and policies. Despite being black-box in nature, machine learning models come with various provisions for their interpretation. Depending on the type of the model, these approaches may be widely different as discussed below.

Models like linear regression which are not only linear but monotonic in their behavior, the coefficients associated with the explanatory variables exhibit their respective influence on the target variable in the model.

Ensemble models such as ada boosting or gradient boosting, are nonlinear but monotonic in their behavior. In these monotonic models, if the explanatory variables are restricted in their values, the restriction will cause either an increase or a decrease in the value of the target variable. This monotone nature of the models simplifies the contributions of the explanatory variables in the predicted values of the target variable by the model.

The complex deep neural network-based models which are not only nonlinear but also non-monotonic have methods like *Shapley additive explanations* (SHAP), and *local interpretable model agnostic explanations* (LIME) for their global and local interpretability.

**Model bias:** Machine learning models are susceptible to four different types of bias. These biases are, (i) sample bias, (ii) bias in measurement, (iii) bias due to algorithms, and (iv) bias towards or against a particular category of people [29]. For example, the algorithm used in building a random forest model has a bias towards input features that have a more distinct set of values. For example, a model built on a random forest algorithm for assessing potential money laundering activities is likely to be biased towards the features with a higher number of levels in its categories (e.g., occupation), while features having a lower number of categorical levels (e.g., country) might be better suited to detect money laundering activities. To counter the issues about algorithmic bias, the validation processes must be equipped with the ability to select the most appropriate algorithm in a given context. Feature engineering and designing challenger models are some of the methods to counter algorithmic biases in models. The bias against or in favor of a particular group of people can be avoided by defining a fair set of policies by the banks and financial institutions. Models should be tested for fairness for various cases, and necessary corrections should be made in case aberrant outputs are produced by the models.

**Extensive feature engineering:** The task of feature engineering involves much more complications in machine learning and deep learning models than classical statistical models. The factors that contribute to the increased complexity in feature engineering in machine learning are as follows: The first and the most obvious reason is the large number of features involved in machine learning models. First, machine-learning models usually involve a very large number of input variables. The second reason is due to the increased use of unstructured data input in machine learning models. Unstructured data like text, images, and speech involve an enormously large number of features after data preprocessing which add to the complexity in modeling. The use of commercial-off-the-shelf frameworks in modeling such as AutoML has also led to an increased number of features as these frameworks automatically generate derived features from the raw data for providing a better fit of the models into the data in the training phase. While such an approach leads to better training, it is very likely to yield overfitted models in most practical use cases. The banks must have a robust feature engineering strategy in place for mitigating their operational and business risks. The feature strategy is likely to be different for a diverse set of applications. In the case of highly critical and risk-intensive applications like the evaluation of credit-worthiness of customers, every single feature in the model needs to be assessed very closely. On the other hand, for routine applications involving low risks, an overall review of the feature engineering process involving a robust data wrangling step may be sufficient.

**Model hyperparameters:** In machine learning models, the algorithms have parameters associated with them that are not parameters of the models. For example, the number of levels in the constituent decision trees of a random forest (also known as the depth of the trees), or the number of hidden layers in the architecture of a deep learning model must be decided before the models can be trained. Stated differently, the parameters are not determined from the training data of the models, and these are called hyperparameters of the models. The values of hyperparameters in machine learning models are most often determined by trial-and-error methods or brute force search methods like grid search. The absence of an exact and efficient way of finding the optimum values of the hyperparameters makes designing machine learning and deep learning models a particularly challenging task as a wrong choice of values of the hyperparameters can lead to imprecise models. Of late, banks and other financial institutions are depending more on sophisticated binary classification models built on support vector machines with the additional capability of text analytics for analyzing customer complaints. However, these models will be difficult to generalize for a multitude of applications as they will be sensitive to the kernel used in the training.

**Production readiness of models:** Unlike statistical models which are designed as an aggregation of rules of codes to be executed in a production system, machine learning models are built on algorithms requiring intensive computations most of the time. However, in many situations, financial model developers ignore this important point and tend to build overly complex models only to find later that the production systems of the banks are unable to support such complexity. A very simple example could be a highly complex deep learning model built for detecting frauds in financial transactions that is unable to satisfy the latency constraint in its response. To ensure that the machine learning models are production-ready, the validation step should make a reliable estimation of the volume of data that the model will require to process on the production architecture [30].

**Dynamic calibration of models:** There is an adaptable class of machine learning models that can dynamically modify their parameters based on the patterns in the data [31]. An obvious example is a model built on the reinforcement learning approach. While the autonomous learning of such models is a great boon, there is also a new type of risk associated with such models. In absence of sufficient external control and with too much emphasis on learning from the short-term patterns on the data, the long-term performance of the models may be adversely affected. The consequence is an additional complexity in financial modeling – when to recalibrate the model dynamically? The dynamic recalibration time will also need to be adapted for different applications like algorithmic trading, creditworthiness determination, etc. A comprehensive validation process should now include a set of policies that will guide the modelers in evaluating dynamic recalibration so that the model can adhere to its intended objectives while appropriate controls are in place for ensuring that risks are mitigated when they emerge. This is surely not going to be an easy task as it will involve the complex job of thresholding, identifying the health of the models based on their critical metrics of performance on out-of-sample data.

**Explainable artificial intelligence:** In the explainable artificial intelligence paradigm, an AI program scans through the code of another AI program and attempts to explain the operating steps and the output yielded by the latter program. This approach can be exploited to adjust the values of the explanatory variables in a predictive model so that the desired value of the target variable (i.e., the output of the model) is obtained [32]. In this way, explainable AI provides a very easy and elegant way to record, analyze and interpret the learning method of a complex model and for repeating the same in the future. Although the computing paradigm is still in research labs, its adoption in a commercial environment especially in the financial industries is not far away.

## 7. Conclusion

In the days to come, the financial industry will show increasingly more reliance on machine learning and artificial intelligence-based emerging methods and models to leverage competitive advantages. While the regulatory and compliance will evolve into a more standardized framework, machine learning will continue to provide the banks and other financial institutions more opportunities to explore and exploit emerging applications, while being more efficient in delivering the existing services. While the emerging techniques discussed in the chapter will play their critical roles in mitigating future risks in models, they will also guide the authorities in designing effective regulations and compliance frameworks in risk-intensive applications like creditworthiness assessment, trade surveillance, and capital asset pricing. The model validation process will increasingly be adapted to mitigate machine learning risks, while considerable effort and time will be spent in fine-tuning the model hypermeters in handling emerging applications. However, banks will have more opportunities to deploy the models in a large gamut of applications, gaining competitive business advantages and mitigating risks in operations.

## Author details

Jaydip Sen[1]*, Rajdeep Sen[2] and Abhishek Dutta[3]

1 Department of Data Science, Praxis Business School, Kolkata, India

2 Independent Researcher and Financial Analyst

3 School of Computing and Analytics, NASHM Knowledge Campus, Kolkata, India

*Address all correspondence to: jaydip.sen@acm.org

**IntechOpen**

# Design and Analysis of Robust Deep Learning Models for Stock Price Prediction

*Jaydip Sen and Sidra Mehtab*

## Abstract

Building predictive models for robust and accurate prediction of stock prices and stock price movement is a challenging research problem to solve. The well-known efficient market hypothesis believes in the impossibility of accurate prediction of future stock prices in an efficient stock market as the stock prices are assumed to be purely stochastic. However, numerous works proposed by researchers have demonstrated that it is possible to predict future stock prices with a high level of precision using sophisticated algorithms, model architectures, and the selection of appropriate variables in the models. This chapter proposes a collection of predictive regression models built on deep learning architecture for robust and precise prediction of the future prices of a stock listed in the diversified sectors in the National Stock Exchange (NSE) of India. The Metastock tool is used to download the historical stock prices over a period of two years (2013–2014) at 5 minutes intervals. While the records for the first year are used to train the models, the testing is carried out using the remaining records. The design approaches of all the models and their performance results are presented in detail. The models are also compared based on their execution time and accuracy of prediction.

**Keywords:** Stock Price Forecasting, Deep Learning, Univariate Analysis, Multivariate Analysis, Time Series Regression, Root Mean Square Error (RMSE), Long-and-Short-Term Memory (LSTM) Network, Convolutional Neural Network (CNN)

## 1. Introduction

Building predictive models for robust and accurate prediction of stock prices and stock price movement is a very challenging research problem. The well-known efficient market hypothesis precludes any possibility of accurate prediction of future stock prices since it assumes stock prices to be purely stochastic in nature. Numerous works in the finance literature have shown that robust and precise prediction of future stock prices is using sophisticated machine learning and deep learning algorithms, model architectures, and selection of appropriate variables in the models.

Technical analysis of stocks has been a very interesting area of work for the researchers engaged in security and portfolio analysis. Numerous approaches to technical analysis have been proposed in the literature. Most of the algorithms here work on searching and finding some pre-identified patterns and sequences in the time series of stock prices. Prior detection of such patterns can be useful for the investors in the stock market in formulating their investment strategies in the market to maximize their profit. A rich set of such patterns has been identified in the finance literature for studying the behavior of stock price time series.

In this chapter, we propose a collection of forecasting models for predicting the prices of a critical stock of the automobile sector of India. The predictive framework consists of four CNN regression models and six models of regression built on the *long-and-short-term memory* (LSTM) architecture. Each model has a different architecture, different shapes of the input data, and different hyperparameter values.

The current work has the following three contributions. First, unlike the currently existing works in the literature, which mostly deal with time-series data of daily or weekly stock prices, the models in this work are built and tested on stock price data at a small interval of 5 minutes. Second, our propositions exploit the power of deep learning, and hence, they achieve a very high degree of precision and robustness in their performance. Among all models proposed in this work, the lowest ratio of the *root mean square error* (RMSE) to the average of the target variable is 0.006967. Finally, the speed of execution of the models is very fast. The fastest model requires 174.78 seconds for the execution of one round on the target hardware platform. It is worth mentioning here that the dataset used for training has 19500 records, while models are tested on 20500 records.

The chapter is organized as follows. Section 2 briefly discusses some related works in the literature. In Section 3, we discuss the method of data acquisition, the methodology followed, and the design details of the ten predictive models proposed by us. Section 4 exhibits the detailed experimental results and their analysis. A comparative study of the performance of the models is also made. In Section 5, we conclude the chapter and identify a few new directions of research.

## 2. Related work

The literature on systems and methods of stock price forecasting is quite rich. Numerous proposals exist on the mechanisms, approaches, and frameworks for predicting future stock prices and stock price movement patterns. At a broad level, these propositions can be classified into four categories. The proposals of the first category are based on different variants of univariate and multivariate regression models. Some of the notable approaches under this category are - *ordinary least square* (OLS) regression, *multivariate adaptive regression spline* (MARS), *penalty-based regression*, *polynomial regression*, etc. [1–4]. These approaches are not, in general, capable of handling the high degree of volatility in the stock price data. Hence, quite often, these models do not yield an acceptable level of accuracy in prediction. *Autoregressive integrated moving average* (ARIMA) and other approaches of econometrics such as *cointegration*, *vector autoregression* (VAR), *causality tests*, and *quantile regression* (QR), are some of the methods which fall under the second category of propositions [5–16]. The methods of this category are superior to the simple regression-based methods. However, if the stock price data are too volatile and exhibit strong randomness, the econometric methods also are found to be inadequate, yielding inaccurate forecasting results. The learning-based approach is the salient characteristic of the propositions of the third category. These proposals are based on various algorithms and architectures of machine learning, deep learning, and reinforcement learning [17–41]. Since the frameworks under this category use complex predictive models working on sophisticated algorithms and architectures, the prediction accuracies of these models are found to be quite accurate in real-world applications. The propositions of the fourth category are broadly based on hybrid models built of machine learning and deep learning algorithms and architectures and also on the relevant inputs of sentiment and news items extracted from the social web [42–47]. These models are found to yield the most accurate prediction of future stock prices and stock price movement patterns. The *information-theoretic* approach and the *wavelet* analysis have also been proposed in stock price prediction [48, 49]. Several *portfolio optimization* methods have also been presented in some works using forecasted stock returns and risks [50–55].

In the following, we briefly discuss the salient features of some of the works under each category. We start with the regression-based proposals.

Enke et al. propose a multi-step approach to stock price prediction using a multiple regression model [2]. The proposition is based on a *differential-evolution-based fuzzy clustering* model and a fuzzy neural network. Ivanovski et al. present a linear regression and correlation study on some important stock prices listed in the Macedonian Stock Exchange [3]. The results of the work indicate a strong relationship between the stock prices and the index values of the stock exchange. Sen and Datta Chaudhuri analyze the trend and the seasonal characteristics of the capital goods sector and the small-cap sector of India using a time series decomposition approach and a linear regression model [4].

Among the econometric approaches, Du proposes an integrated model combining an ARIMA and a backpropagation neural network for predicting the future index values of the Shanghai Stock Exchange [6]. Jarrett and Kyper present an ARIMA-based model for predicting future stock prices [7]. The study conducted by the authors reveals two significant findings: (i) higher accuracy is achieved by models involving fewer parameters, and (ii) the daily return values exhibit a strong autoregressive property. Sen and Datta Chaudhuri different sectors of the Indian stock market using a time series decomposition approach and predict the future stock prices using different types of ARIMA and regression models [9–14, 33]. Zhong and Enke present a gamut of econometric and statistical models, including ARIMA, *generalized autoregressive conditional heteroscedasticity* (GARCH), *smoothing transition autoregressive* (STAR), *linear* and *quadratic discriminant analysis* [16].

Machine learning and deep learning models have found widespread applications in designing predictive frameworks for stock prices. Baek and Kim propose a framework called ModAugNet, which is built on an LSTM deep learning model [17]. Chou and Nguyen preset a *sliding window metaheuristic optimization* method for stock price prediction [19]. Gocken et al. propose a hybrid artificial neural network using harmony search and genetic algorithms to analyze the relationship between various technical indicators of stocks and the index of the Turkish stock market [21]. Mehtab and Sen propose a gamut of models designed using machine learning and deep learning algorithms and architectures for accurate prediction of future stock prices and movement patterns [22–28, 34, 35]. The authors present several models which are built on several variants of *convolutional neural networks* (CNNs) and *long-and-short-term memory networks* (LSTMs) that yield a very high level of prediction accuracy. Zhang et al. present a multi-layer perceptron for financial data mining that is capable of recommending buy or sell strategies based on forecasted prices of stocks [40].

The hybrid models use relevant information in the social web and exploit the power of machine learning and deep learning architectures and algorithms for making predictions with a high level of accuracy. Among some well-known hybrid models, Bollen et al. present a scheme for computing the mood states of the public from the Twitter feeds and use the mood states information as an input to a nonlinear regression model built on a *self-organizing fuzzy neural network* [43]. The model is found to have yielded a prediction accuracy of 86%. Mehtab and Sen propose an LSTM-based predictive model with a sentiment analysis module that analyzes the public sentiment on Twitter and produces a highly accurate forecast of future stock prices [45]. Chen et al. present a scheme that collects relevant news articles from the web, converts the text corpus into a word feature set, and feeds the feature set of words into an LSTM regression model to achieve a highly accurate prediction of the future stock prices [44].

The most formidable challenge in designing a robust predictive model with a high level of precision for stock price forecasting is handling the randomness and the volatility exhibited by the time series. The current work utilizes the power of deep learning models in feature extraction and learning while exploiting their architectural diversity in achieving robustness and accuracy in stock price prediction on very granular time series data.

## 3. Methodology

We propose a gamut of predictive models built on deep learning architectures. We train, validate, and then test the models based on the historical stock price records of a well-known stock listed in the NSE, viz. *Century Textiles*. The historical prices of Century Textiles stock from 31st Dec 2012, a Monday to 9th Jan 2015, a Friday, are collected at 5 minutes intervals using the Metastock tool [56]. We carry out the training and validation of the models using the stock price data from 31st Dec 2012 to 30th Dec 2013. The models are tested based on the records for the remaining period, i.e., from 31st Dec 2013, to 9th Jan 2015. For maintaining uniformity in the sequence, we organize the entire dataset as a sequence of daily records arranged on a weekly basis from Monday to Friday. After the dataset is organized suitably, we split the dataset into two parts – the training set and the test set. While the training dataset consists of 19500 records, there are 20500 tuples in the test data. Every record has five attributes – open, high, low, close, and volume. We have not considered any adjusted attribute (i.e., adjusted close, adjusted volume, etc.) in our analysis.

In this chapter, we present ten regression models for stock price forecasting using a deep learning approach. For the univariate models, the objective is to forecast the future values of the variable *open* based on its past values. On the other hand, for the multivariate models, the job is to predict the future values of *open* using the historical values of all the five attributes in the stock data. The models are tested following an approach known as *multi-step prediction using a walk-forward validation* [22]. In this method, we use the training data for constructing the models. The models are then used for predicting the daily *open* values of the stock prices for the coming week. As a week completes, we include the actual stock price records of the week in the training dataset. With this extended training dataset, the open values are forecasted with a forecast horizon of 5 days so that the forecast for the days in the next week is available. This process continues till all the records in the test dataset are processed.

The suitability of CNNs in building predictive models for predicting future stock prices has been demonstrated in our previous work [22]. In the current work, we present a gamut of deep learning models built on CNN and LSTM architectures and illustrate their efficacy and effectiveness in solving the same problem.

CNNs perform two critical functions for extracting rich feature sets from input data. These functions are: (1) *convolution* and (2) *pooling or sub-sampling* [57]. A rich set of features is extracted by the convolution operation from the input, while the sub-sampling summarizes the salient features in a given locality in the feature space. The result of the final sub-sampling in a CNN is passed on to possibly multiple dense layers. The fully connected layers learn from the extracted features. The fully connected layers provide the network with the power of prediction.

LSTM is an adapted form of a *recurrent neural network* (RNN) and can interpret and then forecast sequential data like text and numerical time series data [57]. The networks have the ability to memorize the information on their past states in some designated cells in memory. These memory cells are called *gates*. The information on the past states, which is stored in the memory cells, is aggregated suitably at the forget gates by removing the irrelevant information. The input gates, on the other hand, receive information available to the network at the current timestamp. Using the information available at the input gates and the forget gates, the computation of the predicted values of the target variable is done by the network. The predicted value at each timestamp is made available through the output gate of the network [57].

The deep learning-based models we present in this paper differ in their design, structure, and dataflows. Our proposition includes four models based on the CNN architecture and six models built on the LSTM network architecture. The proposed models are as follows. The models have been named following a convention. The first part of the model's name indicates the model type (CNN or LSTM), the second part of the name indicates the nature of the input data (univariate or multivariate). Finally, the third part is an integer indicating the size of the input data to the model (5 or 10). The ten models are as follows:

(i) CNN_UNIV_5 – a CNN model with an input of univariate open values of stock price records of the last week, (ii) CNN_UNIV_10 – a CNN model with an input of univariate open values of stock price records of the last couple of weeks, (iii) CNN_MULTV_10 – a CNN model with an input of multivariate stock price records consisting of five attributes of the last couple of weeks, where each variable is passed through a separate channel in a CNN, (iv) CNN_MULTH_10 – a CNN model with the last couple of weeks' multivariate input data where each variable is used in a dedicated CNN and then combined in a multi-headed CNN architecture, (v) LSTM_UNIV_5 – an LSTM with univariate open values of the last week as the input, (vi) LSTM_UNIV_10 – an LSTM model with the last couple of weeks' univariate open values as the input, (vii) LSTM_UNIV_ED_10 – an LSTM having an encoding and decoding ability with univariate open values of the last couple of weeks as the input, (viii) LSTM_MULTV_ED_10 – an LSTM based on encoding and decoding of the multivariate stock price data of five attributes of the last couple of weeks as the input, (ix) LSTM_UNIV_CNN_10 – a model with an encoding CNN and a decoding LSTM with univariate open values of the last couple of weeks as the input, and (x) LSTM_UNIV_CONV_10 – a model having a convolutional block for encoding and an LSTM block for decoding and with univariate open values of the last couple of weeks as the input.

We present a brief discussion on the model design. All the hyperparameters (i.e., the number of nodes in a layer, the size of a convolutional, LSTM or pooling layer, etc.) used in all the models are optimized using grid-search. However, we have not discussed the parameter optimization issues in this work.

## 3.1 The CNN_UNIV_5 model

This CNN model is based on a univariate input of *open* values of the last week's stock price records. The model forecasts the following five values in the sequence as the predicted daily *open* index for the coming week. The model input has a shape (5, 1) as the five values of the last week's daily *open* index are used as the input. Since the input data for the model is too small, a solitary convolutional block and a subsequent max-pooling block are deployed. The convolutional block has a feature space dimension of 16 and the filter (i.e., the kernel) size of 3. The convolutional block enables the model to read each input three times, and for each reading, it extracts 16 features from the input. Hence, the output data shape of the convolutional block is (3,16). The max-pooling layer reduces the dimension of the data by a factor of 1/2. Thus, the max-pooling operation transforms the data shape to (1, 16). The result of the max-pooling layer is transformed into an array structure of one-dimension by a flattening operation. This one-dimensional vector is then passed through a *dense layer* block and fed into the final output layer of the model. The output layer yields the five forecasted *open* values in sequence for the coming week. A batch size of 4 and an epoch number of 20 are used for training the model. The *rectified linear unit* (ReLU) activation function and the Adam optimizer for the *gradient descent algorithm* are used in all layers except the final output layer. In the output layer of the model, the sigmoid is used as the activation function. The use of the activation function and the optimizer is the same for all the models. The schematic architecture of the model is depicted in **Figure 1**.

We compute the number of trainable parameters in the CNN_UNIV_5 model. As the role of the input layer is to provide the input data to the network, there is no learning involved in the input layer. There is no learning in the pooling layers as all these layers do is calculate the local aggregate features. The flatten layers do not involve any learning as well. Hence, in a CNN model, the trainable parameters are involved only in the convolutional layers and the dense layers. The number of trainable parameters ($n_1$) in a one-dimensional convolutional layer is given by (1), where $k$ is the kernel size, and $d$ and $f$ are the sizes of the feature space in the previous layer and the current layer, respectively. Since each element in the feature space has a bias, the term 1 is added in (1)

$$n_1 = (k * d + 1) * f \tag{1}$$

The number of parameters ($n_2$) in a dense layer of a CNN is given by (2), in which $p_{current}$ and $p_{previous}$ refer to the node count in the current layer and the previous layer, respectively. The second term on the right-hand side of (2) refers to the *bias* terms for the nodes in the current layer.

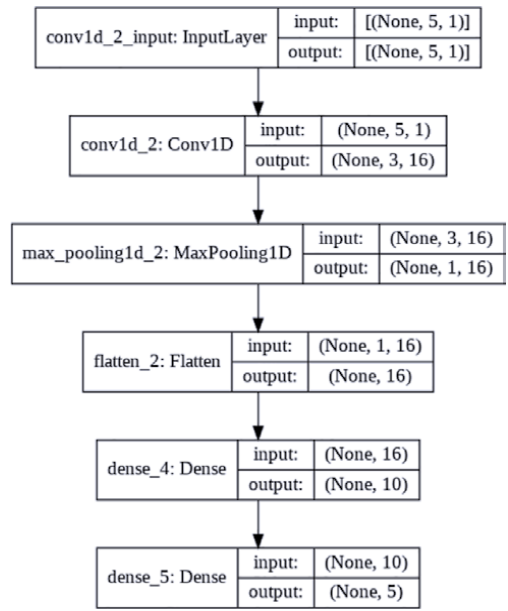

**Figure 1.**
*The schematic architecture of the model CNN_UNIV_5.*

$$n_2 = \left(p_{curr} * p_{prev}\right) + 1 * p_{curr} \tag{2}$$

The computation of the number of parameters in the CNN_UNIV_5 model is presented in **Table 1**. It is observed that the model involves 289 trainable parameters. The number of parameters in the convolutional layer is 64, while the two dense layers involve 170 and 55 parameters, respectively.

## 3.2 The CNN_UNIV_10 model

This model is based on a univariate input of the *open* values of the last couple of weeks' stock price data. The model computes the five forecasted daily *open* values in sequence for the coming week. The structure and the data flow for this model are identical to the CNN_UNIV_5 model. However, the input of the model has a shape of (10, 1). We use 70 epochs and 16 batch-size for training the model. **Figure 2** shows the architecture of the model CNN_UNIV_10. The computation of the number of parameters in the model CNN_UNIV_10 is exhibited in **Table 2.**

| Layer | $k$ | $d$ | $f$ | $p_{\text{prev}}$ | $p_{\text{curr}}$ | $n1$ | $n2$ | #params |
|---|---|---|---|---|---|---|---|---|
| Conv1D (conv1d) | 3 | 1 | 16 | | | 64 | | 64 |
| Dense (dense) | | | | 16 | 10 | | 170 | 170 |
| Dense (dense_1) | | | | 10 | 5 | | 55 | 55 |
| Total #parameters | | | | | 289 | | | |

**Table 1.**
*Computation of the number of params in the model CNN_UNIV_5.*

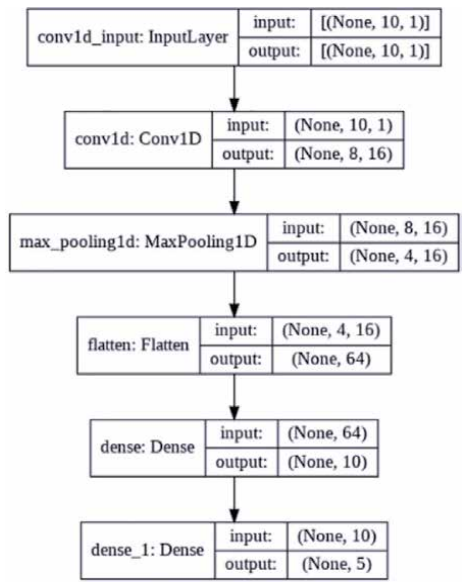

**Figure 2.**
*The architecture of the model CNN_UNIV_10.*

| Layer | *k* | *d* | *f* | $p_{prev}$ | $p_{curr}$ | *n1* | *n2* | **#params** |
|---|---|---|---|---|---|---|---|---|
| Conv1D (conv1d) | 3 | 1 | 16 | | | 64 | | 64 |
| Dense (dense) | | | | 64 | 10 | | 650 | 650 |
| Dense (dense_1) | | | | 10 | 5 | | 55 | 55 |
| **Total #parameters** | | | | | **769** | | | |

**Table 2.**
*The number of parameters in the model CNN_UNIV_10 model.*

It is evident from **Table 2** that the CNN_UNIV_10 involves 769 trainable parameters. The parameter counts for the convolutional layer, and the two dense layers are 64, 650, and 55 respectively.

### 3.3 The CNN_MULTV_10 model

This CNN model is built on the input of the last two weeks' multivariate stock price records data. The five variables of the stock price time series are used in a CNN in five separate channels. The model uses a couple of convolutional layers, each of size (32, 3). The parameter values of the convolutional blocks indicate that 32 features are extracted from the input data by each convolutional layer using a feature map size of 32 and a filter size of 3. The input to the model has a shape of (10, 5), indicating ten records, each record having five features of the stock price data. After the first convolutional operation, the shape of the data is transformed to (8, 32). The value 32 corresponds to the number of features extracted, while the value 8 is obtained by the formula: $f = (k - n) + 1$, where, $k = 10$, $n = 3$, hence, $f = 8$. Similarly, the output data shape of the second convolutional layer is (6, 32). A max-pooling layer reduces the feature space size by a factor of 1/2 producing an output data shape of (3, 32). The max-pooling block's output is then passed on to a third convolutional layer with a feature map of 16 and a kernel size of 3. The data shape of the output from the third convolutional layer becomes (1, 16) following the same computation rule. Finally, another max-pooling block receives the results of the final convolutional layer. This block does not reduce the feature space since the input data shape to it already (1, 16). Hence, and the output of the final max-pooling layer remains unchanged to (1,16). A flatten operation follows that converts the 16 arrays containing one value to a single array containing 16 values. The output of the flatten operation is passed on to a fully connected block having 100 nodes. Finally, the output block with five nodes computes the predicted daily open index of the coming week. The epochs size and the batch size used in training the model are 70 and 16, respectively. **Figure 3** depicts the CNN_MULTV_10 model. **Table 3** shows the computation of the number of trainable parameters involved in the model.

From **Table 3**, it is observed that the total number of trainable parameters in the model CNN_MULTV_10 is 7373. The three convolutional layers *conv1d_4*, *conv1d_5*, and *conv1d_6* involve 512, 3014, and 1552 parameters, respectively. It is to be noted that the value of *k* for the first convolutional layer, *conv1d_4*, is multiplied by a factor of five since there are five attributes in the input data for this layer. The two dense layers, *dense_3* and *dense_4* include 1700 and 505 parameters, respectively.

### 3.4 The CNN_MULTH_10 model

This CNN model uses a dedicated CNN block for each of the five input attributes in the stock price data. In other words, for each input variable, a separate CNN is

Design and Analysis of Robust Deep Learning Models for Stock Price Prediction
DOI: http://dx.doi.org/10.5772/intechopen.99982

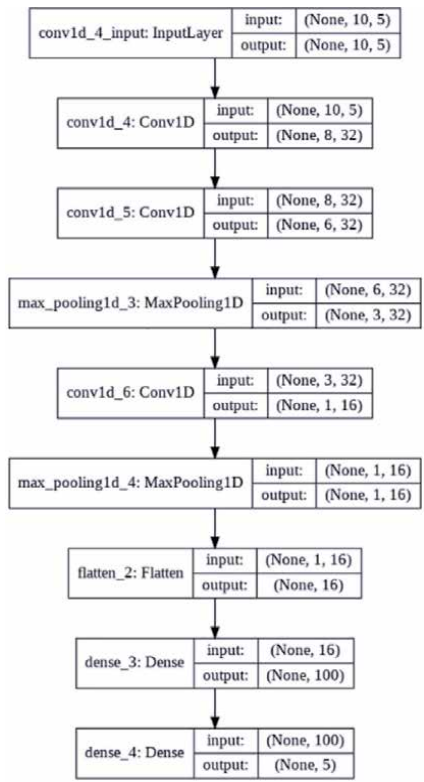

**Figure 3.**
*The schematic architecture of the model CNN_MULTV_10.*

| Layer | k | d | f | $p_{\text{prev}}$ | $p_{\text{curr}}$ | n1 | n2 | #params |
|---|---|---|---|---|---|---|---|---|
| Conv1D (conv1d_4) | 3*5 | 1 | 32 | | | 512 | | 512 |
| Conv1D (conv1d_5) | 3 | 32 | 32 | | | 3104 | | 3014 |
| Conv1D (conv1d_6) | 3 | 32 | 16 | | | 1552 | | 1552 |
| Dense (dense_3) | | | | 16 | 100 | | 1700 | 1700 |
| Dense (dense_4) | | | | 100 | 5 | | 505 | 505 |
| Total #parameters | | | | | 7373 | | | |

**Table 3.**
*The number of parameters in the model CNN_MULTV_10.*

used for feature extrication. We call this a multivariate and multi-headed CNN model. For each sub-CNN model, a couple of convolutional layers were used. The convolutional layers have a feature space dimension of 32 and a filter size (i.e., kernel size) of 3. The convolutional layers are followed by a max-pooling layer. The size of the feature space is reduced by a factor of 1/2 by the max-pooling layer. Following the computation rule discussed under the CNN_MULTV_10 model, the data shape of the output from the max-pooling layer for each sub-CNN model is (3, 32). A flatten operation follows converting the data into a single-dimensional array of size 96 for each input variable. A concatenation operation follows that concatenates the five arrays, each containing 96 values, into a single one-dimensional array of size 96*5 = 480. The output of the concatenation operation is passed successively through two dense layers containing 200 nodes and 100 nodes, respectively. In the end, the output layer having five nodes yields the forecasted five values as the daily *open* stock prices for the coming week. The epoch number and the batch size used in training the model are 70 and 16, respectively. **Figure 4** shows the structure and data flow of the CNN_MULTH_10 model.

**Table 4** presents the necessary calculations for finding the number of parameters in the CNN_MULTH_10 model. Each of the five convolutions layers, *conv1d_1*, *conv1d_3*, *conv1d_5*, *conv1d_7*, and *convid_9*, involves 128 parameters. For each of these layers, $k$ = 3, $d$ = 1 and $f$ = 3, and hence the number of trainable parameters is: (3 * 1 + 1) * 32 = 128. Hence, for the five convolutional layers, the total number of parameters is 128 * 5 = 640. Next, for each of the five convolutional layers, conv1d_2, conv1d_4, conv1d_6, conv1d_8, and con1d_10, involves 3104. Each layer of this group has $k$ = 3, $d$ = 32, and $f$ = 32. Hence the number of trainable parameters for each layer is: (3*32 + 1) * 32 = 3104. Therefore, for the five convolutional layers, the total number of parameters is 3104 * 5 = 15, 520. The dense layers, *dense_1*, *dense_2*, and *dense_3* involve 96200, 20100, and 505 parameters using (2). Hence, the model includes 132,965 parameters.

### 3.5 The LSTM_UNIV_5 model

This model is based on an input of the univariate information of the *open* values of the last week's stock price records. The model predicts the future five values in sequence as the daily *open* index for the coming week. The input has a shape of (5, 1) that indicates that the previous week's daily *open* index values are passed as the input. An LSTM block having 200 nodes receives that data from the input layer. The number of nodes at the LSTM layer is determined using the *grid-search*. The results of the LSTM block are passed on to a fully connected layer (also known as a dense layer) of 100 nodes. Finally, the output layer containing five nodes receives the output of the dense layer and produces the following five future values of open for the coming week. In training the model, 20 epochs and 16 batch-size are used. **Figure 5** presents the structure and data flow of the model.

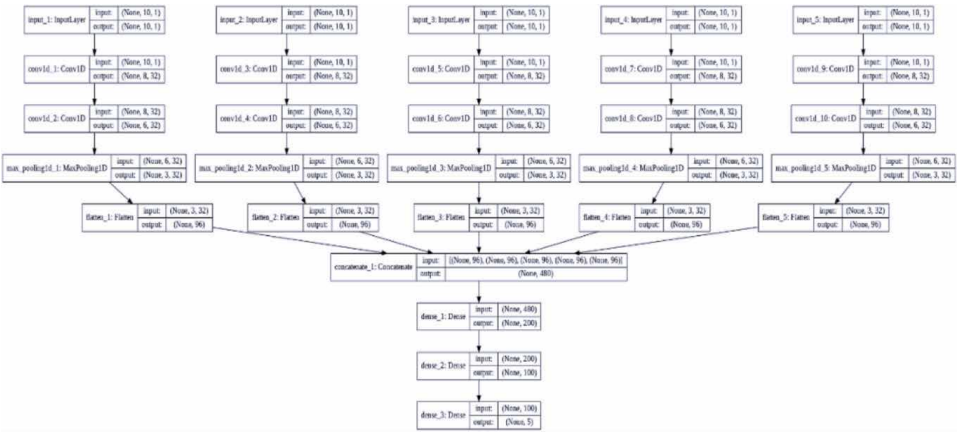

**Figure 4.**
*The schematic architecture of the model CNN_MULTH_10.*

| Layer | k | d | f | $p_{prev}$ | $p_{curr}$ | n1 | n2 | #params |
|---|---|---|---|---|---|---|---|---|
| Conv1D (conv1d_1, conv1d_3, conv1d_5, conv1d_7, conv1d_9) | 3 | 1 | 32 | | | 640 | | |
| | | | | | | | | 640 |
| Conv1D (conv1d_2, convid_4, conv1d_6, conv1d_8, conv1d_10) | 3 | 32 | 32 | | | 15520 | | 15520 |
| Dense (dense_1) | | | | 480 | 200 | | 96200 | 96200 |
| Dense (dense_2) | | | | 200 | 100 | | 20100 | 20100 |
| Dense (dense_3) | | | | 100 | 5 | | 505 | 505 |
| **Total #params** | | | | | 132965 | | | |

**Table 4.**
*The number of parameters in the model CNN_MULTH_10 model.*

As we did in the case of the CNN models, we now compute the number of parameters involved in the LSTM model. The input layers do not have any parameters, as the role of these layers is to just receive and forward the data. There are four gates in an LSTM network that have the same number of parameters. These four gates are known as (i) *forget gate*, (ii) *input gate*, (iii) *input modulation gate*, and the *output gate*. The number of parameters ($n_1$) in each of the gates in an LSTM network is computed using (3), where $x$ denotes the number of LSTM units, and $y$ is the input dimension (i.e., the number of features in the input data)

$$n_1 = (x + y) * x + x \qquad (3)$$

Hence, the total number of parameters in an LSTM layer will be given by 4 * $n_1$. The number of parameters ($n_2$) in a dense layer of an LSTM network is computed using (4), where $p_{prev}$ and $p_{curr}$ are the number of nodes in the previous layer and the current layer, respectively. The bias parameter of each node in the current layer is represented by the last term on the right-hand side of (4).

$$n_2 = \left( p_{prev} * p_{curr} + p_{curr} \right) \qquad (4)$$

The computation of the number of parameters associated with the model LSTM_UNIV_5 is depicted in **Table 5**. In **Table 5**, the number of parameters in the LSTM layer is computed as follows: 4*[(200 + 1) * 200 + 200] = 161,600. The number of parameters in the dense layer, dens_4 is computed as: (200 * 100 + 100) = 20,100. Similarly, the parameters in the dense layers, *dense_5* and

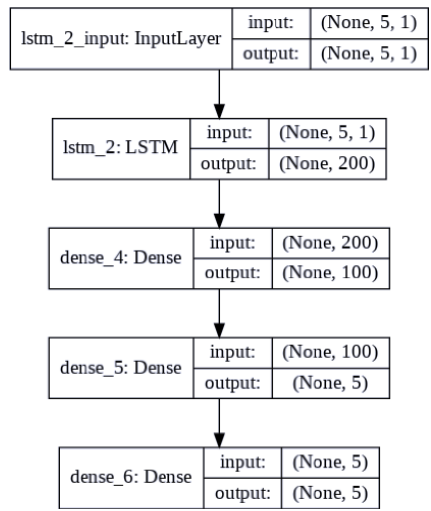

**Figure 5.**
*The schematic architecture of the model LSTM_UNIV_5.*

| Layer | $x$ | $y$ | $p_{\text{prev}}$ | $p_{\text{curr}}$ | n1 | n2 | #params |
|-------|-----|-----|-------|-------|-----|-----|---------|
| LSTM (lstm_2) | 200 | 1 | | | 40,400 | | 161600 |
| Dense (dense_4) | | | 200 | 100 | | 20100 | 20100 |
| Dense (dense_5) | | | 100 | 5 | | 505 | 505 |
| Desne (dense_6) | | | 5 | 5 | | 30 | 30 |
| Total #parameters | | | | 182235 | | | |

**Table 5.**
*The number of parameters in the model LSTM_UNIV_5 model.*

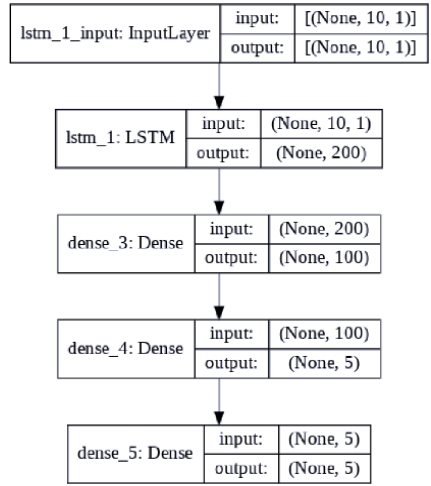

**Figure 6.**
*The schematic architecture of the model LSTM_UNIV_10.*

*dense_6*, are computed. The total number of parameters in the LSTM_UNIV_5 model is found to be 182, 235.

### 3.6 The LSTM_UNIV_10 model

LSTM_UNIV_10 model: This univariate model uses the last couple of weeks' open index input and yields the daily forecasted open values for the coming week. The same values of the parameters and hyperparameters of the model LSTM_UNIV_5 are used here. Only, the input data shape is different. The input data shape of this model is (10, 1). **Figure 6** presents the architecture of this model.

**Table 6** presents the computation of the number of parameters involved in the modelLSTM_UNIV_10. Since the number of parameters in the LSTM layers depends only on the number of features in the input data and the node-count in the LSTM layer, and not on the number of input records in one epoch, the model LSTM_UNIV_10 has an identical number of parameters in the LSTM layer as that of the model LSTM_UNIV_5. Since both the models have the same number of dense layers and have the same architecture for those layers, the total number of parameters for both the models are the same.

### 3.7 The LSTM_UNIV_ED_10 model

This LSTM model has an encoding and decoding capability and is based on the input of the *open* values of the stock price records of the last couple of weeks. The model consists of two LSTM blocks. One LSTM block performs the encoding operation, while the other does the decoding. The encoder LSTM block consists of 200 nodes (determined using the *grid-search* procedure). The input data shape to the encoder LSTM is (10, 1). The encoding layer yields a one-dimensional vector of size 200 – each value corresponding to the feature extracted by a node in the LSTM layer from the ten input values received from the input layer. Corresponding to each timestamp of the output sequence (there are five timestamps for the output sequence for the five forecasted *open* values), the input data features are extracted once. Hence, the data shape from the repeat vector layer's output is (5, 200). It signifies that in total 200 features are extracted from the input for each of the five timestamps corresponding to the model's output (i.e., forecasted) sequence. The second LSTM block decodes the encoded features using 200 nodes.

| Layer | x | y | $p_{prev}$ | $p_{curr}$ | n1 | n2 | #params |
|---|---|---|---|---|---|---|---|
| LSTM (lstm_2) | 200 | 1 | | | 40,400 | | 161600 |
| Dense (dense_4) | | | 200 | 100 | | 20,100 | 20100 |
| Dense (dense_5) | | | 100 | 5 | | 505 | 505 |
| Desne (dense_6) | | | 5 | 5 | | 30 | 30 |
| Total #parameters | | | | 182235 | | | |

**Table 6.**
*The number of parameters in the model LSTM_UNIV_10.*

The decoded result is passed on to a dense layer. The dense layer learns from the decoded values and predicts the future five values of the target variable (i.e., *open*) for the coming week through five nodes in the output layer. However, the forecasted values are not produced in a single timestamp. The forecasted values for the five days are made in five rounds. The round-wise forecasting is done using a *TimeDistributedWrapper* function that synchronizes the decoder LSTM block, the fully connected block, and the output layer in every round. The number of epochs and the batch sizes used in training the model are 70 and 16, respectively. **Figure 7** presents the structure and the data flow of the LSTM_UNIV_ED_10 model.

The computation of the number of parameters in the LSTM_UNIV_ED_10 model is shown in **Table 7**. The input layer and the repeat vector layer do not involve any learning, and hence these layers have no parameters. On the other hand, the two LSTM layers, *lstm_3* and *lstm_4*, and the two dense layers,

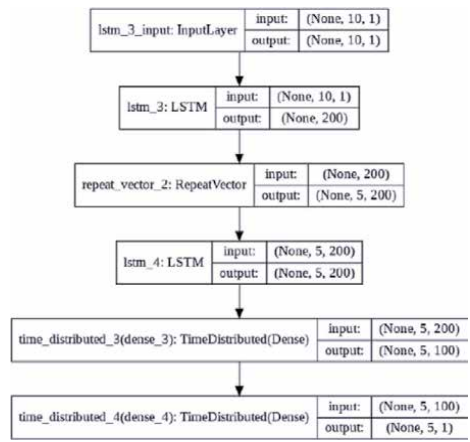

**Figure 7.**
*The schematic architecture of the model LSTM_UNIV_ED_10.*

| Layer | $x$ | $y$ | $p_{\text{prev}}$ | $p_{\text{curr}}$ | $n1$ | $n2$ | #params |
|---|---|---|---|---|---|---|---|
| LSTM (lstm_3) | 200 | 1 | | | 40,400 | | 161600 |
| LSTM (lstm_4) | 200 | 200 | | | | 80, 200 | 320800 |
| Dense (time_dist_dense_3) | | | 200 | 100 | | 20,100 | 20100 |
| Dense (time_dist_dense_4) | | | 100 | 1 | | 101 | 101 |
| **Total #parameters** | | | | | **502601** | | |

**Table 7.**
*The number of parameters in the model LSTM_UNIV_ED_10.*

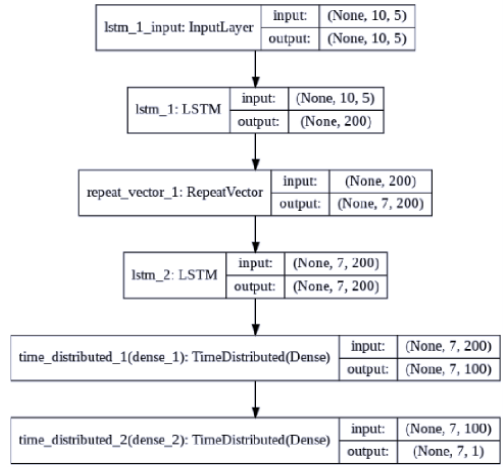

**Figure 8.**
*The schematic architecture of the model LSTM_MULTV_ED_10.*

*time_distributed_3*, and *time_distributed_4* involve learning. The number of parameters in the *lstm_3* layer is computed as: 4 * [(200 + 1) * 200 + 200] = 161, 600. The computation of the number of parameters in the *lstm_4* layer is as follows: 4 * [(200 + 200) * 200 + 200] = 320, 800. The computations of the dense layers' parameters are identical to those in the models discussed earlier. The total number of parameters in this model turns out to be 5,02,601.

### 3.8 The LSTM_MULTV_ED_10 model

This model is a multivariate version of LSTM_UNIV_ED_10. It uses the last couple of weeks' stock price records and includes all the five attributes, i.e., *open*, *high*, *low*, *close*, and *volume*. Hence, the input data shape for the model is (10, 5). We use a batch size of 16 while training the model over 20 epochs. **Figure 8** depicts the architecture of the multivariate encoder-decoder LSTM model.

**Table 8** shows the number of parameters in the LSTM_MULTV_ED_10 model. The computation of the parameters for this model is exactly similar to that for the model LSTM_UNIV_ED_50 expect for the first LSTM layer. The number of parameters in the first LSTM (i.e., the encoder) layer for this model will be different since the number of parameters is dependent on the count of the features in the input data. The computation of the parameter counts in the encoder LSTM layer, *lstm_1*, of the model is done as follows: 4 * [(200 + 5) * 200 + 200] = 164800. The total number of parameters for the model is found to be 505801.

| Layer | x | y | $p_{prev}$ | $p_{curr}$ | n1 | n2 | #params |
|---|---|---|---|---|---|---|---|
| LSTM (lstm_1) | 200 | 5 | | | 41200 | | 164800 |
| LSTM (lstm_2) | 200 | 200 | | | | 80, 200 | 320800 |
| Dense (time_dist_dense_1) | | | 200 | 100 | | 20,100 | 20100 |
| Dense (time_dist_dense_2) | | | 100 | 1 | | 101 | 101 |
| **Total #parameters** | | | | **505801** | | | |

**Table 8.**
*The number of parameters in the model LSTM_MULTV_ED_10.*

### 3.9 The LSTM_UNIV_CNN_10 model

This model is a modified version of the LSTM_UNIV_ED_N_10 model. A dedicated CNN block carries out the encoding operation. CNNs are poor in their ability to learn from sequential data. However, we exploit the power of a one-dimensional CNN in extracting important features from time-series data. After the feature extraction is done, the extracted features are provided as the input into an LSTM block. The LSTM block decodes the features and makes robust forecasting of the future values in the sequence. The CNN block consists of a couple of convolutional layers, each of which has a feature map size of 64 and a kernel size of 3. The input data shape is (10, 1) as the model uses univariate data of the target variable of the past couple of weeks. The output shape of the initial convolutional layer is (8, 64). The value of 8 is arrived at using the computation: (10–3 + 1), while 64 refers to the feature space dimension.

Similarly, the shape of the output of the next convolutional block is (6, 64). A max-pooling block follows, which contracts the feature-space dimension by 1/2. Hence, the output data shape of the max-pooling layer is (3, 64). The max-pooling layer's output is flattened into an array of single-dimension and size 3*64 = 192. The flattened vector is fed into the decoder LSTM block consisting of 200 nodes. The decoder architecture remains identical to the decoder block of the LSTM_UNIV_ED_10 model. We train the model over 20 epochs, with each epoch using 16 records. The structure and the data flow of the model are shown in **Figure 9**.

**Table 9** presents the computation of the number of parameters in the model LSTM_UNIV_CNN_10. The input layer, the max-pooling layer, the flatten operation, and the repeat vector layer do not involve any learning, and hence they have no parameters. The number of parameters in the first convolutional layer is computed as follows: (3 + 1) * 64 = 256. For the second convolutional layer, the number of parameters is computed as: (3 * 64 + 1) * 64 = 12352. The number of parameters for the LSTM layer is computed as follows: 4 * [(200 + 192) * 200 + 200] = 314400. In the case of the first dense layer, the number of parameters is computed as follows: (200 * 100 + 100) = 20100. Finally, the number of parameters in the second dense layer is computed as (100 * 1 + 1) = 101. The total number of parameters in the model is found out to be 347209.

### 3.10 The LSTM_UNIV_CONV_10 model

This model is a modification of the LSTM_UNIV_CNN_10 model. The encoder CNN's convolution operations and the decoding operations of the LSTM submodule are integrated for every round of the sequence in the output. This encoder-decoder model is also known as the Convolutional-LSTM model [58]. This integrated model reads sequential input data, performs convolution operations on the data without any explicit CNN block, and decodes the extracted features using a dedicated LSTM block. The Keras framework contains a class, *ConvLSTM2d*, which is capable of performing two-dimensional convolution operations [58]. The two-dimensional ConvLSTM class is tweaked to enable it to process univariate data of one dimension. The architecture of the model LSTM_UNIV_CONV_10 is represented in **Figure 10**.

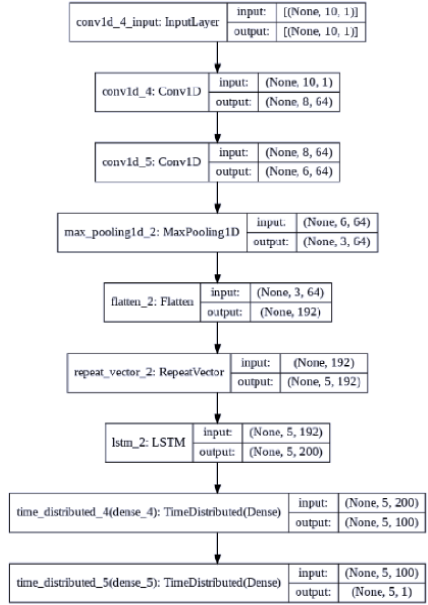

**Figure 9.**
*The schematic architecture of the model LSTM_UNIV_CNN_10.*

| Layer | $k$ | $d$ | $f$ | $x$ | $y$ | $p_{prev}$ | $p_{curr}$ | $n1$ | $n2$ | #param |
|---|---|---|---|---|---|---|---|---|---|---|
| Conv1D (conv1d_4) | 3 | 1 | 64 | | | | | 256 | | 256 |
| Conv1D (conv1d_5) | 3 | 64 | 64 | | | | | 12352 | | 12352 |
| LSTM (lstm_2) | | | | 200 | 192 | | | 78600 | | 314400 |
| Dense (time_dist_4) | | | | | | 200 | 100 | | 20,100 | 20100 |
| Dense (time_dist_5) | | | | | | 100 | 1 | | 101 | 101 |
| **Total #parameters** | | | | | | 347209 | | | | |

**Table 9.**
*The number of parameters in the model LSTM_UNIV_CNN_10.*

The computation of the number of parameters for the LSTM_UNIV_CONV_10 model is shown in **Table 10**. While the input layer, the flatten operation, and the repeat vector layer do not involve any learning, the other layers include trainable parameters. The number of parameters in the convolutional LSTM layer (i.e., *conv_1st_m2d*) is computed as follows: $4*x*[k \ (1 + x) + 1] = 4*64[3$ $(1 + 64) + 1] = 50176$. The number of parameters in the LSTM layer is computed as follows: $4*[(200 + 192)*200 + 100] = 314400$. The number of parameters in the *first time distributed dense layer* is computed as $(200*100 + 100) = 20100$. The computation for the *final dense layer* is as follows: $(100*1 + 1) = 101$. The total number of parameters involved in the model, LSTM_UNIV_CONV_10 is 38,4777.

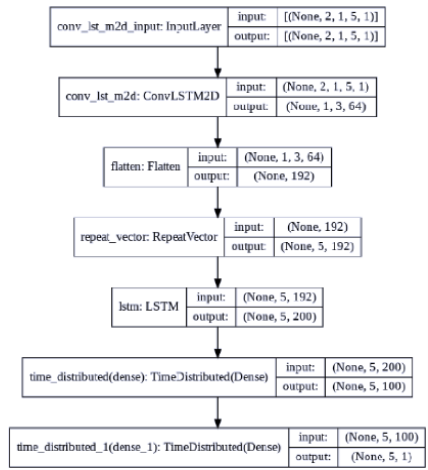

**Figure 10.**
*The schematic architecture of the model LSTM_UNIV_CONV_10.*

| Layer | k | d | f | x | y | $p_{prev}$ | $p_{curr}$ | n1 | n2 | #param |
|---|---|---|---|---|---|---|---|---|---|---|
| ConvLSTM2D(conv_1st_m2d) | 3 | 1 | 64 | 64 | 1 | | | 12544 | | 50176 |
| LSTM (lstm) | | | | 200 | 192 | | | 78600 | | 314400 |
| Dense (time_dist) | | | | | | 200 | 100 | | 20,100 | 20100 |
| Dense (time_dist_1) | | | | | | 100 | 1 | | 101 | 101 |
| **Total #parameters** | | | | | | **384777** | | | | |

**Table 10.**
*Computation of the no. of params in the model LSTM_UNIV_CONV_10.*

## 4. Performance results

We present the results on the performance of the ten deep learning models on the dataset we prepared. We also compare the performances of the models. For designing a robust evaluation framework, we execute every model over ten rounds. The average performance of the ten rounds is considered as the overall performance of the model. We use four metrics for evaluation: (i) average RMSE, (ii) the RMSE for different days (i.e., Monday to Friday) of a week, (iii) the time needed for execution of one round, and (iv) the ratio of the RMSE to the response variable's (i.e., *open* value's) mean value. The models are trained on 19500 historical stock records and then tested on 20250 records. The mean value of the response variable, *open*, of the test dataset is 475.70. All experiments are carried on a system with an Intel i7 CPU with a clock frequency in the range 2.60 GHz – 2.56 GHz and 16GB RAM. The time needed to complete one round of execution of each model is recorded in seconds. The models are built using the Python programming language version 3.7.4 and the frameworks TensorFlow 2.3.0 and Keras 2.4.5.

**Table 11** shows the results of the performance of the CNN_UNIV_5 model. The model takes, on average, 174.78 seconds to finish its one cycle of execution. For this model, the ratio of RMSE to the mean *open* values is 0.007288. The ratio of the RMSE to the average of the actual *open* values for day1 through day5 are 0.0062, 0.0066, 0.0073, 0.0078, and 0.0083, respectively. Here, day1 refers to Monday, and day5 is Friday. In all subsequent Tables, we will use the same notations. The RMSE values of the model CNN_UNIV_N_5 plotted on different days in a week are depicted in **Figure 11** as per record no 2 in **Table 11**.

| No. | Agg RMSE | Day1 | Day2 | Day3 | Day4 | Day5 | Time (sec) |
|---|---|---|---|---|---|---|---|
| 1 | 4.058 | 4.00 | 3.40 | 3.90 | 4.40 | 4.50 | 173.95 |
| 2 | 3.782 | 3.10 | 3.30 | 3.80 | 4.10 | 4.40 | 176.92 |
| 3 | 3.378 | 2.80 | 3.00 | 3.40 | 3.60 | 3.90 | 172.21 |
| 4 | 3.296 | 2.60 | 3.00 | 3.30 | 3.60 | 3.90 | 173.11 |
| 5 | 3.227 | 2.60 | 3.00 | 3.30 | 3.50 | 3.70 | 174.72 |
| 6 | 3.253 | 2.60 | 3.00 | 3.30 | 3.50 | 3.70 | 183.77 |
| 7 | 3.801 | 3.60 | 3.60 | 3.80 | 3.80 | 4.10 | 172.29 |
| 8 | 3.225 | 2.60 | 2.90 | 3.30 | 3.50 | 3.70 | 171.92 |
| 9 | 3.306 | 2.80 | 3.00 | 3.30 | 3.50 | 3.70 | 174.92 |
| 10 | 3.344 | 2.70 | 3.10 | 3.40 | 3.60 | 3.80 | 174.01 |
| **Mean** | **3.467** | 2.94 | 3.13 | 3.48 | 3.71 | 3.94 | **174.78** |
| **RMSE/Mean** | **0.007288** | 0.0062 | 0.0066 | 0.0073 | 0.0078 | 0.0083 | |

**Table 11.**
*The RMSE and the execution time of the CNN_UNIV_5 model.*

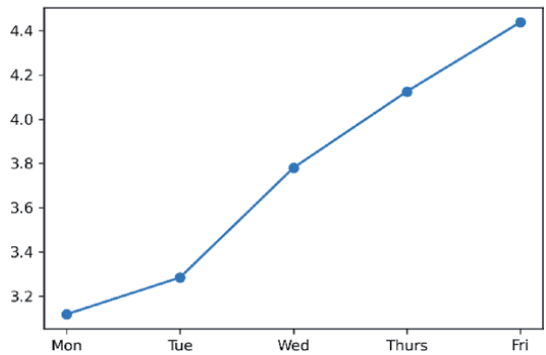

**Figure 11.**
*RMSE vs. day plot of CNN_UNIV_5 (depicted by tuple#2 in **Table 11**).*

**Table 12** depicts the performance results of the model CNN_UNIV_10. The model needs 185.01 seconds on average for one round. The ratio of the RMSE to the average of the *open* values for the model is 0.006967. The ratios of the RMSE to the average *open* values for day1 through day5 for the model are 0.0056, 0.0067, 0.0070, 0.0075, and 0.0080, respectively. **Figure 12** presents the RMSE values for the results of round 7 in **Table 12**.

**Table 13** depicts the performance results of the model CNN_MULTV_10. One round of execution of the model requires 202.78 seconds. The model yields a value of 0.009420 for the ratio of the RMSE to the average of the *open* values. The ratios of the RMSE values to the mean of the *open* values for day1 through day5 of a week are 0.0085, 0.0089, 0.0095, 0.0100, and 0.0101, respectively. The RMSE values of the model CNN_MULTV_N_10 plotted on different days in a week are depicted in **Figure 13** based on record number 6 of **Table 13**.

**Table 14** depicts the results of the model CNN_MULTH_10. The model needs, on average, 215.07 seconds to execute its one round. The ratio of the RMSE to the average of the *open* values is 0.008100. The ratios of the RMSE to the average *open* value for day1 to day5 are 0.0076, 0.0075, 0.0082, 0.0084, and 0.0088, respectively. The pattern of variations exhibited by the model daily RMSE is shown in **Figure 14** as per record no 4 in **Table 14**.

| No. | Agg RMSE | Day1 | Day2 | Day3 | Day4 | Day5 | Time (sec) |
|---|---|---|---|---|---|---|---|
| 1 | 3.165 | 2.50 | 3.20 | 3.10 | 3.50 | 3.50 | 177.86 |
| 2 | 3.813 | 3.30 | 3.90 | 3.30 | 3.60 | 4.80 | 202.25 |
| 3 | 3.230 | 2.60 | 2.90 | 3.30 | 3.50 | 3.80 | 183.45 |
| 4 | 3.209 | 2.50 | 3.10 | 3.40 | 3.40 | 3.60 | 188.35 |
| 5 | 3.176 | 2.80 | 3.00 | 3.10 | 3.40 | 3.60 | 180.30 |
| 6 | 3.233 | 2.60 | 3.00 | 3.30 | 3.50 | 3.70 | 181.20 |
| 7 | 3.312 | 2.70 | 3.20 | 3.20 | 3.50 | 3.80 | 188.81 |
| 8 | 3.082 | 2.20 | 2.80 | 3.30 | 3.30 | 3.50 | 180.89 |
| 9 | 3.772 | 2.80 | 3.70 | 3.90 | 4.30 | 4.10 | 186.23 |
| 10 | 3.150 | 2.40 | 2.90 | 3.20 | 3.50 | 3.60 | 180.78 |
| **Mean** | **3.3142** | 2.64 | 3.17 | 3.31 | 3.55 | 3.80 | **185.01** |
| **RMSE/Mean** | **0.006967** | 0.0056 | 0.0067 | 0.0070 | 0.0075 | 0.0080 | |

**Table 12.**
*The RMSE and the execution time of the CNN_UNIV_10 model.*

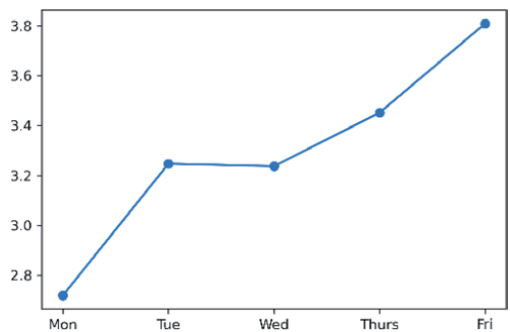

**Figure 12.**
*RMSE vs. day plot of CNN_UNIV_10 (depicted by tuple#7 in **Table 12**).*

The results of the *LSTM_UNIV_5* model are depicted in **Table 15**. The average time needed to complete one round of the model is 371.62 seconds. The ratio of the RMSE and the average value of the target variable is 0.007770. The RMSE values for day1 to day5 are 0.0067, 0.0071, 0.0074, 0.0081, and 0.0086, respectively. The pattern of variation of the daily RMSE is as per record no 9 in **Table 15** is depicted in **Figure 15**.

**Table 16** exhibits the results of the model *LSTM_UNIV_10*. The model yields a value of 0.007380 for the ratio of its RMSE to the average *open* values, while one round of its execution needs 554.47 seconds. The RMSE values for day1 to day5 are 0.0061, 0.0070, 0.0074, 0079, and 0.0083 respectively. The RMSE pattern of the model as per record no 10 in **Table 16** is exhibited in **Figure 16**.

**Table 17** shows that the model LSTM_UNIV_ED_10 needs, on average, 307.27 seconds to execute its one round. The average value of the ratio of the RMSE to the average value of the target variable (i.e., the *open* values) for the model is 0.008350. The daily ratio values for day1 to day 5 of the model are, 0.0067, 0.0078, 0.0085, 0.0090, and 0.0095, respectively. **Figure 17** exhibits the pattern of variation of the daily RMSE as per record no 9 in **Table 17**.

| No. | Agg RMSE | Day1 | Day2 | Day3 | Day4 | Day5 | Time (sec) |
|---|---|---|---|---|---|---|---|
| 1 | 4.525 | 4.00 | 4.30 | 4.50 | 4.70 | 5.00 | 206.92 |
| 2 | 3.606 | 3.10 | 3.30 | 3.70 | 3.80 | 4.00 | 202.61 |
| 3 | 4.830 | 4.60 | 4.70 | 4.70 | 5.10 | 5.00 | 202.87 |
| 4 | 4.938 | 4.40 | 4.80 | 4.70 | 5.30 | 5.40 | 201.49 |
| 5 | 4.193 | 3.50 | 4.00 | 4.10 | 4.60 | 4.60 | 214.66 |
| 6 | 5.101 | 4.70 | 4.90 | 5.20 | 5.30 | 5.30 | 190.73 |
| 7 | 4.751 | 4.40 | 4.50 | 4.80 | 5.00 | 5.00 | 201.73 |
| 8 | 3.927 | 3.20 | 3.70 | 4.00 | 4.30 | 4.40 | 200.04 |
| 9 | 4.267 | 3.90 | 3.80 | 4.50 | 4.60 | 4.40 | 199.09 |
| 10 | 4.661 | 4.40 | 4.50 | 4.60 | 4.90 | 4.90 | 207.62 |
| **Mean** | **4.4799** | 4.02 | 4.25 | 4.53 | 4.76 | 4.80 | **202.78** |
| **RMSE/Mean** | **0.009420** | 0.0085 | 0.0089 | 0.0095 | 0.0100 | 0.0101 | |

**Table 13.**
*The RMSE and the execution time of the CNN_MULTV_10 model.*

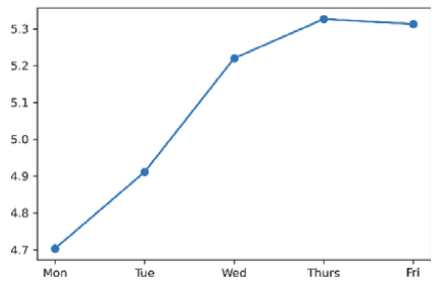

**Figure 13.**
*RMSE vs. day plot of CNN_MULTV_10 (based on tuple#6 in **Table 13**).*

| No. | Agg RMSE | Day1 | Day2 | Day3 | Day4 | Day5 | Time (sec) |
|---|---|---|---|---|---|---|---|
| 1 | 3.338 | 2.70 | 2.80 | 3.30 | 3.70 | 4.00 | 224.63 |
| 2 | 3.264 | 2.80 | 3.10 | 3.30 | 3.50 | 3.70 | 216.44 |
| 3 | 3.015 | 2.30 | 2.70 | 3.10 | 3.30 | 3.50 | 218.14 |
| 4 | 3.692 | 3.20 | 3.40 | 4.00 | 3.80 | 4.00 | 220.01 |
| 5 | 3.444 | 2.80 | 3.20 | 3.40 | 3.80 | 3.90 | 212.54 |
| 6 | 4.019 | 4.50 | 3.70 | 3.70 | 4.20 | 3.90 | 210.95 |
| 7 | 6.988 | 6.40 | 7.40 | 7.20 | 6.80 | 7.10 | 210.24 |
| 8 | 3.133 | 2.50 | 2.80 | 3.20 | 3.40 | 3.60 | 214.48 |
| 9 | 3.278 | 2.40 | 3.10 | 3.70 | 3.40 | 3.60 | 211.53 |
| 10 | 4.469 | 5.90 | 3.60 | 4.00 | 4.10 | 4.40 | 211.78 |
| **Mean** | **3.864** | 3.55 | 3.58 | 3.89 | 4.00 | 4.17 | **215.07** |
| **RMSE/Mean** | **0.008100** | 0.0076 | 0.0075 | 0.0082 | 0.0084 | 0.0088 | |

**Table 14.**
*The RMSE and the execution time of the CNN_MULTH_10 model.*

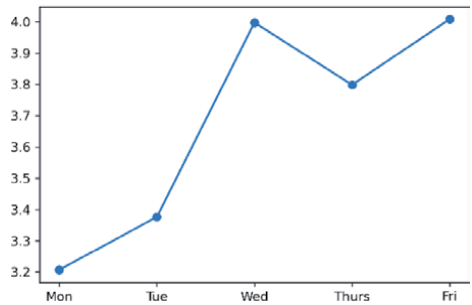

**Figure 14.**
*RMSE vs. day plot of CNN_MULTH_10 (based on tuple#4 in **Table 14**).*

| No. | Agg RMSE | Day1 | Day2 | Day3 | Day4 | Day5 | Time (sec) |
|---|---|---|---|---|---|---|---|
| 1 | 3.125 | 2.40 | 2.90 | 3.00 | 3.50 | 3.70 | 372.28 |
| 2 | 3.376 | 3.00 | 2.90 | 3.40 | 3.90 | 3.70 | 371.73 |
| 3 | 2.979 | 2.10 | 2.60 | 3.00 | 3.30 | 3.70 | 368.72 |
| 4 | 3.390 | 3.20 | 3.40 | 3.30 | 3.60 | 3.50 | 368.58 |
| 5 | 4.387 | 4.20 | 4.60 | 4.10 | 4.40 | 4.60 | 379.10 |
| 6 | 5.173 | 4.40 | 5.10 | 4.60 | 5.20 | 6.30 | 373.84 |
| 7 | 3.434 | 4.30 | 2.60 | 2.90 | 3.70 | 3.50 | 368.91 |
| 8 | 3.979 | 3.70 | 3.10 | 4.60 | 4.30 | 4.10 | 371.02 |
| 9 | 2.892 | 1.90 | 2.50 | 2.90 | 3.30 | 3.50 | 371.95 |
| 10 | 3.683 | 2.70 | 4.00 | 3.30 | 3.50 | 4.60 | 370.07 |
| **Mean** | **3.6418** | 3.19 | 3.37 | 3.51 | 3.87 | 4.12 | **371.62** |
| **RMSE/Mean** | **0.007770** | 0.0067 | 0.0071 | 0.0074 | 0.0081 | 0.0086 | |

**Table 15.**
*The RMSE and the execution time of the LSTM_UNIV_5 model.*

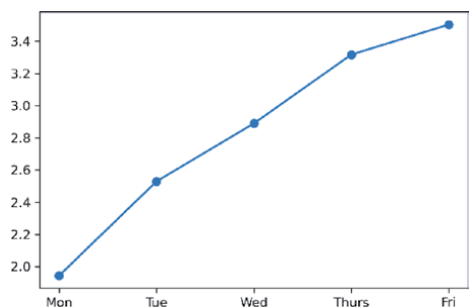

**Figure 15.**
*RMSE vs. day plot of LSTM_UNIV_5 (depicted by tuple#9 in **Table 15**).*

Table 18 shows that the model LSTM_MULTV_ED_10, on average, requires 634.34 seconds to complete the execution of its one round. For this model, the ratio of the RMSE to the average value of the target variable (i.e., the *open* values) is 0.010294. The ratios of the daily RMSE to the mean value of *open* for day1 to day5 are, respectively, 0.0094, 0.0099, 0.0102, 0.0107, and 0.0111. **Figure 18** shows the pattern of the daily RMSE values of the model as per record no 10 in **Table 18**.

| No. | Agg RMSE | Day1 | Day2 | Day3 | Day4 | Day5 | Time (sec) |
|---|---|---|---|---|---|---|---|
| 1 | 3.005 | 2.40 | 2.40 | 2.80 | 3.70 | 3.50 | 547.22 |
| 2 | 3.859 | 3.50 | 3.30 | 3.80 | 3.90 | 4.70 | 554.03 |
| 3 | 4.601 | 4.50 | 4.50 | 4.60 | 4.80 | 4.60 | 550.24 |
| 4 | 3.342 | 2.70 | 4.00 | 3.10 | 3.40 | 3.50 | 555.50 |
| 5 | 4.714 | 4.80 | 4.40 | 4.70 | 4.60 | 5.10 | 563.44 |
| 6 | 3.336 | 2.50 | 3.20 | 3.30 | 3.60 | 3.90 | 553.83 |
| 7 | 3.711 | 3.10 | 4.00 | 4.00 | 3.60 | 3.90 | 559.31 |
| 8 | 2.795 | 1.90 | 2.40 | 2.80 | 3.20 | 3.40 | 552.50 |
| 9 | 3.012 | 1.80 | 2.80 | 2.90 | 3.60 | 3.50 | 551.20 |
| 10 | 2.751 | 1.70 | 2.30 | 3.00 | 3.00 | 3.30 | 557.39 |
| **Mean** | **3.5126** | 2.89 | 3.33 | 3.50 | 3.74 | 3.94 | **554.47** |
| **RMSE/Mean** | **0.007380** | 0.0061 | 0.0070 | 0.0074 | 0.0079 | 0.0083 | |

**Table 16.**
*The RMSE and the execution time of the LSTM_UNIV_10 model.*

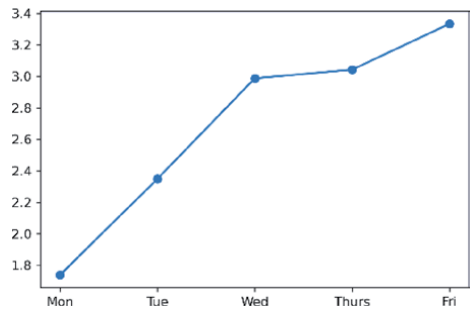

**Figure 16.**
*RMSE vs. day plot of LSTM_UNIV_10 (depicted by tuple#10 in **Table 16**).*

| No. | Agg RMSE | Day1 | Day2 | Day3 | Day4 | Day5 | Time (sec) |
|---|---|---|---|---|---|---|---|
| 1 | 2.975 | 2.00 | 2.70 | 3.00 | 3.40 | 3.60 | 310.28 |
| 2 | 4.856 | 4.10 | 4.60 | 5.00 | 5.20 | 5.30 | 306.22 |
| 3 | 5.500 | 4.30 | 5.20 | 5.50 | 6.00 | 6.40 | 306.08 |
| 4 | 3.656 | 3.20 | 3.40 | 3.70 | 3.90 | 4.10 | 305.64 |
| 5 | 2.859 | 1.90 | 2.60 | 2.90 | 3.20 | 3.40 | 306.03 |
| 6 | 3.887 | 3.30 | 3.60 | 3.90 | 4.20 | 4.40 | 305.34 |
| 7 | 4.007 | 3.60 | 3.70 | 4.00 | 4.10 | 4.50 | 304.69 |
| 8 | 3.489 | 2.70 | 3.20 | 3.60 | 3.80 | 3.90 | 305.26 |
| 9 | 2.944 | 2.10 | 2.80 | 3.00 | 3.20 | 3.50 | 314.37 |
| 10 | 5.497 | 4.70 | 5.10 | 5.60 | 6.00 | 5.90 | 308.78 |
| **Mean** | **3.971** | 3.19 | 3.69 | 4.02 | 4.30 | 4.50 | **307.27** |
| **RMSE/Mean** | **0.008350** | 0.0067 | 0.0078 | 0.0085 | 0.0090 | 0.0095 | |

**Table 17.**
*The RMSE and the execution time of the LSTM_UNIV_ED_10 model.*

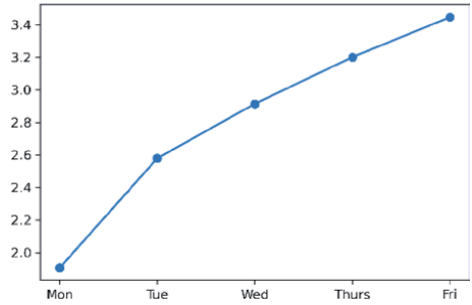

**Figure 17.**
*RMSE vs. day plot of LSTM_UNIV_ED_10 (as per tuple#5 in **Table 17**).*

| No. | Agg RMSE | Day1 | Day2 | Day3 | Day4 | Day5 | Time (sec) |
|-----|----------|------|------|------|------|------|------------|
| 1 | 5.858 | 5.50 | 5.70 | 5.90 | 6.00 | 6.20 | 631.53 |
| 2 | 4.062 | 3.60 | 3.90 | 4.00 | 4.20 | 4.50 | 617.62 |
| 3 | 6.623 | 6.20 | 6.50 | 6.60 | 6.80 | 6.90 | 640.09 |
| 4 | 3.661 | 3.20 | 3.30 | 3.60 | 3.90 | 4.10 | 624.22 |
| 5 | 5.879 | 5.80 | 5.90 | 5.70 | 6.00 | 6.10 | 632.34 |
| 6 | 4.808 | 4.20 | 4.60 | 4.80 | 5.10 | 5.20 | 644.48 |
| 7 | 4.657 | 4.10 | 4.50 | 4.70 | 4.90 | 5.10 | 631.72 |
| 8 | 3.866 | 3.30 | 3.60 | 3.90 | 4.10 | 4.30 | 633.28 |
| 9 | 3.910 | 3.30 | 3.70 | 3.90 | 4.20 | 4.40 | 647.29 |
| 10 | 5.644 | 5.30 | 5.50 | 5.60 | 5.90 | 6.00 | 640.86 |
| **Mean** | **4.897** | 4.50 | 4.72 | 4.87 | 5.11 | 5.28 | **634.34** |
| **RMSE/Mean** | **0.010294** | 0.0094 | 0.0099 | 0.0102 | 0.0107 | 0.0111 | |

**Table 18.**
*The RMSE and the execution time of the LSTM_MULTV_ED_10 model.*

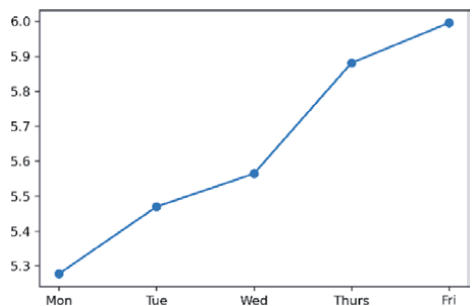

**Figure 18.**
*RMSE vs. day plot of LSTM_MULTV_ED_10 (as per tuple#10 in **Table 18**).*

**Table 19** depicts that the model LSTM_UNIV_CNN_N_10 requires, on average, 222.48 seconds to finish one round. For this model, the ratio of the RMSE to the average value of the target variable (i.e., the *open* values) is found to be 0.007916. The daily RMSE values for day1 to day5 are, 0.0065, 0.0074, 0.0080, 0.0085, and 0.0089 respectively. **Figure 19** depicts the pattern of variation of the daily RMSE values for this model as per record no 3 in **Table 19**.

| No. | Agg RMSE | Day1 | Day2 | Day3 | Day4 | Day5 | Time (sec) |
|---|---|---|---|---|---|---|---|
| 1 | 3.832 | 3.30 | 3.50 | 3.90 | 4.10 | 4.30 | 221.18 |
| 2 | 3.256 | 2.50 | 3.00 | 3.30 | 3.60 | 3.80 | 219.74 |
| 3 | 4.308 | 3.80 | 4.00 | 4.40 | 4.60 | 4.60 | 222.59 |
| 4 | 4.081 | 3.30 | 4.00 | 4.10 | 4.30 | 4.50 | 227.95 |
| 5 | 3.325 | 2.60 | 3.00 | 3.30 | 3.60 | 3.90 | 224.46 |
| 6 | 3.870 | 3.20 | 3.70 | 3.90 | 4.10 | 4.10 | 223.40 |
| 7 | 3.688 | 3.10 | 3.40 | 3.80 | 4.00 | 4.10 | 222.89 |
| 8 | 3.851 | 3.20 | 3.60 | 3.80 | 4.20 | 4.40 | 221.87 |
| 9 | 3.710 | 2.60 | 3.40 | 4.00 | 4.00 | 4.40 | 219.74 |
| 10 | 3.736 | 3.30 | 3.70 | 3.70 | 3.90 | 4.10 | 220.96 |
| **Mean** | **3.766** | 3.09 | 3.53 | 3.82 | 4.04 | 4.22 | **222.48** |
| **RMSE/Mean** | **0.007916** | 0.0065 | 0.0074 | 0.0080 | 0.0085 | 0.0089 | |

**Table 19.**
*The RMSE and the execution time of the LSTM_UNIV_CNN_10 model.*

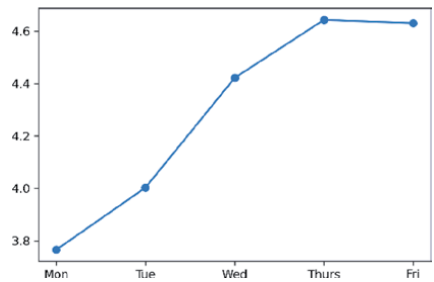

**Figure 19.**
*RMSE vs. day plot of LSTM_UNIV_CNN_10 (as per tuple#3 in* **Table 19**).

| No. | Agg RMSE | Day1 | Day2 | Day3 | Day4 | Day5 | Time (sec) |
|---|---|---|---|---|---|---|---|
| 1 | 3.971 | 3.00 | 3.60 | 4.00 | 4.40 | 4.60 | 263.84 |
| 2 | 3.103 | 2.40 | 2.80 | 3.20 | 3.40 | 3.60 | 262.06 |
| 3 | 3.236 | 2.30 | 2.90 | 3.30 | 3.60 | 3.80 | 266.47 |
| 4 | 4.347 | 3.10 | 4.00 | 4.60 | 4.70 | 5.00 | 257.43 |
| 5 | 2.860 | 2.20 | 2.50 | 2.80 | 3.20 | 3.40 | 260.05 |
| 6 | 3.525 | 2.50 | 3.60 | 3.50 | 3.80 | 4.00 | 282.27 |
| 7 | 3.163 | 2.30 | 2.80 | 3.20 | 3.50 | 3.80 | 265.26 |
| 8 | 2.870 | 2.00 | 2.60 | 2.90 | 3.20 | 3.50 | 272.18 |
| 9 | 3.504 | 2.20 | 3.10 | 3.70 | 3.70 | 4.40 | 265.46 |
| 10 | 5.053 | 4.70 | 4.40 | 5.20 | 5.30 | 5.60 | 264.66 |
| **Mean** | **3.563** | 2.67 | 3.23 | 3.64 | 3.88 | 4.17 | **265.97** |
| **RMSE/Mean** | **0.007490** | 0.0056 | 0.0068 | 0.0077 | 0.0082 | 0.0088 | |

**Table 20.**
*The RMSE and the execution time of the LSTM_UNIV_CONV_10 model.*

The results of the model LSTM_UNIV_CONV_N_10 are presented in **Table 20**. The model completes its one round, on average, in 265.97 seconds. The ratio of the RMSE to the average of the *open* values is 0.007490. The daily RMSE for day1 to day5 are 0.0056, 0.0068, 0.0077, 0.0082, and 0.0088, respectively. **Figure 20** shows the patterns of daily RMSE values for this model as per record no 8 in **Table 20**.

**Table 21** summarizes the performance of the ten models proposed in this chapter. We evaluate the models on two metrics and then rank the models on the basis of each metric. The two metrics used for the model evaluation are: (i) an accuracy matric computed as the ratio of the RMSE to the mean value of the target variable (i.e., *open* values), and (ii) a speed metric as measured by the time (in seconds) required for execution of one round of the model. The number of parameters in each model is also presented. It is noted that the CNN_UNIV_5 model is ranked 1 on its execution speed, while it occupies rank 2 on the accuracy parameter. The CNN_UNIV_10 model, on the other hand, is ranked 2 in terms of its speed of execution, while it is the most accurate model. It is also interesting to note that all the CNN models are faster than their LSTM counterparts. However, there is no appreciable difference in their accuracies except for the multivariate encoder-decoder LSTM model, LSTM_MULTV_ED_10.

Another interesting observation is that the multivariate models are found to be inferior to the corresponding univariate models on the basis of the accuracy metric.

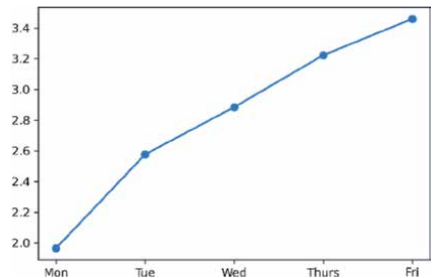

**Figure 20.**
*RMSE vs. day plot of LSTM_UNIV_CONV_10 (as per tuple#8 in Table 20).*

| No. | Model | #param | RMSE/Mean | Rank | Exec. Time (s) | Rank |
|-----|-------|--------|-----------|------|----------------|------|
| 1 | CNN_UNIV_5 | 289 | 0.007288 | 2 | 174.78 | 1 |
| 2 | CNN_UNIV_10 | 769 | 0.006967 | 1 | 180.01 | 2 |
| 3 | CNN_MULTV_10 | 7373 | 0.009420 | 9 | 202.78 | 3 |
| 4 | CNN_MULTH_10 | 132965 | 0.008100 | 7 | 215.07 | 4 |
| 5 | LSTM_UNIV_5 | 182235 | 0.007770 | 5 | 371.62 | 8 |
| 6 | LSTM_UNIV_10 | 182235 | 0.007380 | 3 | 554.47 | 9 |
| 7 | LSTM_UNIV_ED_10 | 502601 | 0.008350 | 8 | 307.27 | 7 |
| 8 | LSTM_MULTV_ED_10 | 505801 | 0.010294 | 10 | 634.34 | 10 |
| 9 | LSTM_UNIV_CNN_10 | 347209 | 0.007916 | 6 | 222.48 | 5 |
| 10 | LSTM_UNIV_CONV_10 | 384777 | 0.007490 | 4 | 265.97 | 6 |

**Table 21.**
*Comparative analysis of the accuracy and execution speed of the models.*

The multivariate models, CNN_MULTV_10 and LSTM_MULTV_ED_10, are ranked 9 and 10, respectively, under the accuracy metric.

Finally, it is observed that the number of parameters in a model has an effect on its execution time, barring some notable exceptions. For the four CNN models, it is noted that with the increase in the number of parameters, there is a monotone increase in the execution time of the models. For the LSTM models, even though the models, LSTM_UNIV_CNN_10, LSTM_UNIV_CONV_10, and LSTM_UNIV_ED_10, have higher number of parameters than the vanilla LSTM models (i.e., LSTM_UNIV_5 and LSTM_UNIV_10), they are faster in execution. Evidently, the univariate encoder-decoder LSTM models are faster even when they involve a higher number of parameters than the vanilla LSTM models.

## 5. Conclusion

Prediction of future stock prices and price movement patterns is a challenging task if the stock price time series has a large amount of volatility. In this chapter, we presented ten deep learning-based regression models for robust and precise prediction of stock prices. Among the ten models, four of them are built on variants of CNN architectures, while the remaining six are constructed using different LSTM architectures. The historical stock price records are collected using the Metastock tool over a span of two years at five minutes intervals. The models are trained using the records of the first year, and then they are tested on the remaining records. The testing is carried out using an approach known as walk-forward validation, in which, based on the last one- or two-weeks historical stock prices, the predictions of stock prices for the five days of the next week are made. The overall RMSE and the RMSE for each day in a week are computed to evaluate the prediction accuracy of the models. The time needed to complete one round of execution of each model is also noted in order to measure the speed of execution of the models. The results revealed some very interesting observations. First, it is found that while the CNN models are faster, in general, the accuracies of both CNN and LSTM models are comparable. Second, the univariate models are faster and more accurate than their multivariate counterparts. And finally, the number of variables in a model has a significant effect on its speed of execution except for the univariate encoder-decoder LSTM models. As a future scope of work, we will design optimized models based on *generative adversarial networks* (GANs) for exploring the possibility of further improving the performance of the models.

## Author details

Jaydip Sen[1]* and Sidra Mehtab[2]

1 Department of Data Science, Praxis Business School, Kolkata, India

2 School of Computing and Analytics, NSHM Knowledge Campus, Kolkata, India

*Address all correspondence to: jaydip.sen@acm.org

**IntechOpen**

# Articulated Human Pose Estimation Using Greedy Approach

*Pooja Kherwa, Sonali Singh, Saheel Ahmed, Pranay Berry and Sahil Khurana*

## Abstract

The goal of this Chapter is to introduce an efficient and standard approach for human pose estimation. This approach is based on a bottom up parsing technique which uses a non-parametric representation known as Greedy Part Association Vector (GPAVs), generates features for localizing anatomical key points for individuals. Taking leaf out of existing state of the art algorithm, this proposed algorithm aims to estimate human pose in real time and optimize its results. This approach simultaneously detects the key points on human body and associates them by learning the global context. However, In order to operate this in real environment where noise is prevalent, systematic sensors error and temporarily crowded public could pose a challenge, an efficient and robust recognition would be crucial. The proposed architecture involves a greedy bottom up parsing that maintains high accuracy while achieving real time performance irrespective of the number of people in the image.

**Keywords:** Neural networks, Pose- estimation, Greedy Search, Neural Network, Heat-maps

## 1. Introduction

Human pose estimation is a complex field of study in artificial intelligence, which requires a depth knowledge of computer vision, calculus, graph theory and biology. Initially this work start by introducing an image to a computer through camera and detect humans in the image known as object detection, as one of computer vision problem. In real world detecting an object from an image [1] and estimating its posture [2, 3] is two different aspects of objects. The latter is a very challenging and complex task. Images are filled with occluded objects, humans in close proximity, occlusions or spatial interference makes the task even more strenuous. One way of solving this problem is to use single person detector for estimation known as top down parsing [4–9]. This approach suffers from preconceived assumptions and lacks robustness. The approach is biased towards early decisions which makes it hard to recover if failed. Besides this, the computational time complexity is commensurate with the number of people in the image which makes it not an ideal approach for practical purpose. On a contrary the bottom up approach seems to perform well as compare to its counterpart. However earlier bottom up versions could not able to reduce the computational complexity as it unable to sustain the benefits of being consistent. For instance, the pioneering work E. Insafutdinov et al. Proposed a bottom up approach that simultaneously detects joints and label them as part candidates [10]. Later it associates them to individual person. Even solving the combinatorial optimization problem over a complete graph is itself NP hard. Another approach built on with stronger joint detectors based on ResNet [11] and provides ranking based on images, significantly improved its runtime but still performs in the order of minutes per image. The approach also requires a separate logistic regression for precise regression. After studying sufficient approaches and their shortcomings in the literature of image processing and object detection, this chapter introduces a efficient approach for human pose estimation.

## 1.1 Contribution of the work

Optimizing the current state of the art results and introducing a new approach to solving this problem is the highlight of this chapter. In this chapter, we presented a bottom up parsing technique which uses a non-parametric representation, features for localizing anatomical key points for individuals. We further introduced a multistage architecture with two parallel branches one of the branches estimates the body joints via hotspots while the other branch captures the orientations of the joints through vectors This proposed approach is based on bottom up parsing, localizes the anatomical key points and associates them using greedy parsing technique known as greedy part association vectors. These 2D vectors aims to provide not only the encoded translator position but also the respective directional orientations of body parts. This approach also able to decouple the dependency of number of persons with running time complexity. Our approach has resulted in competitive performance on some of the best public benchmarks. The model maintains its accuracy while providing real time performance.

This chapter comprises of 6 sections: Section 2 discussed related work, in Section 3, proposed methodology is explained, in details with algorithms, in Section 4 results are discussed, and finally the chapter is concluded with future work in Section 5.

## 2. Related work

The research trend that was primarily focused on detection of objects, visual object tracking and human body part detection, has advanced to pose estimation recently. Various visual tracking architectures have been proposed such as those based on convolutional neural networks and particle filtering and colored area tracking using mean shift tracking through temporal image sequence [12]. A survey of approaches for intruder detection systems in a camera monitored frame for surveillance was explained by C. Savitha and D. Ramesh [13]. A. Shahbaz and K. Jo also proposed a human verifier which is a SVM classifier based on histogram of oriented gradients along with an algorithm for change detection based on Gaussian mixture model [14]. But still there was a need of more precise detection algorithm that would accurately predict minor features as well. Human and Object detection evolved to detection of human body parts. L. Kong, X. Yuan and A.M.Maharajan introduced framework for automated joint detection using depth frames [15]. A cascade of Deep neural networks was used for Pose Estimation formulated as a joint regression problem and cast in DNN [16]. A full image and 7-layered generic convolutional DNN is taken as input to regress the location of each body joint. In [17], long-term temporal coherence was propagated to each stage of the overall video and data of joint position of initial posture was generated. A multi-feature, three-stage deep CNN was adopted to maintain temporal consistency of video by halfway temporal evaluation method and structured space learning. Speeded up Robust features (SURF) and Scale

Invariant Feature Transform (SIFT) was proposed by A. Agarwal, D. Samaiya and K. K. Gupta to deal with blur and illumination changes for different background conditions [18]. Paper [19] aims to improve human ergonomics using Wireless vibrotactile displays in the execution of repetitive or heavy industrial tasks. Different approach was presented to detect human pose. Coarse-Fine Network for Key point Localization (CFN)[20], G-RMI [21] and Regional Multi-person Pose Estimation (RMPE)[22] techniques have been used to implement top-down approach of pose detection (i.e. the person is identified first and then the body parts). An alternate bottom-up approach was proposed by Z. Cao, T. Simon, S. Wei and Y. Sheikh based on Partial Affinity Fields to efficiently detect the 2D pose [23]. X. Chen and G. Yang also presented a generic multi-person bottom-up approach for pose estimation formulated as a set of bipartite graph matching by introducing limb detection heatmaps. These heatmaps represent association of body joint pairs, that are simultaneously learned with joint detection [24]. L. Ke, H. Qi, M. Chang and S. Lyu proposed a deep conv-deconv modules-based pose estimation method via keypoint association using a regression network [25]. K. Akila and S. Chitrakala introduced a highly discriminating HOI descriptor to recognize human action in a video. The focus is to discriminate identical spatio-temporal relations actions by human-object interaction analysis and with similar motion pattern [26] Y. Yang and D. Ramanan proposed methods for pose detection and estimation for static images based on deformable part models with augmentation of standard pictorial structure model by co-occurrence relations between spatial relations of part location and part mixtures [27]. A Three-dimensional (3D) human pose estimation methods are explored and reviewed in a paper, it nvolves estimating the articulated 3D joint locations of a human body from an image or video [28]. One more study includes a 2-D technique which localize dense landmark on the entire body like face, hands and even on skin [29].

## 3. Proposed approach for human pose detection

### 3.1 Methodology

**Figure 1** depicts the methodology of our proposed approach, our approach works as black box which receive an image of a fixed size and produces a 2D anatomical key point of every person in the image. After performing the needed preprocessing, the image is passed through a feed forward convolutional neural network. The architecture has two separate branches that runs simultaneously

   i. On one branch it predicts an approximations represented by a set of hotspots H for each body joint locations while the

   ii. Other branch predicts a set of 2D vectors representing joints associations P for each pair of different joints. Each set H is a collection of $\{H1, H2, H3, ... Hj\}$ j hotspots one for each joint and P is a collection of $\{L1, L2, L3, ....Lk\}$ k part association vector field for each pair or limb. The output of these two branches will be summed up using parsing algorithm and feed forward to multiple layers of convolutional net ultimately giving 2D anatomical key points for every person in the image.

### 3.2 Part detection using heat-maps

The heat maps produced by convolutional neural net are highly reliable supporting features. The heat maps are set of matrices that stores the confidence that the network has that a pixel contains a body joint. As many as 16 matrices for each of the true body joints. The heat map specifies the probability that a particular joint exist within a particular pixel location. The very idea of having heat maps provide support in predicting the joint location. The visual representation of heat maps could give an intuition of a presence of body joint. The darker the shade or sharper the peak represents a high probability of a joint. Several peaks represent a crowded image representing one peak for one person (**Figure 2**).

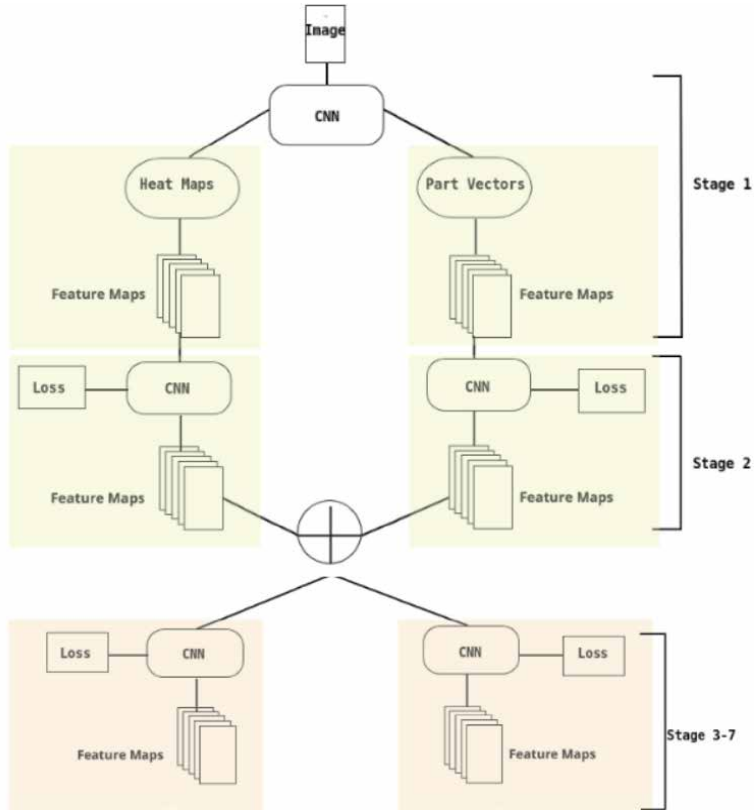

**Figure 1.**
*Schematic diagram of a multistage architecture. Two parallel branches feeds forward the network. Heat maps predicts the approximation while part association vectors predict association and orientations.*

Calculating the confidence map or heat maps $C_{jk}^{*}$ for each joint requires some prior information for comparison. Let $\boldsymbol{x_{jk}}$ be the empirical position of a body joint j of the person k. These confidence maps at any position m can be created by using the empirical position $x_{jk}$. The value of confidence map at location p in $C_{jk}^{*}$ is given by

$$C_{jk}^{*}(m) = exp\left(\frac{-\Delta^2}{\sigma^2}\right) \tag{1}$$

where σ is spread from the mean and $\Delta$ is the absolute difference of $x_{jk}$ and m.

All the confidence maps get aggregated by the network to produce the final confidence map. The final confidence map is generated by the network obtained from the aggregation of the individual maps.

$$C_{jk}^{*}(m) = max\left(C_{jk}^{*}(m)\right) \tag{2}$$

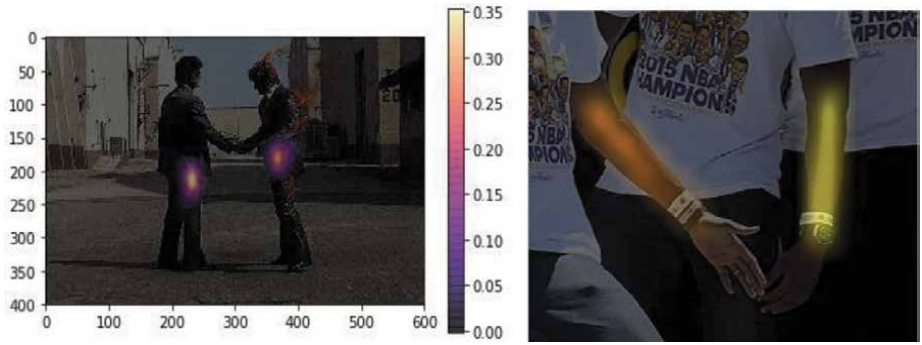

**Figure 2.**
*Illustration of one segment of the pipeline i.e. predicting heat maps through neural network. It gives the confidence metric with regards to the presence of the particular body part in the given pixel.*

These confidence maps are rough approximations, but we need the value for that joint. We need to extract value from the hot spot. For the final aggregated confidence map we take the max of the peak value while suppressing the rest.

### 3.3 Greedy part association vector

The problem that comes while detecting the pose is that even if we have all the anatomical key points how we are going to associate them. The hotspot or the key points itself have no idea of the context on how they are connected. One way to approach this problem is to use a geometrical line midpoint formula. But the given approach would suffer when the image is crowded as it would tend to give false association. The reason behind the false association is the limitation of the approach as it tend to encode only the position of the pair and not the orientations and also it reduces the base support to a single point. In order to address this issue, we want to implement a greedy approach known as greedy part association vector which will preserve the position along with the orientation across the entire area of pair support. Greedy part association vectors are the 2D vector fields that provides information regarding the position and the orientation of the pairs. These are a set of coupled pair with one representing x axis and the other representing the y axis. There are around 38 GPAVs per pair and numerically index as well (**Figure 3**).

Consider a limb j with 2 points at $x_1$ and $x_2$ for $k^{th}$ person in the image. The limb will have many points between $x_1$ and $x_2$. The greedy part association vector at any point c between $x_1$ and $x_2$ for $k^{th}$ person in the image represented by $G^*_{j,k}$ can be calculated as.

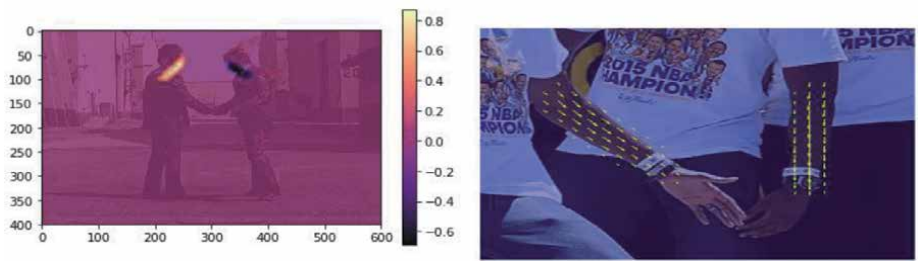

**Figure 3.**
*Illustration of the other segment greedy part vectors, preserving the position along with the orientations and finally associates the joints through greedy parsing.*

$$G^*_{j,k} = \hat{c} \text{ if c is on limb j and person k.Or 0 otherwise.} \tag{3}$$

where $\hat{c}$ a unit vector along the direction of limb equivalent to

$$\frac{x_2 - x_1}{\sqrt{x_2^2 - x_1^2}} \tag{4}$$

The empirical value of final greedy part association vector will be the average of GPAVs of all the person in the image.

$$G^*_j = \frac{\sum_k G^*_{j,k}}{n_j(c)} \tag{5}$$

where $G^*_{j,k}$ is the greedy part association vector at any point and $n_j(c)$ is the total number of vectors at the same point c among all people.

### 3.4 Multi person pose estimation

After getting the part candidates using non-maximum suppression, we need to associate those body parts to forms pairs. For each body part there are n numbers of part candidates for association. On an abstract level one-part can form association with every possible part candidate forming a complete graph (**Figure 4**).

For example, we have detected a set of plausible neck candidates and a set of hip candidates. For each neck candidates there is a possible connection with the right hip candidates giving a complete bipartite graph having the nodes as part candidates and the edges as possible connections. We need to associate only the optimal part giving rise to a problem of N dimensional matching problem which itself a NP hard problem. In order to solve this optimal matching problem, we need to assign weights to each of possible connection. This is where the greedy part association vectors come into the pipeline. These weights are assigned using the aggregated greedy part association vector.

In order to measure the association between two detected part candidates. We need to integrate over the predicted greedy part association vector found in previous section, along these two detected part candidates. This integral will give assign a score to each of the possible connections and store the scores in a complete bipartite graph. We need to find the directional orientation of the limb with respect to these detected part candidates. Empirically we have two detected part candidates namely $t_1$ and $t_2$ and the predicted part association vector $G_j$. An integral over the curve will give a measure of confidence in their association.

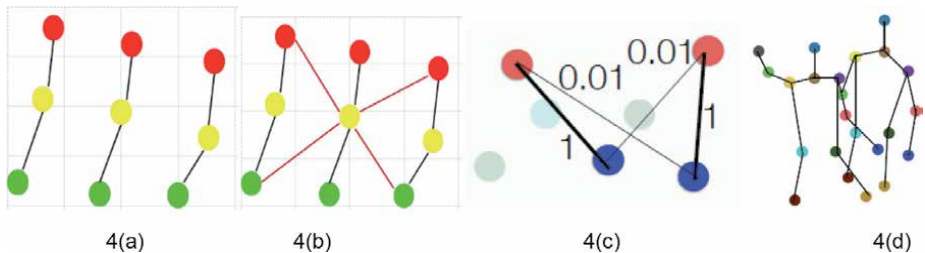

**Figure 4.**
*(a–d) Solving the assignment problem for associating body joints to form the right pair. Assigning weights to each possible connection with the help of greedy association vectors.*

$$E = \int_{i=0}^{i=1} G_j(c(m)) . \hat{d} . dm \qquad (6)$$

where $G_j(c(m))$ greedy part association vector and $\hat{d}$ is a unit vector along the direction two non-zero vectors $t_1$ and $t_2$.

After assigning weights to the edges our aim is to find the edge for a pair of joints with the maximum weight. For this we choose the most intuitive approach. We started with sorting the scores in descending manner followed by selecting the connection with the max score. We then move to the next possible connection if none of the parts have been assigned a score, this is a final connection. Repeat the third step until done.

The final step involves merging the whole detected part candidates with optimal scores and forming a complete 2D stick figure of human structure. One way to approach this problem is that let us say each pair of part candidates belong a unique person in the image that way we have a set of humans i.e. $\{H_1, H_2, H_3, \ldots . H_k\}$ where k is the total number of final connection. Each human in the set contain a pair i.e. pair of body parts. Let represent the pairs as a tuple of indices one in x direction and one in y direction. $H_i = \{(m_{idx}, m_x, m_y), (n_{idx}, n_x, n_y)\}$. Now comes the merging we conclude that if two human set shares any index coordinates with other set means that they share a body part. We merge the two sets and delete the other. We perform the same steps for all of the sets until no two human share a part ultimately giving a human structure.

# 4. Results

For the training and evaluating the final build we used a subset of a state-of-the-art public dataset, the COCO dataset. COCO dataset is collection of 100 K images with diverse instances. We have used a subset of those person instances with annotated key points. We have trained our model on 3 K images, cross validated on 1100 images and tested on 568 images. The metric used for evaluation is OKS stands for Object key point similarity. The COCO evaluation is based on mean average precision calculates over different OKS threshold. The minimum OKS value that can have is 0.5. We are only interested in key points that lie within 2.77 of the standard deviation (**Figure 5**).

Above table compares the performance of our model with the other state of the art model. **Table 1** shows the mAP performance comparison of our model with others on a testing dataset of 568 images. We can see clearly our novel approach outperforms the previous key point benchmarks. We can also see our model achieved a significant rise in mean average precision of 6.5%. Our inference time is 3 order less. **Table 2** presents the performance comparison on a complete testing dataset of 1000 images. Here again we can see our model outperforming the rest. Our model achieved a rise of almost 2.5% in mean average precision as compare to other models. The above comparison of our model with earlier state of the art bottom up approaches presents the significance of our model.

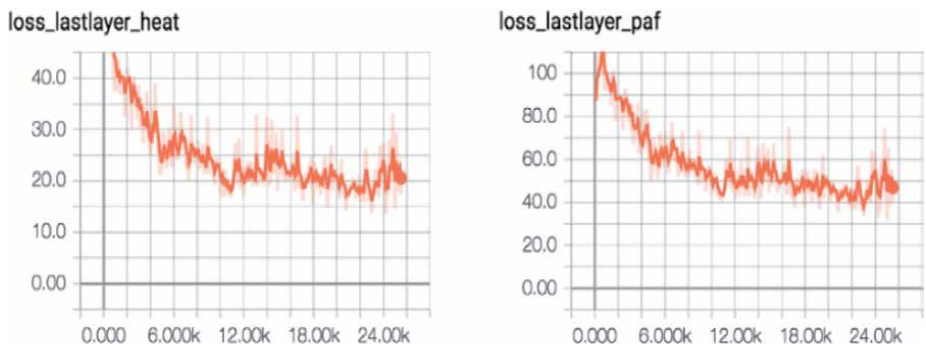

**Figure 5.**
*Convergence of training losses for both the heat maps (L) and greedy part vectors (R).*

| Method | Head | Shoulder | Elbow | Hip | Knee | Ankle | Wrist | mAp |
|---|---|---|---|---|---|---|---|---|
| Deep cut | 73.4 | 71.8 | 57.9 | 56.7 | 44.0 | 32.0 | 39.9 | 54.1 |
| Iqbal et al | 70.0 | 65.2 | 56.4 | 52.7 | 47.9 | 44.5 | 46.1 | 54.7 |
| Deeper cut | 87.9 | 84.0 | 71.9 | 68.8 | 63.8 | 58.1 | 63.9 | 71.2 |
| **Proposed Approach** | **90.7** | **90.9** | **79.8** | **76.1** | **70.2** | **66.3** | **70.5** | **77.7** |

**Table 1.**
*mAP performance comparison of our model with others on a testing dataset of 568 images.*

| Method | Head | Shoulder | Elbow | Hip | Knee | Ankle | Wrist | mAp |
|---|---|---|---|---|---|---|---|---|
| Deep cut | 78.4 | 72.5 | 60.2 | 57.2 | 52.0 | 45.0 | 51.0 | 54.1 |
| Iqbal et al | 58.4 | 53.9 | 44.5 | 42.2 | 36.7 | 31.1 | 35.0 | 54.7 |
| Deeper cut | 87.9 | 84.0 | 71.9 | 68.8 | 63.8 | 58.1 | 63.9 | 71.2 |
| **Proposed Approach** | **90.1** | **87.9** | **75.8** | **73.1** | **65.2** | **60.3** | **66.5** | **73.7** |

**Table 2.**
*Performance comparison on a complete testing dataset of 1000 images.*

## 5. Conclusion and future work

Solving one of the complex problems in computer vision was a huge challenge. Optimizing the current state of the art results and introducing a new approach to solving this problem is the highlight of this chapter. In this chapter, we presented a bottom up parsing technique which uses a non-parametric representation, features for localizing anatomical key points for individuals. We further introduced a multistage architecture with two parallel branches one of the branches estimates the body joints via hotspots while the other branch captures the orientations of the joints through vectors. We ran our model on a publicly available COCO dataset for training, cross validation and testing. Finally, we evaluated the results and achieved a mean average precision of 77.7. We compare our results with existing models and achieved and a significant rise of 2.5% in mAP with less inference time. We have showed the results in **Tables 1 and 2**. We aim to expand our project in future by proposing a framework for human pose comparator based on the underlying technology used in single person pose estimation to compare the detected pose with that of the target in real-time. This would be done by developing a model to act as an activity evaluator by learning physical moves using key points detection performed by the source and compare the results with the moves performed by the target along with a scoring mechanism that would decide how well the two sequence of poses match. In a nutshell, we aim to build an efficient comparison mechanism that would accurately generate the similarity scores based on the series of poses between the source and the target as the future scope of this project.

## Author details

Pooja Kherwa*, Sonali Singh, Saheel Ahmed, Pranay Berry and Sahil Khurana
Maharaja Surajmal Institute of Technology, New Delhi, India

*Address all correspondence to: poona281280@gmail.com

IntechOpen

# Ensemble Machine Learning Algorithms for Prediction and Classification of Medical Images

*Racheal S. Akinbo and Oladunni A. Daramola*

## Abstract

The employment of machine learning algorithms in disease classification has evolved as a precision medicine for scientific innovation. The geometric growth in various machine learning systems has paved the way for more research in the medical imaging process. This research aims to promote the development of machine learning algorithms for the classification of medical images. Automated classification of medical images is a fascinating application of machine learning and they have the possibility of higher predictability and accuracy. The technological advancement in the processing of medical imaging will help to reduce the complexities of diseases and some existing constraints will be greatly minimized. This research exposes the main ensemble learning techniques as it covers the theoretical background of machine learning, applications, comparison of machine learning and deep learning, ensemble learning with reviews of state-of the art literature, framework, and analysis. The work extends to medical image types, applications, benefits, and operations. We proposed the application of the ensemble machine learning approach in the classification of medical images for better performance and accuracy. The integration of advanced technology in clinical imaging will help in the prompt classification, prediction, early detection, and a better interpretation of medical images, this will, in turn, improves the quality of life and expands the clinical bearing for machine learning applications.

**Keywords:** machine learning, medical image, classification, ensemble method

## 1. Introduction

The application of technology to the health field is growing at a rapid pace and medical imaging techniques is part of the advancement of technology in the simplification of the medical imaging processes. It refers to an aspect of medical operations in this dispensation as it has overridden the traditional processes. Technological advancement is majorly responsible for the improvement in medicine via enhancement of imaging. The traditional perspectives on the clarification and diagnosis of the outcome of image processes require lot of processing time, human errors are foreseeable and the general outcome is unable to properly aligned with the history as the former ones is not easily available for comparison. These limitations motivated this research work so as to give insight to the applications of Ensemble Machine-Learning Algorithms for the Prediction and Classification of Medical Images.

## 1.1 Medical imaging

Medical Imaging refers to the application of different techniques to get various image modalities from the human body especially the affected area for further processing and to assist in diagnosis and the treatment of the patients [1]. Medical Image analysis is very important in today's world to be able to meet up with the growing population and in the current trend of a low medical expert in an expanding population. The healthcare industry has witnessed different technological disruptions that benefit humankind and more progress are being made. Having precision in the medical analysis of images will enhance the faster diagnosis and the treatment plan will be predicted along the line, this process increases the turnaround time with lots of lives being saved. It was reported that the market capacity of medical image analysis of software was estimated at USD 2.41 Billion in 2019, globally. In addition, the medical image processing market is forecast to reach about 8.1% in 2027 [2].

In medical imaging, radiology is a branch of medicine that uses imaging technology to diagnose and treat disease. The different types of diagnostic radiology examination comprise, Ultrasound, Plain X-Rays, Mammography, Fluoroscopy, Computed tomography (CT), including CT angiography, Magnetic Resonance Imaging (MRI), and Magnetic Resonance Angiography (MRA), Nuclear medicine, and Positron Emission Tomography [3]. Furthermore, the recent advancements in image processing as of the year 2020 are, EVP (Enhanced Visualization Processing) Plus, Bone Suppression, Pediatric Capabilities, Tube and Line Visualization, Longlength Imaging, and Pneumothorax Visualization. The operation of computermediated imaging processing is done using some computational framework, programming language, and algorithms which will make the prediction and classification to be an automated process and produce the result analysis [4].

The techniques involve professional in the medical field to use a particular device to create computerized images of an affected area in the body for diagnosis that will lead to treatment. The process may not necessarily include the opening of the affected part before using the radiological equipment in viewing the area that needed diagnosis. The medical personnel will later produce the image of the affected part from the device and then summarized the image analysis. Moreover, some medical imaging involves not just bones, blood vessels, and tissue without tearing apart the affected skin. Generally, the imaging techniques allow the healthcare providers to determine the type of treatment that is required for the ailment. Medical imaging has brought a major improvement in medicine as the difficult part or layers of the internal system of humans and animals can now be done using technological devices. The major technological improvement has decreased the manual work of the health providers and thereby give rise to specific and better treatment.

### 1.1.1 Medical image applications

The utilization of medical image in ultrasound, presents the internal part of a human or even animals; to be examined under an ultrasound device applicable joints, muscles, breast, blood vessels, pelvic, bones, and kidneys to say the least [5]. The X-ray is another category that employ the electromagnetic radiation and penetrates through the outer skin, layer to disclose the internal components [6]. Computer Tomography is a medical image that has an opening area in a circular form, the patient will be placed on it and slide inside; it produces images in a cross-sectional way [7].

Magnetic Resonance Imaging harness radio waves with magnetic fields to develop images with no harmful radiation compared to x-rays. It can generate images from soft tissue, bones, organs, cartilage, brains breast, spinal cord, liver, prostate, and ligaments, etc. [8]. Positron emission tomography (PET) scan can be used to scan the whole body or part of it. It uses a type of tool called tracer which will be swallowed or injected by the patient and then lie on the PET scanner to be examined by detecting the gamma rays by the device which is converted to images. PET scans can be applied in the diagnosis of conditions like brain disorders, tumors, and heart-related diseases, etc. [9]. The application of Ensemble Machine Learning technologies will improve the processing of medical images as it allows management, monitoring, early detection, diagnosis, assessment, and treatment of various medical challenges.

### 1.1.2 Challenges in medical image classifications

The application of Machine Learning vis-à-vis image generations promised a better way to visualize images and generally improves the medical condition of humans. This has brought out candid information from the input sample and it has encouraged a better decision to the treatment pattern of the concerned patient with the overall benefit of good living. These techniques are mostly used in the radiology field and pathology. The traditional model of result interpretation presents the problems of data bottlenecks, reliability, accuracy, and speed, and most of the issues with the traditional methods are being addressed with the machine learning algorithms and techniques. The more the technology is advancing, the more there is a need to have better analysis and clarity in medical imaging prediction and classification. The challenges that are associated with Machine Learning applications are; data availability, validation of methods, patient-specific model faster and accurate algorithms [10]. The learning models should be built into clinical operations and be intuitive so as not to cause serious damages. Training of the care providers and the documentations of analysis of the algorithms is another challenge that is needed to be looked into for future adaptation of research students and the users [11].

### 1.1.3 The benefits of medical image processing

The main advantage of medical image processing is the opportunity to explore the internal system of the human organ called anatomy, it is such an interesting thing to be able to view the inner system and see how things work. The rate of death is drastically reduced as more outcomes of the image processes allows timely intervention and prompt treatment of the ailment. Another benefit is the deep knowledge of the internal anatomy which helps to enhance treatment and diagnosis outcomes.

### 1.1.4 Medical image professional

The professionals that are involved in the operation of medical imaging is the clinicians, radiologist, and engineers' radiographers, radiologists, and engineers to know the anatomy patients. The medical imaging device is used to do imaging is operated by the radiographers and the result is sent to the caregiver to interpret it to the patient. One of the significant of Machine Learning in the aspect imaging is the enhancement, interpretation and analysis of results better than the manual result from human.

## 1.2 Machine learning

The advancement in computational applications and frameworks provides solutions to our everyday problems. Machine learning is one of the computational applications of algorithms and statistical models with the use of algorithms and statistical models, to carry out a task without explicit instructions but with the use of patterns to give inference. Machine learning is referred to the use of computer algorithms that support systems operation in training to automatically learn and enhance data to predict or classify the nature of such data through the use of patterns [12]. Generally, machine learning is a subfield of artificial intelligence that allows the systems to make decisions autonomously with no external support. The decision is made by finding valuable hidden layers of patterns within the complex data.

The machine-learning approach depends on the data type for input and output operation and problem type which is based on the applications on data for decision making and an embedded instruction to carry out the assignment with minimum supervision from the programmers [12]. Machine learning is classified as supervised learning, semi-supervised learning, and unsupervised and reinforcement learning while there are few hybrid approaches and other common methods [13].

### 1.2.1 Machine learning techniques

The categories of Machine Learning Techniques are mainly divided into four categories: Supervised learning, Unsupervised learning, Semi-supervised learning, and Reinforcement learning [12]. The techniques are discussed further according to their applicability of solving real-world problems.

### 1.2.1.1 Supervised learning

In the supervised learning category of machine learning, the algorithms (step by step method of solving a problem in a particular format) operates in such a way that it will develop a mathematical model (translating or encoding a problem into a mathematical formulations) of the data which comprises the inputs (data sent to a computer system) and the expected outputs (processed information sent out from a computer) [14]. The data supplied is also categorized as the training data which comprises the sets of training examples with one or more inputs. The mathematical modeling is applied in the supervised learning uses array vector (feature vector for extraction) and the data to be trained by matrix. The algorithm that enhances and improves the outcomes in the accuracy of the outputs for classification or prediction purposes has learned the task and therefore it can give a good outcome [15].

### 1.2.1.2 Unsupervised learning

Unsupervised learning algorithms operate in such a way that it takes set of data and detect the patterns in it for grouping or clustering purpose. Unsupervised learning algorithms identify resemblance in the data and react based on the presence or absence of such identity in each new piece of data. The algorithms learn from test data that is not labeled, classified, or categorized. Unsupervised learning analyzes unlabeled datasets without the need for human interference, i.e., a data-driven process [16]. The unsupervised learning tasks that are common are anomaly detection, dimension reduction, clustering, density estimation, feature learning, finding association rules, etc. [17].

### 1.2.1.3 Semi-supervised learning

The semi-supervised learning is situated between unsupervised learning (with no labeled training data) and supervised learning (with labeled training data). It is a hybrid form of machine learning techniques because it operates on labeled and unlabeled data which brings a better accuracy. The major aim of unsupervised learning is to give great outcomes for prediction than the ones done with labeled data. The application areas of semi-supervised learning are text classification, fraud detection, machine translation, etc. [18].

### 1.2.1.4 Reinforcement learning

Reinforcement learning in machine learning parlance refers is concerned with the use of software agents and machines to make the decision automatically in an environment to improve efficiency. Generally, reinforcement learning is used in operation research, game theory, information theory, swarm-intelligence, and genetics algorithms, etc. The learning uses the reward or penalty system, and the primary goal is to use leading obtained from environmental parameters to validate the reward or to minimize the risk involved. The algorithms are used in autonomous vehicles or in learning to play a game against a human opponent, it is an effective tool in training AI models for increase automation which is used in robotics, autonomous driving tasks, manufacturing, and supply chain logistics [12].

### 1.2.2 Applications of machine learning

The areas of applications of machine learning to various fields are enormous such as agriculture, engineering, medical diagnosis, natural language processing, banking, bioinformatics, games, insurance speech recognition, and recommended system etc. [19], used machine learning technology to make a medical diagnosis in developing a cure for Covid-19. Similarly, machine learning is also applied in [20] work to predict visitors' behavior in marine protected areas, while [21] applied machine learning to smartphone performance optimization. According to [22] by Mayo, the most common data science/machine learning methods used from the period of 2018–2019 are regression, decision trees/rules, clustering, visualization, random forests, statistics, K-nearest neighbors, time series, ensemble methods, text mining, principal component analysis (PCA), boosting, neural networks (deep learning), gradient boosted machines, anomaly/deviation detection, neural networks Convolutional Neural Networks (CNN) and support vector machine (SVM).

### 1.2.3 Machine learning and deep learning

Deep learning refers to a distinctive subtype variant in the machine learning, also a subclass in the domain of artificial intelligence (AI). Furthermore, Machine learning primarily means a computer that learns from data and makes predictions using algorithms. Machine learning yields to some environmental parameters, conversely, the deep learning operates in a quick manner and adapt to it using constant feedback in building on the models. Deep Learning system leverage on the Neural Networks which imitates the brain of human with an embedded multiple-layer architecture. It also learns through the data to carry intelligent decisions [23].

### 1.2.4 Ensemble machine learning

Ensemble learning is a general meta-approach to machine learning as it looks for the best predictive performance using a combination of the methods to achieve the best accuracy [24]. The use of different machine learning algorithms individually may not be able to give the best outcomes, hence the combination of the algorithms will combine all the strength of the model and brings out a better accuracy. The diagram below shows different Machine learning algorithms. The Ensemble learning methodology for the prediction and classification of medical images has been established to have a better result than using a single classifier. The reviews on elated works in artificial intelligence systems to detect fractures in the body cited few works on Convolutional Neural Networks (CNN) for fracture detection [25].

The authors also noted that stacking with Random Forest and Support Vector algorithm, with neural networks were mostly engaged. The development of an Ensemble deep learning application for ear disease using otoendoscopy images by [26]. They perform well with the average of accuracy taken as 93.67% for the five-fold cross-validation using learning models based on ResNet101 and Inception-V3. Furthermore, another author developed a three-dimensional bone model system which is based on employment of x-ray images for distal forearm engaging the convolutional neural networks [27]. The deep learning framework is employed in estimating and to construct a high accuracy for three-dimensional model of bones. The result gives correctness of the evaluation of CNN to reduce exposure to computer tomograph device and cost. In summary, the application of Ensemble methods to medical imaging can be perused with all intent as the accuracy recorded is far more than the single classifiers or the traditional methods.

There are three classes of ensemble learning, bagging, stacking and boosting. The bagging is concerned with having many decisions on a different sample of the very same dataset and get the average of the prediction; while the stacking is concerned with the fitting of many different types of models on the same data while using another type of model to learn the combined predictions. The boosting involves the addition of ensemble members in a sequential manner that will correct the former prediction by the other models then gives the average of the predictions (**Figure 1**) [24].

### 1.3 Neural networks

The Neural Networks is an aspect of machine-learning that comprises different node layers which include the input, hidden, and output layers. The Network is used in most deep learning architectures. Neural Networks works in a manner that the nodes connect with their different weight and threshold. More so, if a node for instance has an output that is more than that of the threshold, then it will be triggered and will send the data involved to the layer that is next and if not, there will be no data being activated to the succeeding layer in the network.

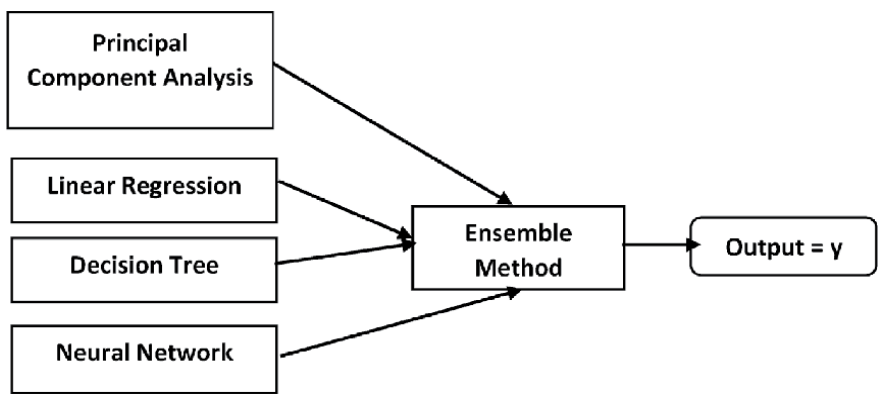

**Figure 1.**
*Ensemble machine learning combine several models for better accuracy.*

### 1.3.1 Convolutional neural network

The traditional manual process employed in the prediction and classification of images convincingly, wastes time, wrong diagnosis is another major problem attributed to it. The convolutional neural network provides a better and more scalable method in the medical imaging process. The CNN involves the identification of images through a computational approach that combines linear algebra and matrix multiplications. CNN outperformed other networks in applications like image processing and speech recognition, The CNN has three parts, the convolutional, pooling, and fully-connected layer. The convolutional part is where the major computation happens to be the building block among the three and it comprises the data, filter, and feature area. Pooling layer is responsible for the data sample dimension reduction known as downsampling. The pooling layers also holds a filter and it moves over the input but may not have weight. The pooling is sub-divided into Max and Average pooling with the functions of calculating the maximum and or average value respectively [28]. The fully-connected layer, output layers are fully joined via a node to former layer and do classification tasks through the feature extracted from the preceding layer.

## 2. Literature review

Visual Attention Mechanism-MCNN was developed using classification Algorithm According to the authors, the motivation was a result of the complexity of medical images as a traditional way of analyzing medical images is presented with some disadvantages and it may not be able to meet up with the growing demands. The methodology is based on a medical classification algorithm that uses visual attention mechanism multiscale convolutional neural network. The results show that the medical image classification was more accurate with stability, robustness and improved with the employment of the deep learning methods than the traditional methods [29].

In the article titled "Machine Learning: Algorithms, Real-World Applications and Research Directions", they stated the positive impact and the present challenges of Machine Learning algorithms to the real-world application and research direction. The authors explained further that is an enormous advantage to the intelligent and automatic operations of machine learning to a different field of study [13].

The primary goal of machine learning is in the automation of human assistance by training an algorithm on suitable data Machine Learning as they focused on learning types. The author stated different machine learning techniques and the ways to apply them to various operations. The algorithms are used for many applications that include data classification, prediction, or pattern recognition [30].

Medical image analysis based on a deep learning approach which helps in different clinical applications as it occurs in medical procedures for monitoring, diagnosis, and detection was proposed for early detection, monitoring, diagnosis, and treatment evaluation of various medical conditions. The methodology employed artificial neural networks and the detailed analysis of deep learning algorithms and delivers great medical imaging applications [31].

The applications of Machine Learning Predictive Models in the Chronic Disease Diagnosis, presented different reviews on the employment of machine learning predictive models for the diagnosis of chronic diseases. The motivation of the authors arises from the major health cost of disease especially for those that attract lifelong treatment. The methodology of the predictive models involves the analysis of 453 papers from 2015 and 2019, PubMed (Medline), and Cumulative Index to Nursing and Allied Health Literature (CINAHL) libraries. The result of the investigation shows that the support vector machines (SVM), logistic regression (LR), clustering was among the most commonly used models [32].

An ensemble approach in the classification of bone fracture using different CNN is presented. The authors were motivated by the need to help the emergency nature of bone fracture for a prompt response than the usual process of going through the X-ray and then forwarded to the doctors for better interpretations of the result. The process can take a long time as nothing significant can be until the result is out. To increase the processes that are involved in carrying out this important aspect of a medical emergency, a new method is proposed by the author. The research employs ensemble machine learning techniques using CNN to classify the bone fracture images with stacking methodology for reliable and robust classification. The result shows that the ensemble method is more reliable and forms a robust output than the manual works from the providers [33].

The classification of shoulder images using X-ray images with deep learning ensemble models for diagnosis, with the data gathered from the X-ray images from magnetic resonance imaging and computer tomography. The target of the research is to determine the state of the images through classification using artificial intelligence. The study employs twenty-six deep learning models to detect the shoulder fracture and evaluate using the musculoskeletal radiographs datasets, together with ensemble learning models. The twenty-eight classification was performed and the overall accuracy was from Cohen's kappa. The best score was 0.6942, taking an ensemble of ResNet34, DenseNet169, DenseNet201, and a sub-ensemble of different convolution networks [34].

In 2019 some authors constructed Xception, Resnet, and Inception-V3 CNNs to determine appearance of fractures on the ankle through ensemble methods. The approach involved using five hundred and ninety-six ankle cases both the normal and abnormal for the processing. The programming language applied is Python with the TensorFlow framework. The outputs were in radiographic views of one and three. The ensembles were made from combined convolutional networks, and the voting approach was applied to incorporate results from the views and ensemble models. The results show 81% accuracy despite the small dataset used [35].

Furthermore, in 2021 an investigation was done for the detection of tuberculosis in x-ray images of the chest via Ensemble Learning method together with hybrid feature descriptors as Tuberculosis (TB) is a major health challenge that has a record of high mortality and early diagnosis is key to early control of the disease. The author proposed an innovative approach to TB detection that combines handcrafted features with deep features using convolutional neural networks through Ensemble Learning. The dataset captured from Montgomery and Shenzhen for the critical evaluation of the system. The result shows the distinction of the Ensemble machine Learning method over the other classifier as a single unit in classifications as the operating characteristics curve reaches 0.99 and 0.97 respectively from the Shenzhen and Montgomery [36].

In the Ensembles Learning for COVID-19 Detection using Chest X-Rays. The development applied Ensemble approach for the X-ray classification in detecting pulmonary manifestation of COVID-19 is established. The methodology entails a customized convolutional neural network and ImageNet pre-trained models while the dataset is from publicly available collections. The best predictions from the best accurate models are combined via different Ensemble approaches for better performance evaluation. The result shows a major improvement of 99.01% accuracy and the area under the curve to be 0.9972 for the detection of COVID-19 on the dataset collections. The blend of the iterative, Ensemble, and modality bases knowledge transfer shows a major improvement in the prediction level [37].

## 3. Ensemble methods and design analysis

The ensemble system of machine learning application is a type of method that allows the combination of multiple models to produce an output that is improved and enhanced than applying the single model. The ensemble predictive model generated is to have a decrease in variance (bagging), bias (boosting), or to improve the predictions (stacking). The ensemble machine learning methods consist of sequential, parallel, homogenous and heterogenous ensemble, while the technical classifications of ensemble method are, bagging, boosting, stacking, and random forest [38].

### 3.1 Ensemble methods

In the continuous method, the base learners (where the data dependency stays) are being initiated successively. Moreover, there is a dependency on the former data of all other data in the base level and to get the performance analysis of the system, the wrongfully-labeled are adjusted based on the weight. An example of this type of analysis is the boosting method. The parallel methodology ensures that the base learner is initiated in a parallel manner and the data dependency is not available, and all the data are generated separately. A very good example of this model is the stacking method [38].

The homogenous ensemble method can be applied to a large number of the dataset as it employed the combination of the very same types of classifiers. The dataset is always different for each classifier and the model works well following the results collection for each of the classifier models. The feature selection method is the same for different training of data. The major disadvantage of this type of model is that the computational cost is very expensive. The popular type of this model is the bagging and boosting method. Conversely, the heterogeneous ensemble method combines different classifiers and each of them is generated on the same data, these types of methods are used for small datasets and the feature selection method is different for the same training dataset. An example of this type of classifier is stacking [38].

### 3.2 Technical classification of ensemble approach

#### 3.2.1 Bagging

The bagging method ensemble learning represents the bootstrapping and aggregation of combining two models into one ensemble model. Bagging uses random sampling to reduce the variance of the model by creating more data at the training stage of the model and every element in bagging has a way to appear in the new dataset. This method decreases the variance and narrowly changes the prediction to an expected output [38].

#### 3.2.2 Boosting

The boosting method of ensemble classification makes use of a continuous method of classifying based on the features that the next model will utilize. The boosting methods make a stronger learner out of a weak learner model using weight average. The much stronger trained model relies on the multiple weak trained models. The weak learner is the one with the less correlated true classification and the next one is improved over the last one which will be a bit more correlated, the aggregation of the weak learners makes up the strong learner that is well correlated with accurate classification [38].

### 3.2.3 Stacking

In the stacking method, a different classifier is also used using the regression techniques or multiple combinations of the models. The operation involves training the lower-level with the entire dataset in which the outcomes are used to train the combined model. This is different from boosting in the sense that the lower level is in line with parallel trained. The prediction from the outcome of the lower level forms the next model as a stack. The stack operation is such that the top of the stack is the most trained and has goo prediction while the down of the stack is the least trained. The predictions continue until the best surfaced with fewer errors [38].

### 3.2.4 Random forest

The random forest models employ tree operations to carry out the sampling of the dataset. The tree is fitted on bootstrap samples and the output is aggregated together to reduce variance. It uses the features to sample the dataset in a random subset for the creation of the tree to reduce the correlation of the outputs. This model of random forest is best for determining the missing data as it selects a random subset of the sample to reduce the possibility of having common prediction values, with each tree having a different structure. The resultant variance result from the average of the lesser prediction from different trees gives better output [38].

## 3.3 Ensemble framework

In machine learning, ensemble methods and/or models employ the blend of different learning algorithms to give a better predictive performance that displays an outcome that surpasses what could be gathered from the single application of any of the learning models. Experimentally, ensembles models produce a more improved output as there is major distinctiveness among the model. The ensemble framework is depicted in **Figures 2** and **3** below.

## 3.4 Ensemble framework component

The framework of ensemble learning as depicted in **Figure 2** consists of the dataset, samples of data, different types of learners and the predicted output. The dataset denotes the medical images such as Ultrasound, Plain X-Rays, Mammography, Fluoroscopy, Computed tomography (CT), including CT angiography, Magnetic Resonance Imaging (MRI), and Magnetic Resonance Angiography (MRA), Nuclear medicine, and Positron Emission Tomography e.tc. The samples are the data to be trained by each model of the classifier before combing them together for the ensemble classifier to aggregate all the model predictions.

## 3.5 The classification and prediction block diagram

The block diagram of the system incorporates the following: The dataset is the pre-processed images that will be divided into training and validation set. The input function is then activated while estimate or facilitates construction and training of the models with appropriate APIs that include four sequential actions, they are training, evaluation, prediction, and export. The training and evaluation will be looped. The training setcompute the loss and adjust the weights of the model using gradient descent, the evaluation set validate the model while training and adjust the learning rate then and take the best version of the model, while the test set is used to compare different model approaches, and report the final result of the model (**Figure 4**).

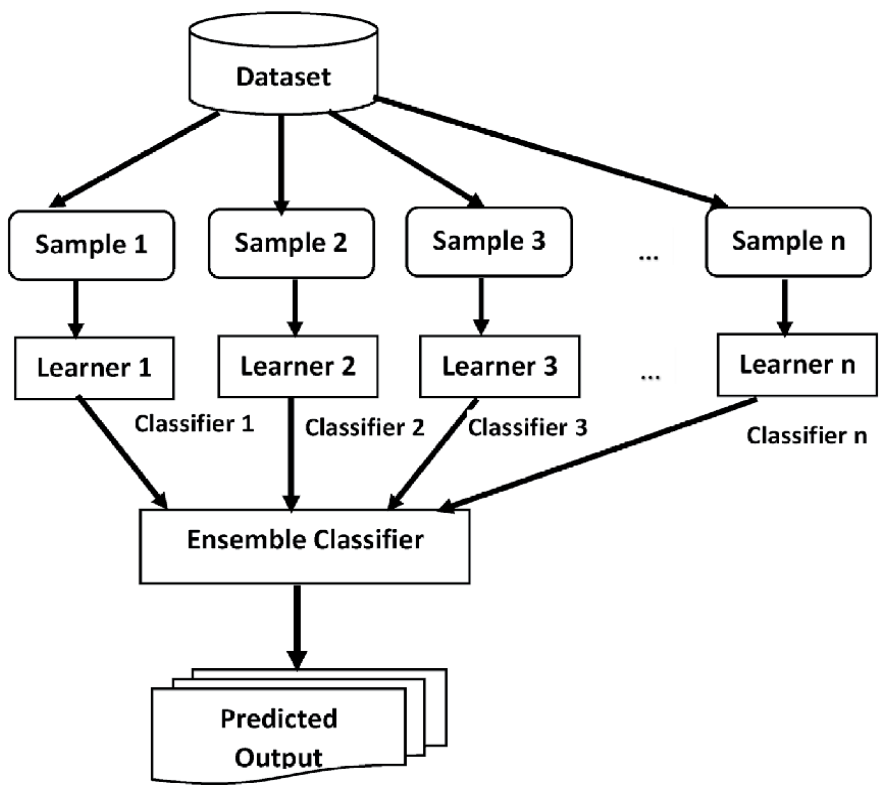

**Figure 2.**
*Ensemble framework for prediction and classification of medical images.*

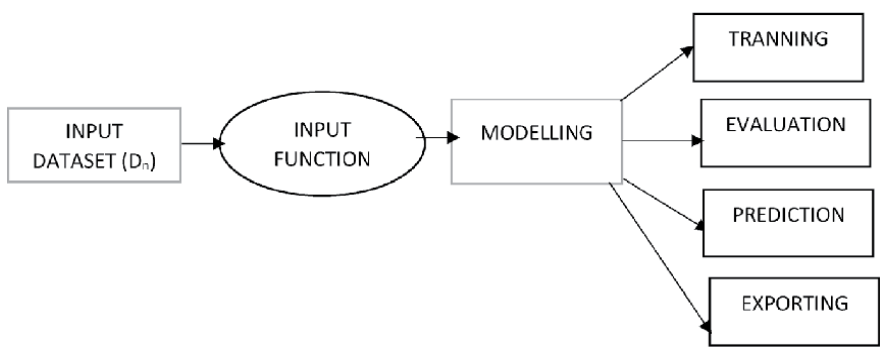

**Figure 3.**
*Block diagram of the classification and prediction.*

## 3.6 Methodology

### 3.6.1 Techniques/method(s)

Different tools that can be adopted for this type of research include the following:

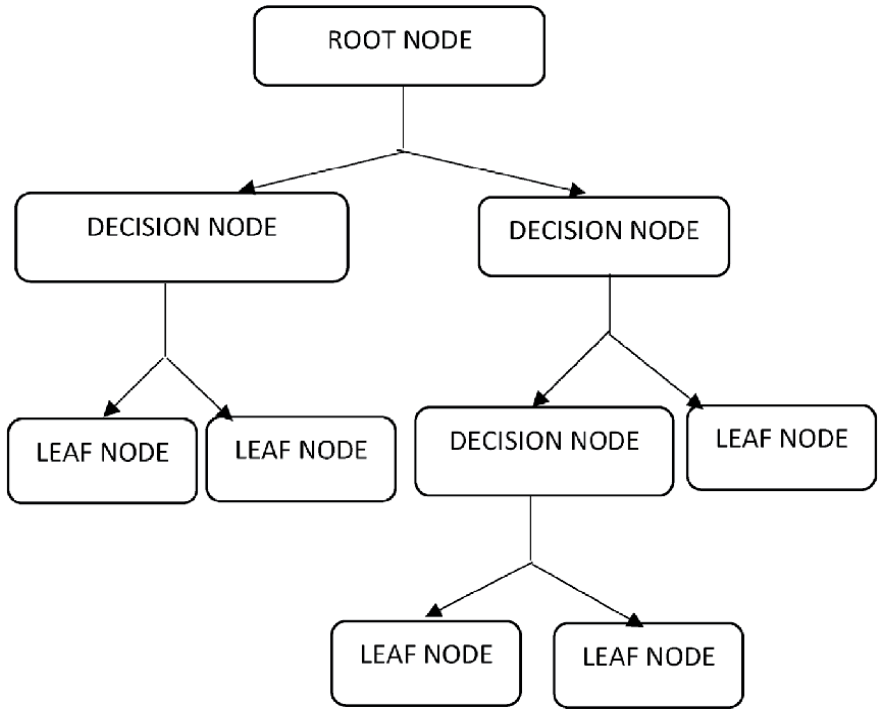

**Figure 4.**
*Decision tree structure with tree and subtree.*

    i. TensorFlow image recognition with object detection.

    ii. Python programming language for scripting.

    iii. Evaluation using MATLAB

### 3.6.2 Data extraction

Data will be extracted from the medical images represented by the distribution of pixels. To improve the data, standardization of the data is necessary before the application of the principal component analysis and these will enable each of the feature to have its mean = 0 while the variance = 1 Mathematically, this can be done by subtracting the mean and dividing by the standard deviation for each value of each variable. PCA makes maximum variability in the dataset more visible by rotating the axes.

$$Z = \frac{x - \mu}{\sigma} \tag{1}$$

### 3.6.3 The use of covariance matrix

The next step is to create a covariance matrix by constructing a square matrix to express the correlation between two or more features in a multidimensional dataset.

### 3.6.3.1 Principal component analysis

The PCA is used for technique dimensionality reduction with the operation of using measurement vector for distance classification, it explores the pattern recognition techniques [39]. It is best used for feature extraction, removal of redundancy, and feature extractions. A simple analysis of the PCA algorithm for image classification is given below. The PCA employs the Eigenvectors and Eigenvalues by the sorting of the eigenvectors in highest to lowest order, then select the number of principal components using the following equations:

$$PC1 = w_{1,1} \text{ (Feature A)} + w_{2,1} \text{ (Feature B)} \dots + w_{n,1} \text{ (Feature N)} \tag{2}$$

$$PC1 = w_{1,2} \text{ (Feature A)} + w_{2,2} \text{ (Feature B)} \dots + w_{n,2} \text{ (Feature N)} \tag{3}$$

$$PC1 = w_{1,3} \text{ (Feature A)} + w_{2,3} \text{ (Feature B)} \dots + w_{n,3} \text{ (Feature N)} \tag{4}$$

The first of the principal components (PC1) is a synthetic variable built as a linear combination to determine the magnitude and the direction of the maximum variance in the dataset. The second principal component (PC2) is also a synthetic linear combination that captures the remaining variance in the data set and is not correlated with PC1. The rest of the principal components likewise capture the remaining variation without being correlated with the previous component. PCA allows resizing of medical images, patterns recognition, dimensionality reduction, and visualization of multidimensional data. The covariance matrix is symmetrical and has the form:

$$C = \begin{matrix} w11 & w12 & w13 \\ w21 & w22 & w23 \\ wn1 & wn2 & wn3 \end{matrix} \tag{5}$$

eigenvalues of matrix C are the variances of the principal components.

$$PCi = w_{i1}X1 + w_{i2}X2 + \dots w_{ip}Xp \tag{6}$$

Where $var(Pci_i) = \lambda_i$ and the constants $w_{i1}, w_{i2}, \dots, w_{ip}$ are the elements of the corresponding eigenvector,

$$\text{Therefore, } w^2_{i1} + w^2_{i2} + \dots + w^2_{ip} = 1 \tag{7}$$

The sum of variances of the PCA is equal to the sum of variances of the original variances. The principal components contain all the variation of the original data.

### 3.6.3.2 Logistic regressions

Logistic regression is a linear algorithm (with a non-linear transform on output). It can have a linear relationship between the input variables and the output. Data transformation can come from the input variables result in a more accurate model [40]. In the following Eqs. (8)-(10) z is the output variable, x is the input variable where w and b will be initialized as zeros, to begin with, and they will be modified by numbers of iterations while training the model. The output z is passed through a non-linear function. The commonly used nonlinear function is the sigmoid function that returns a value between 0 and 1. The logistic regression uses the basic linear regression formula as:

$$\text{for a } 0 \leq h\,\theta(x) \leq 1 \text{ (binary)} \tag{8}$$

$$h_{\theta(x)} = (\omega x + b), \text{ where } \theta \text{ is the vector of parameters (w, b)} \tag{9}$$

$$g(z) = \frac{1}{1 + e^{-z}} \text{ Where g(z) is the Sigmoid function} \tag{10}$$

$$h_{\theta(x)} = \frac{1}{1 + e^{-(\omega x + b)}} \tag{11}$$

To train a logistic classifier $h_{\theta(x)}$ for each class of y prediction gives:

$$y_{predict} = g(z) = \frac{1}{1 + e^{-z}} \text{ (Sigmoid Function) which is optimized.} \tag{12}$$

### 3.6.3.3 Decision tree

The building of the decision tree begins with the root and then moves to the sub-nodes. The nodes represent characteristics features that represent the decision points, hence the information is classified. The nodes are linked at different levels to each other by branches that represent different decisions made by testing the status of the features in the node. It is a type of supervised machine learning algorithm that uses the if and else statement with a tree data structure for its operation [41].

The decision tree algorithm is based on the use of Shannon Entropy which determine the amounts of information of an event. If the probability distribution is given P as P = (p1, p2, P3, … , Pn), with a sample data S, the information carried by this distribution is called the Entropy of P and is given by the following equation, where (Pi) is the probability that the number (i) will appear during the process.

$$P = (p1, p2, P3, \ldots, Pn) \tag{13}$$

$$\text{Entropy } (P) = \sum_{i=1}^{n} Pi \, X \, log \, (Pi) \tag{14}$$

$$Gain(P, T) = \text{Entropy } (P) - \sum_{j=1}^{n} (Pj \, X \, Entropy(Pj)) \tag{15}$$

Where Pj is a set data of all values of (T).

### 3.6.3.4 Artificial neural network

Neural networks use set of algorithms to recognize patterns in medical images through machine perception, labelling of the raw input. The layers are made of *nodes* [42]. loosely patterned. A node combines input from the data with a set of coefficients, or weights, that either amplify or dampen that input, thereby assigning significance to inputs concerning the task the algorithm is trying to learn. These input-weight products are summed and then the sum is passed through a node's so-called activation function, to determine whether and to what extent that signal should progress further through the network to affect the outcome (**Figure 5**).

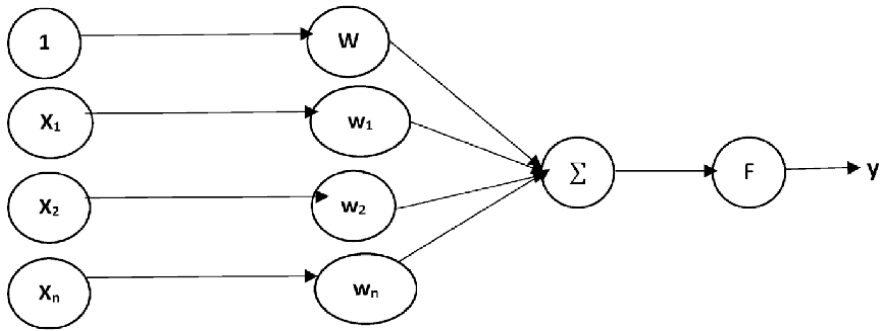

**Figure 5.**
*Neural network modeling and output function.*

$$\text{The input layer is denoted by } X_1, X_2, X_3, \dots X_n, \qquad (16)$$

the first input is $X_1$ and the other several inputs to $X_n$.

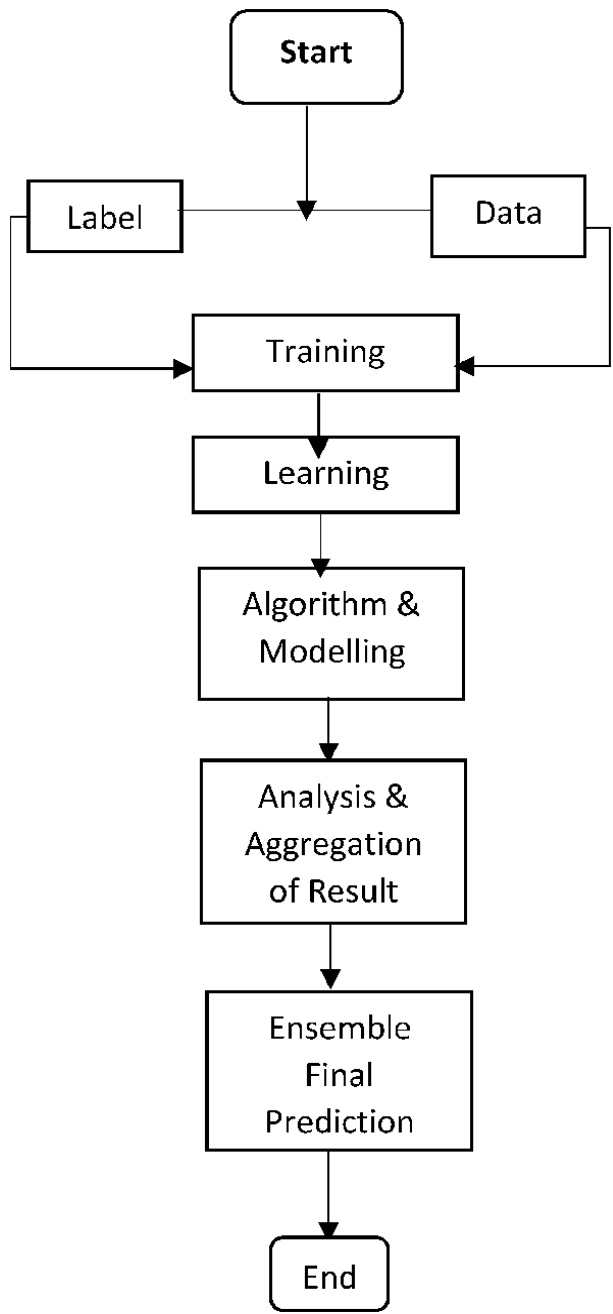

**Figure 6.**
*Flowchart diagram of the classification and prediction.*

The connection weight is also denoted by $W_1, W_2, W_3, ... W_n$. (17)

The weight signifies the strength of the node.

$$a = \sum_{j=1}^{n} w_\theta x_j + b \qquad (18)$$

Where a is the output generated from multiple input. The output is a weighted combination of all the inputs. This output a is fed into a transfer function f to produce y.

### 3.7 The system flowchart

Image Pre-processing: The aim of this process is to improve the image data (features) by suppressing unwanted distortions and enhancement of some important image features so that our Computer Vision models can benefit from this improved data to work on. Detection of an object: Detection refers to the localization of an object which means the segmentation of the image and identifying the position of the object of interest. Feature extraction and Training: This is a crucial step wherein statistical or deep learning methods are used to identify the most interesting patterns of the image, features that might be unique to a particular class, and that will, later on, help the model to differentiate between different classes. This process where the model learns the features from the dataset is called model training. Classification of the object: This step categorizes detected objects into predefined classes by using a suitable classification technique that compares the image patterns with the target patterns (**Figure 6**).

## 4. Discussion

Ensemble machine learning application method for the classification and prediction of medical images is proposed in this study. The study focuses on the use of different machine learning algorithm as a combined model to get better result as a result of the aggregation of the models. The images will be first pre-processed, augmented, fed into the classifier for training and testing for evaluation and then the predicted result.

## 5. Conclusion

Conclusively, the traditional approach of reading results from medical images by the care providers cannot be devoid of errors as well as the time take to predict the result of such images. In case of emergencies, early detection of any ailment will attract prompt, treatment and time management is also a vital aspect of the process. The application of Ensemble machine learning algorithms to medical imaging is a justifiable approach to obtain a better accuracy compared to the single classifier or even the traditional reading of results from the radiologist. This paper established a brief study that enclosed different machine learning algorithms and the combinations to make an ensemble learning classifier. More so, it stated the positive effect of having an ensemble approach for the prediction and classification of medical imaging. The paper also includes some of the reviews that are relevant to the study area. The explanation of medical images is one of the most demanding aspect of prediction and classification of diseases that occurs daily in medical diagnosis. The application of machine learning in the classification and prediction of medical images is growing at a geometric rate. A high level of precision and knowledge in the specific field may be required for accurate inference. The effectiveness and the efficiency of the combinations of different machine learning algorithm will enable better understanding of biological integration and analysis of knowledge as it improves access and transition in healthcare. The impact will reduce cost, earlier detection of diseases and accurate interpretation of results than using the single model. The research has the potential to cause a major shift in the field of medicine.

## Acknowledgements

The authors are grateful to the participants who contributed to this research especially Mr. Gowin Akinbo for his immense contribution in proof reading this paper for publication.

## Conflict of interest

The authors declare no conflicts of interest for this publication.

## Author details

Racheal S. Akinbo* and Oladunni A. Daramola
Federal University of Technology, Akure, Nigeria

*Address all correspondence to: rachea.akinbo@gmail.com

IntechOpen

# Delivering Precision Medicine to Patients with Spinal Cord Disorders; Insights into Applications of Bioinformatics and Machine Learning from Studies of Degenerative Cervical Myelopathy

*Kalum J. Ost, David W. Anderson and David W. Cadotte*

## Abstract

With the common adoption of electronic health records and new technologies capable of producing an unprecedented scale of data, a shift must occur in how we practice medicine in order to utilize these resources. We are entering an era in which the capacity of even the most clever human doctor simply is insufficient. As such, realizing "personalized" or "precision" medicine requires new methods that can leverage the massive amounts of data now available. Machine learning techniques provide one important toolkit in this venture, as they are fundamentally designed to deal with (and, in fact, benefit from) massive datasets. The clinical applications for such machine learning systems are still in their infancy, however, and the field of medicine presents a unique set of design considerations. In this chapter, we will walk through how we selected and adjusted the "Progressive Learning framework" to account for these considerations in the case of Degenerative Cervical Myeolopathy. We additionally compare a model designed with these techniques to similar static models run in "perfect world" scenarios (free of the clinical issues address), and we use simulated clinical data acquisition scenarios to demonstrate the advantages of our machine learning approach in providing personalized diagnoses.

**Keywords:** Precision Medicine, Personalized Medicine, Neural Networks, Degenerative Cervical Myelopathy, Spinal Cord Injury, Continual Learning, Bioinformatics

## 1. Introduction

The classical practice of medical is undergoing a transition as new scales of data production become increasingly common. This transition presents the field with major analytical challenges that necessitate new and creative ways of engaging. The use of machine learning is one of the most important developments supporting this transition, as its methods are ideally suited (and in fact benefit from) the massive scale of data collection that is increasingly becoming the norm. With the rapid growth in high-throughput technologies for genetics, proteomics, and other biological metrics, alongside the recent wide-spread adaptation of electronic health records [1, 2], more data than ever has become available to feed into a machine learning system. The classical practice of medicine is typified by giants such as Dr. William Osler [3] whose diagnostic acumen improved the lives of many. In the very near future (and, in many cases, the present), physicians and diagnosticians will work with more data than they could possibly interpret. Machine learning is one of many tools which will help alleviate this, helping to guide many diagnostic and therapeutic decisions made by the clinical team, and if implemented well, should support patients' overall health. This potential realization of "precision medicine" is based on the belief that each patient has unique characteristics which should be accounted for when treating them [4].

While precision medicine has already demonstrated major benefits in fields like pharmacology [5] and oncology [6–8], a number of potential applications remain in other medical fields. In this chapter, we will demonstrate this using spinal cord disease, specifically by examining its application to Degenerative Cervical Myelopathy (DCM). DCM is a condition when the bones and joints of the human neck (cervical spine) degenerate with age, causing a slow progressive 'squeeze' of the spinal cord. This progressive condition has a significant effect on patient quality of life. Symptoms include pain, numbness, dexterity loss, gait imbalance, and sphincter dysfunction [9, 10], with symptoms often not appearing until permanent damage has already occurred [11]. MRI scans are typically used as part of the diagnostic process, and demographic factors have also been shown to be effective in predicting DCM severity [12]. An additional challenge is that patients can exhibit the hallmarks of DCM without developing symptoms [13], suggesting that a wide range of factors may be contributing to the illness's severity. Despite all of this, research into precision medical approaches and diagnostics have been sorely lacking; to the best of our knowledge, only 4 published studies involving DCM (also referred to as Cervical Spondylotic Myelopathy) exist which utilize machine learning [7, 14–16], coming from only three different groups (including our own), and with only one utilizing MRI data [16].

In this chapter, we will discuss how we went about designing a machine learning process, focusing on considerations required for clinical data specifically. We first explore how data should be managed and stored, before moving into data preparation procedures. Finally, we move onto the design considerations for the machine learning model. We will focus on models made for diagnostic prediction, rather than outcome prediction; however, we intend this only as a first step in using machine learning to support patient care, with future work moving toward models that provide personalized therapeutic recommendations as well. Throughout this chapter we will apply the techniques being discussed to DCM to help contextualize them. Some preliminary results for the final resulting system will also be shown, as a 'proof-of-concept', using the CIFAR-10 dataset modified to replicate clinical circumstances. We hope that this will provide a road-map for future machine learning driven precision medicine projects to follow.

## 2. Precision medicine machine learning system design

We have previously published work using spinal cord metrics generated by the Spinal Cord Toolbox [17] alongside simple linear and logistic regression models [16]. While it found moderate success, our results suggested that complex aspects of the spinal cord morphology are likely the key to an accurate model, with simple regression analyses alone appearing to be insufficient. Said study also only used MRI-derived metrics, resulting in our model being unable to use non-imaging attributes to support its diagnostic conclusions, something which has been shown to aid model accuracy in other studies [18]. Finally, our prior models were static in nature, and thus had to be rebuilt each time new data became available. While this may be tractable for simple models (which can be rebuilt very quickly), more complex models require more computational investment, and as such would become far too difficult to manage as the dataset grows. As an additional concern, there is reason to believe that the trends in our collected metrics are likely to change over time as societal, behavioral, and environmental changes occur, influencing DCM epidemiology [19], resulting in prior trends becoming obsolete or less significant. As such, an ideal model would be able to adapt to these changes as they arise, without the need of manual correction.

## 2.1 Data management

As previously mentioned, a key consideration in the clinical use of machine learning is that clinical data does not remain fixed. As new patients arrive and have their data collected and current patients see their disease state change, the relevant data that can be leveraged will change and expand over time. One possible approach is to retrain our machine learning model from scratch each time we update our dataset; this would become incredibly time and resource consuming as the dataset grows, however. Thankfully, advancements in *continual learning* in the last 5 years provide an elegant solution [20] (which we discuss in Section 3). To use these techniques effectively, we will need to consider the best way of optimizing how data is collected, stored, accessed, processed, and reported. Ideally, these data management systems should be malleable, extendable, and easy to use, so they may remain useful long-term in an ever-changing clinical environment. This section will focus on detailing methodologies for achieving this, accounting for the challenges presented by ongoing clinical data collection in the process.

### 2.1.1 Acquisition and storage

Ideally, our clinical dataset would include any and all relevant features that can be reliably and cost-effectively obtained. In reality, the specific data elements (or "features") will vary both across patients and over time (as new diagnostic tests come available or as ethical rules/constraints are updated). As such, an ideal data management approach should be capable of adapting to variable data feature collection over time, while still allowing new patients to be included. For ethical reasons, the storage system also needs to be set up so that data can be easily removed, should patients request their data be purged or if privacy rules require it.

In our facility, we addressed these considerations by creating a non-relational document database system using MongoDB. This allows for new features to be added and removed on-the-fly via a modular framework of 'forms', which specify sets of related features that should exist inside a single document 'type'. These documents can then be stored within a larger super-document (which we will refer to as a 'record') for each specific patient. This results in a large dataset containing all relevant features organized in an individual-specific manner. Each form acts as a 'schema', specifying what features can be expected to exist within each patient's record. With MongoDB, this allows features to be added and removed as needed without restructuring the entire database [21], which would risk data loss. If new features are desired, one can simply write a new form containing said features and add it to the system; previous entries without these new features can then be treated as containing "null" values in their place, thereby enabling them to still be included in any analyses performed on the data. Should features need to be removed, the form containing them can either be revised or deleted entirely. This results in the features effectively being masked from analysis access without deleting them from the database itself, allowing for their recovery in the future.

Our system also has the added benefit of allowing for the creation of 'output' forms, which capture and store metrics generated from data analyses. This enables the same system that collects the data to also report these analytical results back to the original submitter via the same interface. These output forms can also be stored alongside forms containing the original values that were provided to the analysis, making both easily accessible when calculating error/loss.

In our DCM dataset, all features (including MRI sequences) were collected at the time of diagnosis and consent to participate in our longitudinal registry associated with the Canadian Spine Outcomes and Research Network [22]. This registry collects hundreds of metrics on each patient, including a number of common diagnostic tests, with each being stored in the database as a single form. Most notably, this includes the modified Japanese Orthopedic Association (mJOA) scale form [23]. This is important for our study as we used this diagnostic assessment of DCM severity as the target metric for model training purposes. The MRI sequence form (which contains our MRI sequences alongside metadata associated with how they were obtained) and demographic information about the patient (including metrics such as name, age, and sex, among others) are also represented by one form each within our system. A simplified visualization of this structure can be seen in **Figure 1**.

This system can also allow pre-built structures to be re-created within it. For example, our MRI data is currently stored using the Brain Imaging Data Structure (BIDS) format [24]. This standardized data structure has directory hierarchies according to the contents of the file, with metadata describing the contents of the directory "falling through" to sub-directories and documents nested within it. These nested directories can then contain new metadata which overrides some or all of the previously set values, allowing for more granular metadata specification. Such a structure is conducive to our system, with said "nested" directories acting as features within forms, or forms within records; features could even consist of sets of sub-features (such as our MRI feature, which contains the MRI image *and* its associated metadata bundled together). Such nested structure can then specify "override" values, as they become needed.

## 2.2 Cleaning and preparation

The raw data collected in a clinical setting is almost never "analysis ready", as factors like human error and/or missing data fields must be contended with. Strategies for "cleaning" data can vary from dataset to dataset, but for precision medicine models there are some common standards. First, such protocols should work on a per-record basis, not a full-data basis. This is to avoid the circumstance where adding entries with extreme values would skew the dataset's distribution, compromising the model's prior training (as the input metrics are, in effect, rescaled), resulting in an unrealistic drop in model accuracy. Per-record internal normalization, however, typically works well, so long as it remains consistent over the period of the model's use. Some exceptions to this exist; for example, exclusion methods may need to be aware of the entire dataset to identify erroneous records. Likewise, imputation methods will need to "tap into" other available data to fill in missing or incorrect data points within each record.

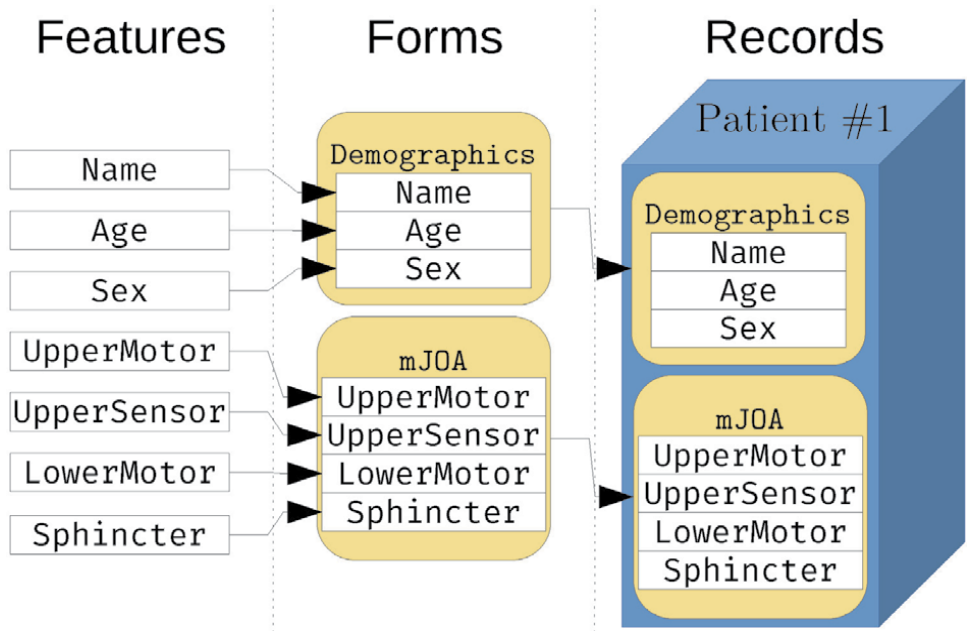

**Figure 1.**
*A simplified example of how data is stored and managed in our theoretical system. Each feature tracked is first bundled into a 'model', which groups related features together alongside a descriptive label. These models act as a schema for any data analysis procedures to hook into, and can be modified, removed, and created as needed. Model instances are then stored in 'records', which represent one entry for any analysis system which requires it (in our case, that of one patient enrolled in our DCM study). A data structure like this can be easily achieved with any non-relational database system; in our case, we opted to use MongoDB.*

It is often the case that data is obtained from multiple different sources (e.g. different clinics, practitioners, hospitals, labs, databases, etc.), which may have varying protocols and/or environmental differences that can structurally influence the resulting measurements. If the model could be retrained from scratch every time new data was obtained, these batch effects could be easily removed [25]. In iteratively trained systems, however, this would result in the same issue as full-data normalization; new entries causing fall-through changes in the entire dataset. However, under the assumption that batch effects have less influence on the data than 'true' contributing effects, it has been shown that systems which learn iteratively can integrate batch effect compensation directly into their training for both numeric [26] and imaging [27] metrics, thereby resolving the issue.

Coming back to our DCM example, our data consists of demographic information (which included a mix of numerical and categorical data), diagnostic data (also numerical and categorical), and 3-dimensional MRI sequences data (which also contains meta-data describing its acquisition method). For numerical and categorical data, our processing procedures are minimal, consisting of a quick manual review to confirm that all required features were present. As our dataset was relatively large, we opted to simply drop entries which contained malformed or missing data. New patient entries with errors were either met with a request to the supplier for corrected data, or had true values imputed for prediction purposes [28]. Categorical data is then one-hot encoded, while numerical data is scaled between 0 and 1 for values with know minimums and maximums. We had access to multiple different MRI sequencing methodologies as well, but focus on T2w sagittal oriented sequences based on our prior tests with the data [16]. MRI sequences are then resampled to a voxel size of $1mm^3$ and the signal values normalized to a 0 to 1 range.

Unlike our numerical results, this was done based on per-image signal minimum and maximum, in an attempt to account for variations in signal intensity variation, aiding in batch effect removal in the process.

## 3. Machine learning model design

Like the data management system, machine learning models designed for precision medicine need to be able to accept new data on an ongoing basis. The data contents may change over time as new discoveries about the illness are made, though it can be safely assumed that new data will be related to old data in some way. Contents of new data cannot be expected to be well distributed across target all target metrics. All of these requirements make precision medicinal systems a perfect use-case for continual learning systems.

Continual learning systems are characterized by their iterative training, as well as the ability to 'recall' what they learn from prior tasks to help solve new ones. Each of these tasks are assumed to be related, but contain non-trivial variation. This means the model must be flexible to change, while avoiding completely reconstructing itself after each new task, which could result in it 'forgetting' useful prior learning. These capabilities are referred to, respectively, as forward transfer (the ability to leverage prior learning to improve future analyses) and backward transfer (the ability leverage new knowledge to help with prior tasks).

Promising progress has been made in designing continual learning systems [20], to the point of a preliminary frameworks being devised to develop them. For this chapter, we will be using Fayek et. al's *Progressive Learning* framework [29] as a baseline reference, though some changes were made to account for precision medicine applications.

### 3.1 Initial network structure

All networks need to start somewhere, which for all intents and purposes acts like a classical static machine learning system. Neural networks are the system of choice for these processes, as they allow for multiple data types to be analyzed simultaneously, being able to be constructed in a modular fashion to match up with our data storage structure detailed prior. Depending on the data, what this entails will differ. For data with implicit relations between features (such as MRI images with their spatial relations), Convolutional Neural Network (CNN) systems have been shown to be extremely effective [30]. CNNs are also among the most computational efficient neural networks to train and run [31, 32], making them ideal for low resource systems. For other data, a Densely Connected Learning Networks (DCLNs) may be more appropriate. The complexity of these networks can be tuned to fit the problem. These models tend to be over-parameterized, however, potentially causing them to "stick" in one place or over-fit to training data; this is mediated somewhat via model pruning, discussed later in this section. The choice of available models is ever-changing, however, so one should find the model structure which best fits their specific case.

For progressively learning models, one further constraint exists; it must be able to be cleanly divide its layers into 'blocks'. As discussed in Fayek et. al's framework [29], this is necessary to allow for the model to progress over time. How these blocks are formed can be arbitrary, so long as each block is capable of being generalized to accept data containing the same features, but of different shape (as the size of the input data grows resulting from the concatenation operation discussed later in this section). One should also keep in mind that the block containing the output layer will be reset every progressive iteration, and should be kept as lightweight as possible.

For DCM, this would be accomplished via multiple layers running in parallel. For MRI inputs, being 3D spatial sequences, something like a 3D DenseNet similar to that employed by Ke et al. [33] could be used. The DenseNet could be run alongside DCLN blocks in parallel to read and interpret our linear data (demographics, for example), grouped with the DenseNet blocks to form the initial progressive learning model. A diagram of this structure, using the same structure mentioned prior (**Figure 1**), with a simplified "MRI" model present, is shown in **Figure 2**.

For the purposes of comparison with the original *Progressive Learning* framework, however, our testing system will instead use their initial model structure [29].

### 3.2 Iterative training data considerations

Once this initial framework is in place, it then needs to prove capable of accepting new patient data, updating itself as it does so. Given that the measurements of patients enrolling in clinical illness studies can be sporadic in terms of when and how often they are made, data for this system will need to be collected over time until a sufficiently large 'batch' of new records is acquired. Ideally large batches would be collected which are sizable enough to be split into multiple smaller batches, allowing for curriculum formation as detailed in the subsequent section. In many cases this is simply not feasible due to the time required to obtain such large batches. In this circumstance, each batch acts as a single 'curriculum' provided to our network in effectively random order. Thankfully, the curriculum stage appears to be the least significant stage of the Progressive Learning framework [29]. The size of these batches will depend heavily on how much data one expects to be able to collect in a given period of time and how regularly one wishes to update the model. For categorical data, each batch should include at least two of every category (one for testing, one for validation), which may influence how many samples one needs to acquire. We recommend a slightly larger batch sizes when linear data is brought into the fold, to account for the increased variety. With our DCM dataset, with a categorical output metric (the mJOA-derived DCM severity class, consisting of 4 classes), a batch of 20 patient records was selected. Data augmentation of this *new* data can also be utilized to increase the number of effective records being submitted. However, one should avoid using data from records in previous training cycles, as it can lead to a the model failing to adopt novel data trends in newer results.

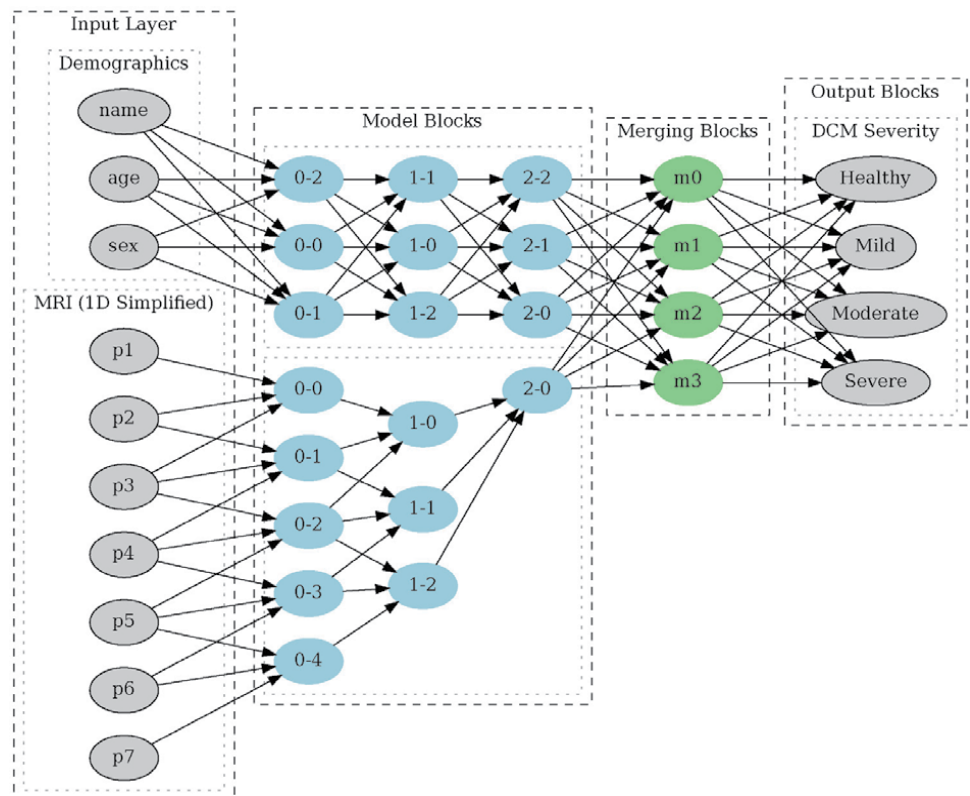

**Figure 2.**
*An example of the initial neural network structure for use in precision medicinal systems. Note that each form receives its own "branch" block (presented within the model block column) which is used to interpret the form's contents. As a result, each branch's structure can be tailored to suit the form's contents, allowing for modular addition or removal of model's feeding into the network's design as needed. The results of each of these branches' interpretations are then submitted to a set of "merging" blocks, which attempts to combine these results together in a sensible manner, before a final "output" layer reports the model's predictions for the input. The output layer is also modular, allowing for extension and/or revision as desired.*

### 3.3 Continual learning

Here, we will focus on detailing a framework based on Fayek et. al's Progressive Learning Framework [29], which consistes of three stages; *curriculum*, *progression* and *pruning*.

#### 3.3.1 Curriculum

Given sufficiently large batches of new data can be collected in a timely manner, one can utilize the curriculum stage; at least three times the number of records per batch being collected in a 6 month period seems to be a good cutoff for this, though this can differ depending how rapidly one expects disease trends to change. This stage, as described in Fayek et. al's framework [29], is composed of two steps; curricula creation and task ordering. In the creation step, the batch is split into sub-batches, with each being known as a 'curriculum'. How this is done depends on the data at hand (i.e. categorical data requires that each curriculum contains data from each category), but can otherwise be performed arbitrarily. Once these curricula are formed they are sorted based on an estimate of how "difficult" they are, from least to most. Difficulty estimation can be as simple as running a regression on the data and using the resulting loss metric. The sorted set of curricula are then submitted to the network for the progression and pruning stages, one at a time. This allows for the network to learn incrementally, picking up the "easier" trends from earlier curricula before being tasked with learning more "difficult" trends in later ones.

In precision medicine, however, collecting sufficient data in a useful time span is often not possible. In this case, this stage can be safely skipped; the smaller batches will simply act as randomly sampled, unordered curricula. How "large" this is depends on the batch size selected earlier; one should collect at least three batches worth to warrant the additional complexity of the curriculum stage. For our DCM setup, we fell below this level, as we intend on updating as often as possible, and as such intended on utilizing new batches as soon as they were collected. We believe that in most precision medicine examples this is likely to be the case, though in some situations (such as the ongoing COVID-19 pandemic), the scale of patient data collection may make the curriculum stage worth considering. Employing few-shot learning techniques may also allow for smaller subsets of data to form multiple batches as well, though the efficacy of such procedures has yet to be tested in this context.

### 3.3.2 Progression

In this stage, new blocks of layers are generated and concatenated to previously generated blocks in the model. The new input-accepting block is simply stacked adjacent to the prior input blocks in the model, ready to receive input from records in our dataset. Each subsequent new block, however, receives the concatenated outputs of *all* blocks from the prior layer, allowing it to include features learned in previous training cycles. The final block, which contains the output layer, is then regenerated entirely, resulting in some lost training progress that is, thankfully, usually quickly resolved as the model begins re-training.

The contents of these added blocks depends on the desired task and computational resources available. Large, more complex blocks require more computational resources and are more likely to result in over-fitting, but can enable rapid adaption of the network and better forward transfer. In the original framework [29], these blocks were simply copies of the original block's architecture, but reduced to approximately half the parameters. However, one could instead cycle through a set of varying block-types, based on how well the model performed and whether new data trends are expected to have appeared. They could also be changed as the model evolves and new effective model designs are discovered, though how effective this is in practice has yet to be seen.

Once these blocks are added, the network is then retrained on the new batch of data, generally with the same training setup used for the original set of blocks. During this retraining, prior block's parameters can be frozen, locking in what they had learned prior while still allowing them to contribute to the model's overall prediction. This prevents catastrophic forgetting of previously learned tasks, should they need to be recalled, though this usually comes at the cost of reduced overall training effectiveness. However, if one does not expect to need to re-evaluate records which have already been tested before, one can deviate from Fayek's original design and instead allow prior blocks to change along with the rest of the model. An example of progression (with two simple DCLN blocks being added) is shown in **Figure 3**.

For our DCM data, this is a pretty straightforward decision. New blocks would simply consist of new 3D DenseNet blocks run in parallel to simple DCLN layers, both containing approximately half the parameters as the original block set. The output block is then simply a linear layer which is fed into a SoftMax function for final categorical prediction. As we do not expect prior records to need to be re-tested, we also allow prior blocks to be updated during each training cycle.

### 3.3.3 Pruning

In this stage, a portion of parameters in the new blocks are dropped from the network. What parts of the model are allowed to be pruned depends on how the prior progression stage was accomplished; if previously trained blocks have been frozen, then only newly added elements should be allowed to be pruned to avoid catastrophic loss of prior training. Otherwise, the entire model can be pruned, just as it has been allowed to be trained and updated. The pruning system can also vary in how it determines which parameters are to be pruned, though dropping parameters with the lowest absolute value weights is the most straightforward. These can also be grouped as well, with Fayek et al. choosing to prune greedily layer-by-layer. However, we have found that considering all parameters at once is also effective. The proportion $q$ dropped per cycle will depend on the computational resources and time available. Smaller increments will take longer to run, whereas larger values will tend to land further away from the "optimal" state of the network. An example of the pruning stage is shown in **Figure 4**.

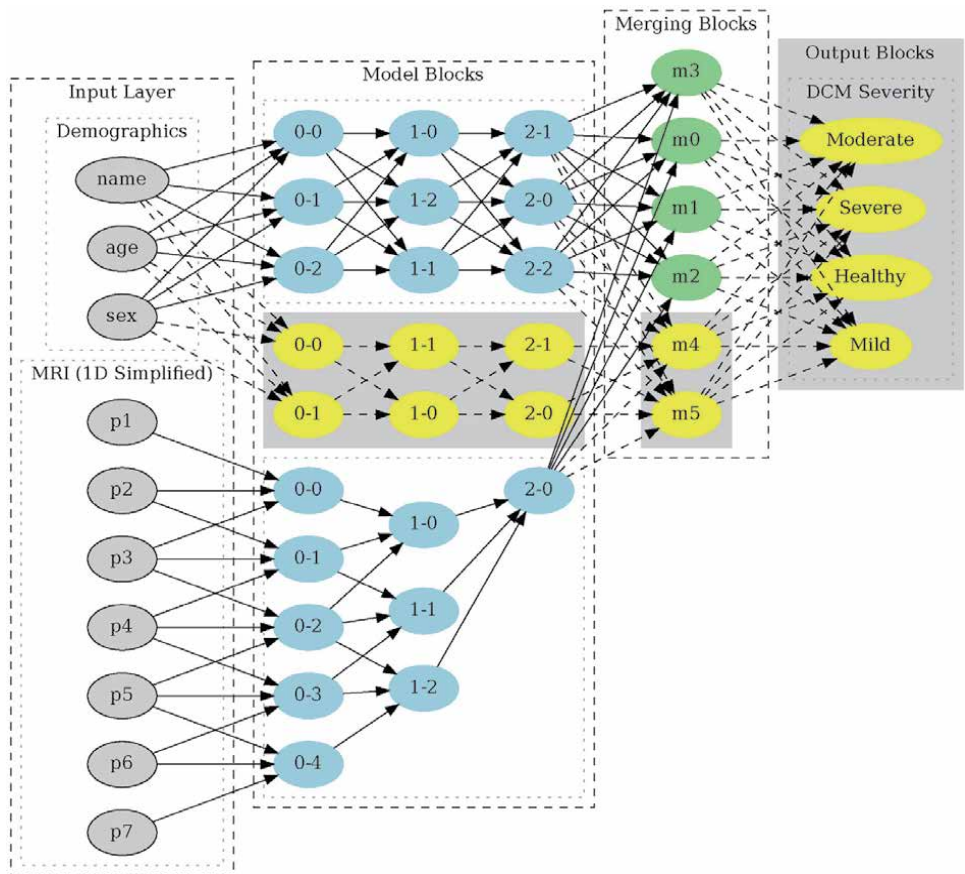

**Figure 3.**
*An example of the progression stage, building off of the initial model shown in **Figure 2**. New nodes are contained within the gray boxes, with hashed lines indicating the new connections formed as a result. Note that input connections are specific to each form, only connecting to one's inputs (in this case, only the Demographic's input), and not to those in the other branches (such as the MRI branch); this allows for shortcomings in particular model's contributions to be accounting for independently, without an extreme growth in network complexity. Note as well that the merging layer (representing all non-input receiving blocks) forms connections with all prior block outputs, however, regardless of which forms have received a new connected block. The entire output block is also regenerated at this stage, providing some learning plasticity at the expense of initial learning.*

The network, now lacking the pruned parameters, is then retrained for a (much shorter) duration to account for their loss. In Fayek et al's example, this process is then repeated with progressively larger $q$ proportions until a loss in performance is observed. Alternatively, one can instead repeatedly drop the same percentile of parameters each cycle from the *previously pruned* network. This has the benefit of reducing the time taken per cycle slightly (the same weights do not need to be pruned every cycle), while also leading to the total proportion of the pruned model increasing more gradually per cycle, improving the odds that the model lands closer to the true optimal size for the system. This pruning system has the potential to be much slower, however (should the rare circumstance occur where all the new parameters are useless, requiring more iterations overall). As such, in time-limited systems, Fayek's approach remains more effective.

This stage also allows for, in theory, dynamic feature removal. Should a model (or feature within said model) cease to be available, one can simply explicitly prune the parameters associated with that feature, in effect performing a targeted pruning cycle. One would need to re-enable training of previously trained nodes to account for this, however, leading to the possibility of reduced backward transfer. Depending on how significant the to-be-removed features have become in the network, this may need to be done over multiple pruning cycles; this should allow the network to adapt to changes over time, reducing the risk of it getting 'stuck' in a sub-optimal state.

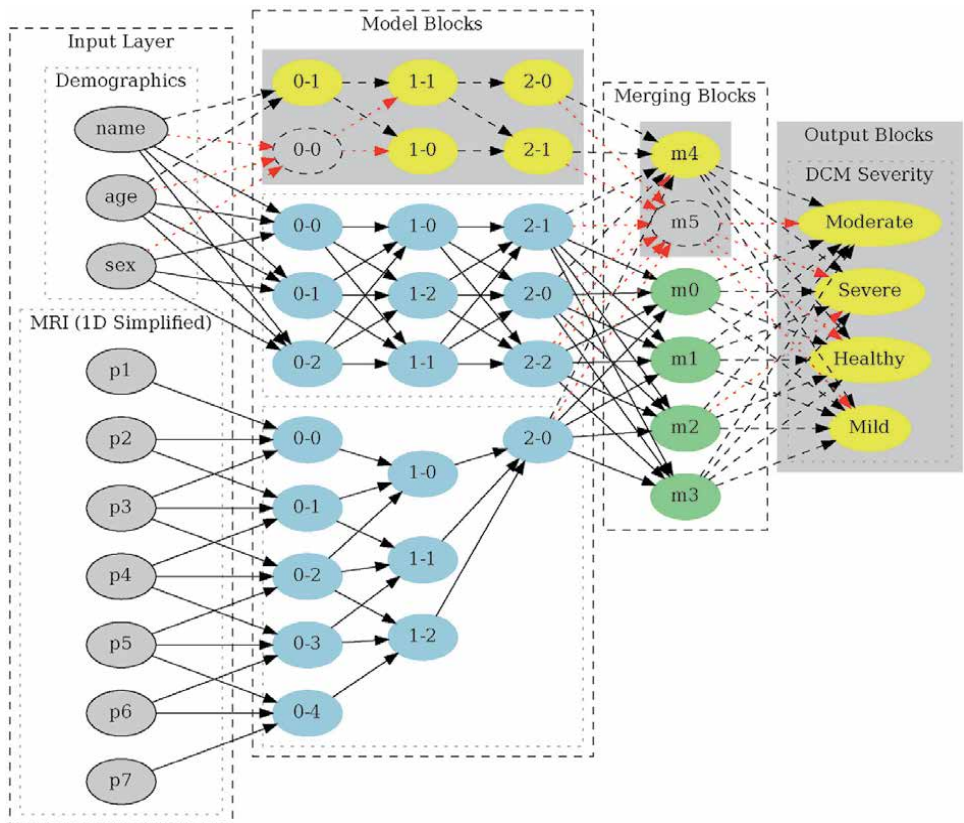

**Figure 4.**
*An example of the pruning stage, building off of the progression network shown in **Figure 3**. Note that only newly formed connections are targeted for pruning by default, with pre-existing connections remaining safe. Parameters themselves can also be effectively lost entirely (shown as nodes with no fill and a dashed outline) should all connections leading into them get removed. This results in all connections leading out of them also getting pruned by proxy.*

For our DCM data, the complexity of the illness and scope of the data makes it extremely unlikely for a worst-case pruning issue to occur. As such, a 10% lowest absolute weight pruning system, applied globally, is selected as our starting point, iteratively applied until a loss of mean accuracy over 10 post-prune correction epochs is observed.

## 4. Protocol assessment

### 4.1 Progressive learning ConvNet

#### 4.1.1 Methodology

To confirm our protocol functions effectively in practice, we replicated the CIFAR-100 analysis used in the original *Progressive Learning* paper [29], with a few major changes to replicate a precision medicine environment (i.e. the kind of clinical context in which data is typically collected). First, only the CIFAR-10 dataset was used [34], rather than it being used as the model initialization dataset. This was done to better reflect the clinical environment, where we are unlikely to have a pre-established dataset which is known to be effective at preparing our model for the task. The 10 categories of the CIFAR-10 dataset also represent the average granularity usually used to assess many illnesses. Second, our datasets were randomly split into 10 subsets of 6000 random images, with 5000 used for training and the remaining 1000 used for validation. The contents of these subsets was completely random, allowing for imbalance in the number of elements from each of the 10 categories, reflecting how data collected in a clinical setting could occur. Third, we skipped the curriculum stage, again to reflect the circumstances of clinical data collection (wherein the scale of collection is insufficient). Fourth, our framework was implemented in PyTorch [35] rather than TensorFlow [36] due to its more robust network pruning support. Finally, data augmentation was performed on each image to both discourage the model from memorizing data and to simulate human error/variation in clinically acquired data. This results in a problem which is slightly more difficult than the original setup devised by Fayek et al., though for parity sake, we continued to use the same Convolution Network design.

We tested 6 procedures, representing combinations of two different variations. The first variation was whether the learning model trained as an independent learning model, a progressive learning model with prior blocks frozen, or a progressive learning model with prior blocks being freely pruned and updated. For independent learning procedures, the model was completely reset after each training cycle, whereas for progressive learning procedures the model persisted across cycles (allowing for it to "apply" prior knowledge to new data). The second was whether data was provided in batches (similar to a clinical setting), or submitted all at once (the "ideal" for machine learning analyses). In batched procedures, data was submitted one subset at a time, as described prior. A strict max wall time of 8 hours was put in place for all protocols to simulate the limited resources (in both time and hardware) that clinical settings often have. All protocols were run on a single Tesla V100-PCIE-16GB GPU with 16GB of RAM and two Intel(R) Xeon(R) Gold 6148 CPUs run at 2.40GHz (speeding up initial protocol setup).

The initial architecture for all procedures is shown in **Table 1**. For progressive learning procedures, new blocks were added which were half the size of the original blocks, set to receive the the concatenated outputs of all blocks in the prior layer of each set of blocks. All parameters were initialized randomly using PyTorch version 1.8.1 default settings. We used an ADAM optimizer with a learning rate of 0.001, first moment $\beta_1$ of 0.99, second moment $\beta_2$ of 0.999, and weight decay $\lambda$ of 0.001 during training. For progressive learning models, an identical optimizer with one tenth the learning rate was used for post-pruning model optimization. Each cycle consisted of 90 epochs of training. Progressive procedures were given 10 epochs per pruning cycle, with pruning being repeated until the mean accuracy of the prior set of epochs was greater than that of the new set of epochs, with the model's state being restored to the prior before continuing. The model's training and validation accuracy was evaluated and reported once per epoch. Protocol efficacy was measured via the max validation accuracy of the model over all cycles and epochs and mean best-accuracy-per-cycle (BAPC) for all cycles.

*4.1.2 Results*

For our full datasets, the model achieved diagnostic classification accuracy values of 80-85% for most of the results. The simple model, without progression and with full access to the entire dataset, reached a max accuracy of 81.13%, with a mean BAPC of 80.71%. Adding progression to the process further improved this, primarily though the pruning stage, with a max accuracy of of 84.80%. However, the mean BAPC dropped to 77.44%, as prior frozen parameters in the model appeared to make the model "stagnate". Allowing the model to update and prune these carried-over parameters improves things substantially, leading to a max accuracy of 90.66% and a mean BAPC of 84.83%.

| Block Number | Type | Size | Other |
|---|---|---|---|
| 1 | 2DConvolution | 32, 3x3 | Stride = 1 |
| | 2DBatchNorm | | |
| | ReLU | | |
| | [*Concatenation*] | | |
| 2 | 2DConvolution | 32, 3x3 | Stride = 1 |
| | 2DBatchNorm | | |
| | ReLU | | |
| | 2DMaxPooling | 2x2 | Stride = 2 |
| | Dropout | | r = 0.25 |
| | [*Concatenation*] | | |
| 3 | 2DConvolution | 64, 3x3 | Stride = 1 |
| | 2DBatchNorm | | |
| | ReLU | | |
| | [*Concatenation*] | | |
| 4 | 2DConvolution | 64, 3x3 | Stride = 1 |
| | 2DBatchNorm | | |
| | ReLU | | |
| | 2DMaxPooling | 2x2 | Stride = 2 |
| | Dropout | | r = 0.25 |
| | [*Concatenation*] | | |
| 5 | Flatten | | |
| | Linear | 512 | |
| | 1DBatchNorm | 512 | |
| | ReLU | | |
| | Dropout | | r = 0.5 |
| 6 | [*Concatenation*] | | |
| | Linear | 20 | |
| | Softmax | | |

**Table 1.**
*The basic structure of the convolutional neural network being tested on the CIFAR-10 dataset. Based on the model used by Fayek et al. [29]. [Concatenation] indicates where the output of one set of blocks would be concatenated together before being fed into new blocks in the following layer, and can be ignored for independent learning tasks.*

When data was batched, a noticeable drop in accuracy was observed, as expected. Without progressive learning, our model's max observed accuracy was only 73.7% (a drop of 7.37%), with a mean BAPC of 71.75%. The progressive model with frozen priors initially performed better, reaching its maximum accuracy of 75.9% in its first cycle, but rapidly fell off, having a mean BAPC of 66.0%. Allowing the model to update its priors greatly improve the results, however, leading to a maximum accuracy of 82.4% and a mean BAPC of 79.02%, competing with the static model trained on all data at once.

A plot of each model's accuracy for each model setup, taken over the entire duration (all cycles and epochs) for both the training and validation assessments, is shown in **Figure 5**.

## 4.2 DenseNet

### 4.2.1 Methodology

To confirm that the success of the setup suggested by Fayek et al. was not due to random chance, we also applied the technique to another model which is effective at predicting the CIFAR-10 dataset; the DenseNet architecture [37]. DenseNets are characterized by their "blocks" of densely connected convolution layer chains, leading to a model which can utilize simpler features identified in early convolutions to inform later layers that would, in a more linear setup, not be connected together at all. These blocks are a perfect fit for our method, as they can be generated and added to our progressive learning network just like any other set of layers. DenseNets have also been shown to have better accuracy than classical convolution nets within the CIFAR-10 dataset, reaching error rates of less than 10% in many cases [37]. However, the dense connections make the networks extremely complex, and they are generally highly over-parameterized as well, making them prone to over-fitting in some cases.

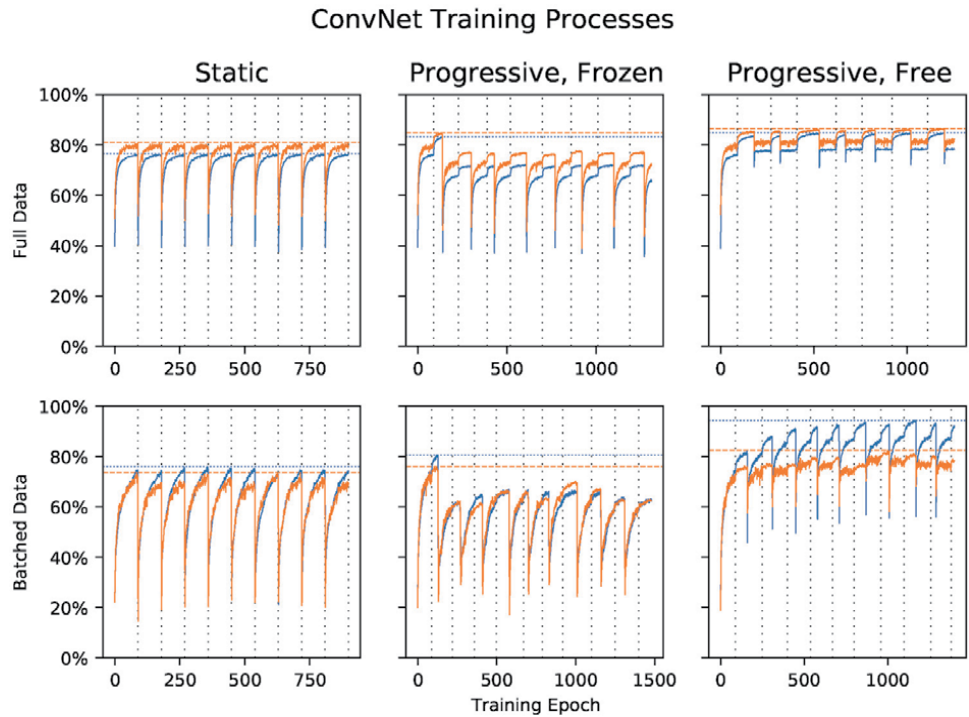

**Figure 5.**
*The training progression of the ConvNet model replicated from Fayek et. al's study [29] in various forms. From left to right, the model on its own, reset after every cycle (static), a progressively learning model, with prior traits frozen (progressive, frozen), and a progressively learning model, with all traits open to training and pruning each cycle (progressive, free). Training accuracy is shown in blue, with validation accuracy shown in orange. The maximum observed accuracy for each is indicated via a horizontal dotted (training) or dashed (validation) line. The dotted horizontal lines indicate where the training of the model for a given cycle was complete (not including the pruning of progressive models). Note that the total number of epochs taken between these cycles differs from cycle to cycle in progressive models, as a result of the pruning stage cycling until a validation accuracy loss was observed.*

Our testing methodology was largely identical to that of the Convolutional network tested in the previous section. One change was to use a Stochastic Gradient Descent (SGD) optimizer with an initial learning rate of 0.1. Training was done in batches of 64, for a total of 300 epochs per cycle. The learning rate is reduced by a factor of 10 when 50% and 75% of the epochs for each cycle has passed. The SGD optimizer was set with a weight decay of 0.0001 and a momentum of 0.9. Dropout layers with a drop rate of 0.2 were added after each block as well. The initial architecture for the network was based on the 'densenet-169' architecture, and is shown in **Table 2**, having a growth rate of 32, an initial feature count of 64, and consisting of 4 blocks of densely connected convolution layers, each with 6, 12, 32, and 32 convolution layers respectively. For progressive learning systems, new blocks followed the same architecture with half the growth rate (16) and initial features (32). These changes were made to maintain parity with the original DenseNet CIFAR-10 test [37].

| Block Number | Type | Size | Other |
|---|---|---|---|
| 1 | 2DConvolution | 64, 7x7 | Stride = 2, Padding = 3 |
| | 2DBatchNorm | | |
| | ReLU | | |
| | 2DMaxPooling | 3x3 | Stride = 2, Padding = 1 |
| | [Concatenation] | | |
| 2 | Dense Block | Layers = 6 | Bottleneck = 4, r = 0.2 |
| | Transition Block | | |
| | [Concatenation] | | |
| 3 | Dense Block | Layers = 12 | Bottleneck = 4, r = 0.2 |
| | Transition Block | | |
| | [Concatenation] | | |
| 4 | Dense Block | Layers = 32 | Bottleneck = 4, r = 0.2 |
| | Transition Block | | |
| | [Concatenation] | | |
| 5 | Dense Block | Layers = 32 | Bottleneck = 4, r = 0.2 |
| | Transition Block | | |
| | [Concatenation] | | |
| 6 | 2DBatchNorm | 1664 | |
| | [Concatenation] | | |
| 7 | ReLU | | |
| | 2DAdaptiveAveragePool | 1x1 | |
| | Flatten | | |
| | Linear | 1664 | |

**Table 2.**
*The structure of the DenseNet mdoel being tested on the CIFAR-10 dataset. Based on the model used by Huang et al. [37]. Dense block indicates a densely connected convolution block, with transition block indicating a transition layer, both being detailed in Huang et. al's original paper. [Concatenation] indicates where the output of one set of blocks would be concatenated together before being fed into new blocks in the following layer, and can be ignored for independent learning tasks. Where it appears, r indicates dropout rate for the associated block.*

### 4.2.2 Results

For our full datasets, we saw an accuracy values of around 90%. The simple model, without progression, reached a max accuracy of exactly 90%, but was only able to run one cycle to completion before the 8 hour time limit was reached. Adding progression to the process improved this slightly, resulting in a max accuracy of of 90.66%, but only barely completed its first pruning cycle before the time limit was reached. As a result, the same accuracy was observed for both progressive models with and without priors being trainable, as no new prior blocks were added. Slight variations were still observed, however, due to how the model's initialization process differs.

When data was batched, a much more significant drop in accuracy occured as compared to the Convolutional network. Without progressive learning, our model's max observed accuracy was only 69.9% (a drop of more than 30%), with a mean BAPC of 65.87%. However, it was able to run for all 10 cycles within the allotted

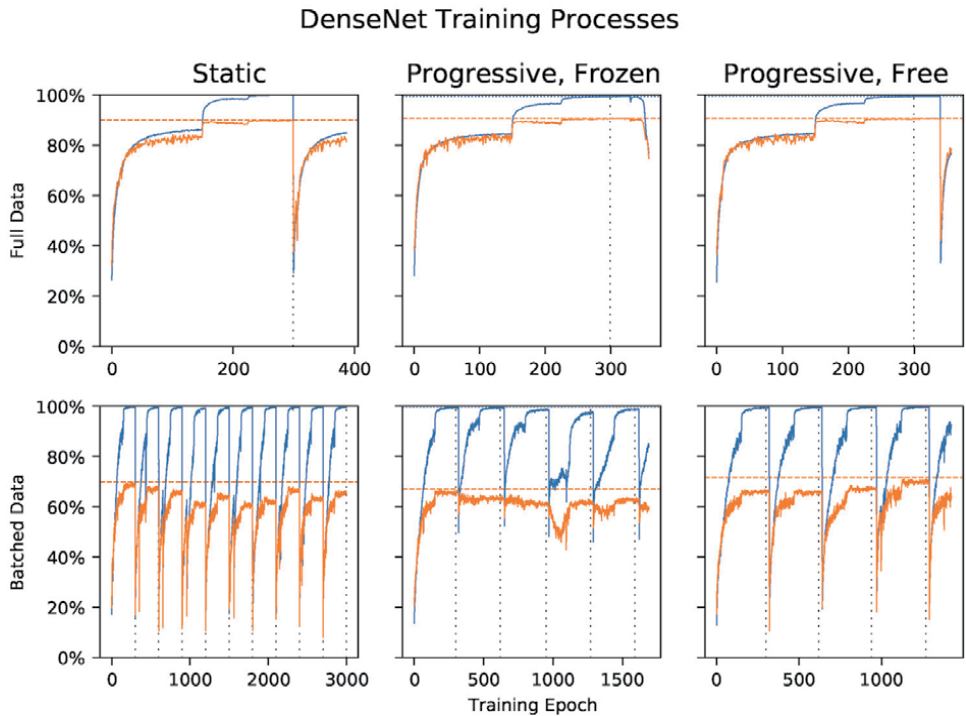

**Figure 6.**
*The training progression of the DeepNet model replicated from Huang et. al's original 'densenet-169' model [37] in various forms. From left to right, the model on its own, reset after every cycle (static), a progressively learning model, with prior traits frozen (progressive, frozen), and a progressively learning model, with all traits open to training and pruning each cycle (progressive, free). Training accuracy is shown in blue, with validation accuracy shown in orange. The maximum observed accuracy for each is indicated via a horizontal dotted (training) or dashed (validation) line. The dotted horizontal lines indicate where the training of the model for a given cycle was complete (not including the pruning of progressive models). Note that the total number of epochs taken between these cycles differs from cycle to cycle in progressive models, as a result of the pruning stage cycling until a validation accuracy loss was observed.*

8 hour time span. The progressive model with frozen priors performed even worse, reaching a maximum accuracy of 67.1% in its first cycle, and only completing 5 cycles before the time limit, only reaching a mean BAPC of 64.28%. Allowing the model to update its priors somewhat improved the results, leading to a maximum accuracy of 71.7% and a mean BAPC of 68.28%, showing some slight recovery over the static model in a batch scenario. However, it also only managed to run through 5 cycles before the time limit was reached.

A plot of each model's accuracy for each model setup, taken over the entire duration (all cycles and epochs) for both the training and validation assessments, is summarized in **Figure 6**.

## 5. Discussion and conclusions

The results of our tests show great promise for how these approaches to machine learning use in precision medicine can be used, while nonetheless highlighting some significant shortcomings which will need to be considered should this framework become common practice. Most notably, we see that model's which over-fit the available data are extremely detrimental to the system, even if the underlying model would be better with all data immediately available to it. This is shown very clearly with the effectiveness of pruning in all our models, with clear gains in accuracy observed, likely as a result of the process helping counteract over-fitting resulting from over-parameterized models. Finding an "ideal" model for a given task is already a difficult task, and our results show that this is only exacerbated by the conditions of a clinical environment. Nevertheless, there is clearly potential in this framework, with the Convolutional network tested on clinical-like batch data being near identical in effectiveness to its static counterpart trained on the full dataset.

We are also optimistic that many opportunities remain for improvement in progressive learning implementations. Our current implementation of the progressive learning framework is locked to a specific set of initial data inputs, being unable to add new ones should they become available. In theory, this could be as simple as adding a new set of initial blocks to the existing network, in effect acting like a "progression" stage with custom new blocks (as well as an update to existing new block generation procedures to match). However, this has a number of issues that we have not, at present, found a way to resolve. First, each branch is likely to "learn" at different rates, resulting in one set of blocks associated with a given set of input data containing more redundant features per-progression stage than the rest. This proves problematic during pruning, however; we either over-prune blocks with important features within them, or under-prune those which contain an abundance of redundant and/or noise-contributing features. We believe this can be resolved, but were simply unable to do so by the time of this publication.

Another potential improvement would be to "carry-over" output layers weights between progression stages. This would allow for the network to have better forward transfer, so long as the task's end goal (categorical prediction, single metric estimation etc.) remains the same. In our implementation, this is currently not the case, with the output layer being regenerated every cycle, keeping it in line with the original Progressive Learning framework's design [29]. The difficulty of implementing such as system, as well as its effectiveness in improving model outcomes, has yet to be tested.

One other major hurdle is that of long term memory cost. As currently implemented, pruning does not actually remove parameters from the model; it simply masks them out during training and evaluation, preventing them from contributing to the models predictions. While this improves the speed and accuracy of the model being generated, its memory footprint expands infinitely as more cycles are run. Resolving this issue is difficult, however, requiring the model to effectively fully re-construct itself to account for any now-removed parameters. Doing so would allow the model to come to a "static" memory cost, as the number of pruned parameters approaches the number of new ones added every cycle. In turn, this would enable applications where the model is expected to existing for very long duration in limited resource systems. Such compression techniques are an ongoing field of research at time of writing; as such, we believe such a implementation will be possible in the near future.

Finally, testing our methodology on a real-world clinical dataset is needed before we can be sure it is truly effective. While the CIFAR-10 dataset [34] has been shown to work effectively for machine learning testing purposes, our assumptions about clinical data still need to be confirmed. We intend to put our framework to the test soon, assessing its effectiveness at predicting DCM severity using the DCM data mentioned throughout this chapter; nevertheless, this framework should be considered experimental until such results (from ourselves or others) are acquired. Continual learning systems trained for clinical data also retain the limitations of continual learning, such as increased potential to over-fit and the inability to transfer new knowledge obtained to help with the understanding of prior knowledge. Modifications to the progression procedure have been proposed to amend this [29], though these have not been tested at time of writing.

Overall, however, we believe our framework for machine learning system design in precision medicine should work well as a road-map for future research, even though refinements remain to be made. With systems such as the Progressive Learning framework available, these new systems can adapt to changes in data trends while accepting new data in effectively random batches, both important requirements for a clinical environment. Well designed data storage and management also allows such systems to easily access, update, and report important metrics to all necessary parties, while remaining open to changes as new research is completed. Through the application of these techniques, modern medicine should be able to not only adapt to the age of information, but to benefit immensely from it.

## Abbreviations

| | |
|---|---|
| DCM | Degenerative Cervical Myeolopathy |
| CIFAR | Canadian Institute for Advanced Research |
| mJOA | Modified Japanese Orthopedic Association |
| BIDS | Brain Imaging Data Structure |
| CNN | Convolutional Neural Network |
| DCLN | Deeply Connected Learning Network |
| BAPC | Best Accuracy Per Cycle |
| SGD | Stochastic Gradient Descent |

## Author details

Kalum J. Ost[1,2], David W. Anderson[2†] and David W. Cadotte[1,2,3,4*†]

1 Hotchkiss Brain Institute, Calgary, Alberta, Canada

2 Cumming School of Medicine, Calgary, Alberta, Canada

3 Division of Neurosurgery, Departments of Clinical Neurosciences and Radiology, University of Calgary, Alberta, Canada

4 Combined Orthopedic and Neurosurgery Spine Program, University of Calgary, Alberta, Canada

*Address all correspondence to: david.cadotte@ucalgary.ca

† These authors contributed equally.

IntechOpen

# Enhancing Program Management with Predictive Analytics Algorithms (PAAs)

*Bongs Lainjo*

## Abstract

The increase in the amount of big data and the emergence of analytics technologies has created the opportunity for applying algorithm development techniques using machine learning (ML) languages to predict future events. To conduct inclusive analyses of contemporary literature of existing relevant narratives with a focus on program management themes, including state-of-the art methodologies on current plausible predictive analytics models. The methodology used is the review and applications of relevant programming platforms available. Program management requires the utilization of the existing ML languages in understanding future events. Enabling decision makers to make strategic - goals, objectives, and missions. The use of PAAs has gained thematic significance in automotive industries, energy sector, financial organizations, industrial operations, medical services, governments, and more. PAAs are important in promoting the management of future events such as workflow or operational activities in a way that institutions can schedule their activities in order to optimize performance. It enables organizations to use existing big data to predict future performance and mitigate risks. The improvements in information technology and data analytics procedures have resulted in the ability of businesses to make effective use of historical data in making predictions. This enables evidence-based planning, mitigating risks, and optimizing production.

**Keywords:** program-management, algorithm-development, predictive-analytics-models, machine-learning, evidence-based-planning, optimizing

## 1. Introduction

This chapter examines the current knowledge and scholarly information about predictive analytics algorithms (PAAs) by focusing on the concept of working principles on which they are used to predict future events and the procedures followed in creating them. The PAAs have been used extensively in predicting future events in healthcare practice, manufacturing companies, businesses, education, sports, and agriculture. The main programming languages used to create PAAs are Java, C, and Python amongst others. The forms of algorithms that are commonly used are brute force algorithm, simple recursive algorithm, backtracking algorithm, randomized algorithm, and dynamic programming algorithms.

## 2. Background

Over the years, the concept and principles of data management have remained mostly unchanged. What has changed, however, includes the introduction of a complex, state-of-the-art, sophisticated, and integrated technological ecosystem: big data, cloud computing, and analytics [1]. The dynamics of this system have moved the way data are managed to a higher level, and institutions (public, private, sports, healthcare, and more) have capitalized on this! They have maximized their respective productivity levels using these systems with no reservations. As expected, these innovative developments come with significant risks from reliability to privacy and security concerns. Data are only as good and useful as their level of validity and reliability. Analytics, mentioned earlier, is one of the major components of the ecosystem that is used in transforming data into information. It is a sub-system that is also as useful as the reliability of the data used in performing different analytical interventions. At the conceptual level, analytics is an algorithm-driven strategy [2]. It facilitates the transformation of complex (generally) historical data sets into meaningful outcomes used for predicting future events. Its effectiveness has transformed and refined different sets of intended results. Institutions have used its predictive capabilities to optimize resources, streamline activities and increase productivity—ultimately becoming more competitive. The key players involved in the management and utilization of these ecosystems are the service providers (SPs) and their clients (users) [3].

It has been difficult for equipment manufacturers to develop innovative products using hardware alone. Those involved in product development have been able to add capabilities by applying solutions that improve customer satisfaction and value creation. Predictive analytics programs and equipment have been effective in promoting the anticipation of failures and provide forecasts for energy requirements while reducing the cost of operations. Predictive analytic models are used by companies in developing forecasts and creating plans for better utilization of resources. Before PAAs are used, the developer must review the available data and create/test mathematical models that incorporate computational processes in predicting future outcomes. The models provide forecasts of future outcomes based on a particular metric such as the associated parameter changes.

This chapter looks at the scope, thematic applicability, challenges, and prognoses of predictive analytics with life case studies from different institutions. It also highlights limitations, implications, and potential vulnerabilities. In this study, a select number of key institutions are included. These serve as examples of classical life case studies meant to help readers resonate with their own different and unique challenges. The various organizations are reviewed and analyzed on multi-dimensional thematic platforms. These include problem statements, strategic approaches, relevant processes, algorithmic layouts, programming descriptions, pilot testing, process reviews, initial implementation, and challenges and lessons learned. The relevant contents of these themes are only limited by the inability to access reliable, valid, evidence-based, useful, and compelling sources of information. Every attempt is made to address these limitations, and at the same time, prioritize available sources based on their pragmatic perspectives, simplicity, and authenticity. The select institutions include business (e-commerce, banking, finance, marketing, and more), health, education, government, sports, agriculture, social media, and so on. One invaluable approach applied in developing this narrative is an extensive review of available and contemporary literature. While the topic remains new and evolving, available documentation does indicate an inclusive degree of participation by different stakeholders. Key limitations like technical inability to develop and implement the various models have not been a significant deterrent. Readers need to consider this chapter as an evidence-based, knowledge-sharing cursory or soft-core and easy to understand demonstration of the strength, scope, and application of PAAs in optimizing program management challenges.

## 3. Quality of data (QOD)

My experience dealing with data of different types and categories spans over four decades. From attending a survey technician-training program after high school to studying in an engineering school, data management has played and continues to play a very significant role in my professional life. As well, the challenges encountered over this period of time continue to evolve exponentially! The most recent paradigm transformation in data management is in the proliferation of analytics — a domain that has enabled businesses, industry, academia, banks, etc. to exhale and address competing forces with might and vitality.

One adage that strongly and appropriately describes different forms of data is "garbage in garbage out" (GIGO). Interestingly, this adage is not just limited to conventional data as described in the previous paragraph—it also includes a human dimension. For example, healthy eating habits correlate positively with an improved quality of life and health.

The importance and significance of good data cannot be adequately emphasized in general, and more specifically and critically in data-intensive methodologies like analytics.

Here is a personal and professional life case study example. In 1992, Columbia University (CU) recruited me as a Senior Data Management Advisor. My very first assignment was to recalculate the incidence rate of HIV/AIDS. Four years earlier, CU had launched a project that was primarily managing an open HIV/AIDS cohort. That is a population of interest that recruited new members as the study progressed.

The project's focus was to manage a cohort of over 13,000 participants and produce periodic reports (in this case every six months) on the dynamics of the epidemic. The milestones were morbidity rates — incidence and prevalence.

The week when my assignment began coincided with a scientific conference in Holland where Dr. Maria Wawer (my boss) and other colleagues were presenting papers on the project findings. During that first week of the conference, Dr. Wawer contacted me to inquire about what incidence rates I had come up with. In the meantime, because of my limited knowledge of the data set, I recruited two experts who had been with the project as consultants during and since its interception. I identified what I believed were the most critical issues to be addressed before starting the computations and subsequent analysis.

The team was then assigned specific tasks. These included cleaning the relevant data set: generating frequency tables; identifying outliers; triangulating with both source data (original questionnaires), laboratory technicians (serology test results), and survey team members. After completing this cleaning and validation process (including correcting the numerous inconsistencies), we proceeded to perform the calculations using the statistical package — Statistical Package for Social Sciences (SPSS). This phase of the assignment went very well. After compiling the results, I then submitted the findings (as earlier agreed) to Dr. Wawer who was still at the conference in Holland. The recalculated rates this time were one infected case lower than what was being presented at the conference. And that, as it turned out, was a big deal! I received immediate feedback as anticipated, highlighting the fact that I was new to the project team with a limited understanding of the data sets.

During one of our weekly team meetings (post-conference), primarily to review what had gone wrong with our incidence rate, one of my colleagues was so embarrassed and distraught that he started shedding tears. Since no amount of consolation could calm him the meeting was immediately adjourned. In the meantime, members of a similar and "competing" project were constantly and consistently asking us what the real incidence rate was. What should they quote in their papers? As the message continued to spread, our team agreed on a consensus response, which was that the team was still in the review and validation process after which the final and latest incidence rates would be disclosed. This resolution served very well in mitigating further concerns.

During this process, our team went back to the drawing board to confirm what the real rates were. After our earlier computations and part of the triangulation process, we had actually conducted a recount of the new infections. The numbers were consistent with our findings. This recounting exercise was again conducted in addition to further calculations. And this time every degree of verification confirmed our results: there was one infected case too many!

And what is the message? PAAs and other quantitative methods are only as valid, reliable, and useful as the quality of data used.

## 3.1 Objectives

The objectives of this chapter are to examine:

- the current literature on PAAs with the focus on methods in which they are used to enable prediction of future events.

- case studies of the use of PAAs in industrial applications

- the conceptual framework on which PAAs are used to develop a machine language that enables prediction of future outcomes.

## 3.2 Theoretical frameworks

Descriptive highlights on which this framework's algorithm is based are as follows:

- A collection of literature materials explaining the concept of PAAs

- Relevant and applicable models used are reviewed;

- And simultaneously analyzing available literature material;

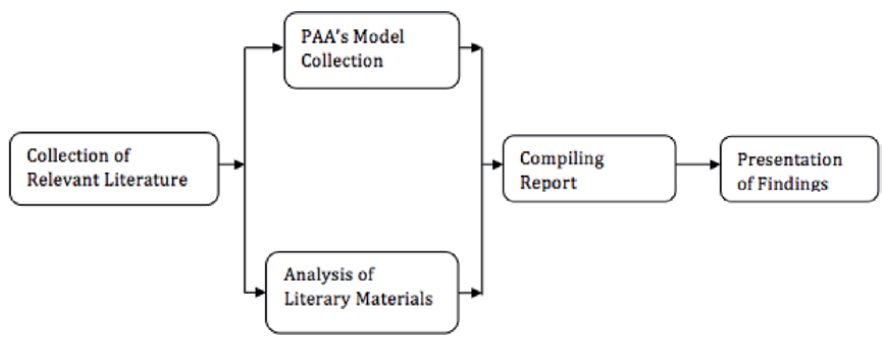

**Figure 1.**
*Theoretical framework [3].*

- An outcome report is compiled and;

- Findings are presented to relevant parties

- The required theoretical framework is as illustrated in **Figure 1**.

## 4. Scorecard

### 4.1 Description of the conceptual framework

A scorecard is a technique of measuring the performance of an organization in its entirety rather than focusing on a particular process or component of activities, tasks, and operations [4]. A balanced scorecard can be used to test the effectiveness of a program such as the ability of the program to be achieved at a reduced cost, increased efficiency, reduced efforts, and a high accuracy in producing the required outcomes. Previously, a balanced scorecard was designed to enable the assessment of the performance of companies and the extent to which its strategic decisions can be made to achieve the desired operational outcomes. It has been a relevant tool for companies in the assessment of the performance of internal processes and providing opportunities for learning and growth [5]. In spite of the perception that a balanced scorecard is used as a tool for measuring performance, it can be used in the measurement of other activities such as operational efficiency, effective time utilization, and the level of competitiveness of an organization in a particular industry.

### 4.2 How it works

A balanced scorecard (BSC) is used in deciding what a business is trying to achieve, to align resources in a manner that the regular activities of a business are achieved, and to create priorities for the provision of products and services to customers. It is composed of small boxes containing elements of mission, vision, core values of an organization, strategic areas of focus, and the activities in which a business will undertake to achieve continuous improvement [6].

BSC is primarily used by businesses, government agencies, and non-profit institutions. The working principle of a BSC is that an organization can be viewed from a number of perspectives, which can be used to create objectives, targets, and actions in relation to various points of views. The main perspectives of a BSC are listed below.

- Financial performance: The performance of an organization is viewed in terms of the effectiveness of its use of financial services.

- Customers/stakeholder needs: The BSC measures performance in terms of the ability to meet customer expectations.

- Internal procedures: The performance of an organization is viewed based on the quality and efficiency of production of a particular product, service, or major business processes.

- Capacity of an organization: From this perspective, an organizational performance is viewed based on its ability to utilize resources, technology, human capital, and other capabilities that create an environment for the achievement of a high performance.

### 4.3 When it is used to create PAAs

BSC can be used during the creation of PAAs by enabling the formulation of the performance features of the algorithms. The algorithms for analyzing an organization's performance can be analyzed using a BSC composed of capacity components such as the ability to be produced at low cost, ease of operation by the users, reduced likelihood of breakdown, and the ability to provide accurate forecast of an organization's performance (**Figure 2**) [7].

### 4.4 Strengths and weaknesses of the model

The strength of a balanced scorecard is that it provides the opportunity for putting all the operations of a business into consideration. It also accounts for the impacts of different components on each other rather than examining the manner in which a particular component operates or achieves its intended goals [8]. When a BSC has been integrated into the functions of an organization, it can be used as a tool for monitoring the achievement of goals and objectives.

The disadvantage of a BSC is that it focuses on the impacts in general, which neglects the performance of an individual or a particular process within a set of processes. There is the possibility of perverting a scorecard by using it as a tool for monitoring employees rather than the performance of a business [9]. It also takes into account a large number of variables to constitute a practicable scorecard, making it challenging to manage.

In Louisiana University College (LCU) of Engineering, a ClearPoint Strategic balanced scorecard software is used to align the activities such as enrollment, assessment of students, and improvement of the infrastructure of the department according to its vision, mission, and goals. The outcomes of the balanced scorecard enabled members of the institution to understand their performances in relation to the target outcomes that need to be achieved [10]. Due to this strategic plan, there has been increased enrollment in the college and it is considered to be the fifth fastest growing college of engineering in the U.S.

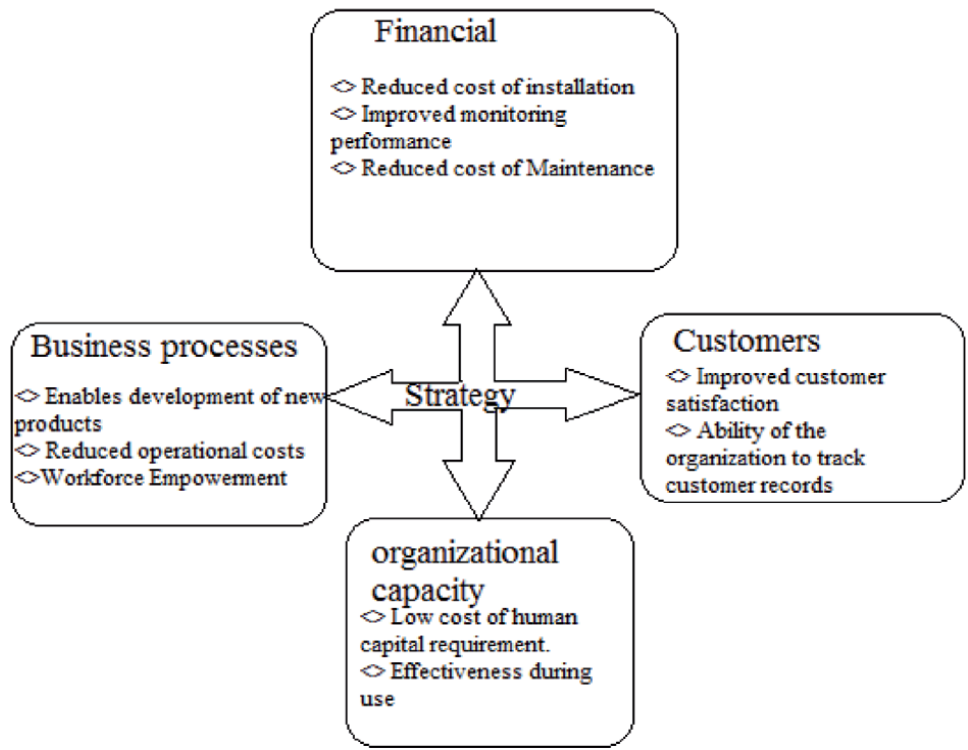

**Figure 2.**
*Balanced scorecard for the implementation of PAAs [7].*

## 5. Current models of designing PAAs

### 5.1 Forecasting and PAAs

Forecasting and analytics algorithms are used to create a model of a future event. An example of a common future event forecasted in many businesses is sales volumes. PAAs are used by sales managers to compare the outputs of the algorithms with achieved results, and to discuss the variations with their representatives who examine them and make estimates [11]. Forecasting algorithms also provide salespeople with the opportunities to know when they need to communicate prospects based on changes in algorithms, which have an impact on the buying decisions of customers.

### 5.2 Statistical models

Time series algorithm is a common statistical model of PAAs and is categorized into frequency-based algorithms and time-domain algorithms. Frequency-domain algorithms consist of spectral and wavelength analyses, while time-domain methods include algorithms used during auto-correlation and cross-correlation analyses [12]. Another commonly used statistical algorithm is the market segmentation algorithm that is extensively used in customer profiling depending on particular characteristics or priorities of a business.

### 5.3 Linear regression models

In simplistic terms, linear regression algorithms are used in modeling relationships between observed (dependent) and design (independent) variables. It is based on the least squares method that fits the best line and results into the minimal sum of squared errors between the expected and actual data points. Linear regression algorithms are used to make decisions such as the most suitable marketing mix to achieve optimized sales when particular investment channels are used. An example of a case where linear regression is used is at Cable Company X in the United States, where a program is used to determine the effect of variables that predict truck rolls within seven days. The variables used are downstream power, upstream power, and downstream signal-to-noise ratio [13]. The results that are statistically significant provide an insight on the interventions that need to be made to prevent truck roll.

### 5.4 Multiple regression models

Multiple regression analyses are used when product pricing is required across an industry such as real estate pricing and marketing organizations in order to establish the impact of a campaign. It is a broader category of regressions that incorporates both linear and nonlinear regressions and uses explanatory variables to perform an analysis [14]. The main application of multiple regression algorithms in practical situations is social science research, the analysis of the behavior of a device, or in the insurance industry to estimate the worthiness of a claim. Multiple regression analysis was used to examine the factors that affected the outcome of a referendum in which the United Kingdom opted to leave the European Union. The research involved the application of multivariate regression analysis in which the Logistic (Logit) Model was combined with real data to determine the statistically significant factors that have an impact on the voting preference in a simultaneous manner, in addition to the odds ratio that supports Leave or Remain [15]. The results of the multiple regressions showed that the gender of voters, age, and level of education were statistically significant factors, while country of birth was a statistically insignificant factor.

## 5.5 Multivariate regression model

In multivariate regression models, the value of a single variable is predicted using a number of independent variables. It is also used in the estimation of the relationship between predictors and responses. Predictors constitute continuous, categorical, or a combination of both. Multivariate analysis measures multivariate probability distributions in the context of their impacts on the observed data [10]. An example of such a model is multivariate analysis of covariance (MANOVA), which performs the analysis of variance that covers instances where more than one variable is analyzed simultaneously. Principal component analysis (PCA) is a multivariate analysis that enables the creation of a new set of orthogonal variables containing similar data as the original set. Multivariate regression analysis has been used by DHL, a global delivery company to predict future status of global trade, in its Global Trade Barometer program. A machine-learning language is used to input collected data related to different intermediate commodities that range from clothes, bumpers, or mobile devices [16]. The program leverages artificial intelligence and multivariate analysis PAAs to create a single data that enables understanding of the effects of a number of variables on a single variable. The output can be used by stakeholders to make decisions such as planning the capacity for future demands of their services and benchmarking on the forecasts to understand the industry's competitiveness.

## 5.6 Decision tree

Decision-tree algorithms are classified into supervised learning algorithms. They are used to create models for solving regression and classification problems. The goal of creating a decision tree is to generate values that can be used to predict the outcomes of a particular class or target variables by applying learning decision rules derived from past data [17]. The concept of tree representation of algorithms is used to solve a problem. Corresponding attributes are used in various internal nodes of the decision tree while class label is made at the leaf node. Pouch, a British plugin company developed an artificial intelligence (AI) chatbot, which informs customers of Black Friday discounts. The bot is available to users on Facebook Messenger and uses decision-tree logic to understand people's preferences [18]. The decision tree enables users to search the directory according to codes such as departments and their products, brands, and voucher codes of their preferences.

Milwaukee-based Aurora Health Care uses the technique of decision tree in the design of a "digital concierge," which operates on the principle of AI. The organization has cooperated with developers from Microsoft's arm of healthcare innovation in the development of a tool that simplifies decision-making in relation to patient care. The concept of decision tree is applied through a chatbot program, which can be accessed via a web browser [19]. This computer code enables mapping out symptoms and the common descriptions used by people to describe their health issues.

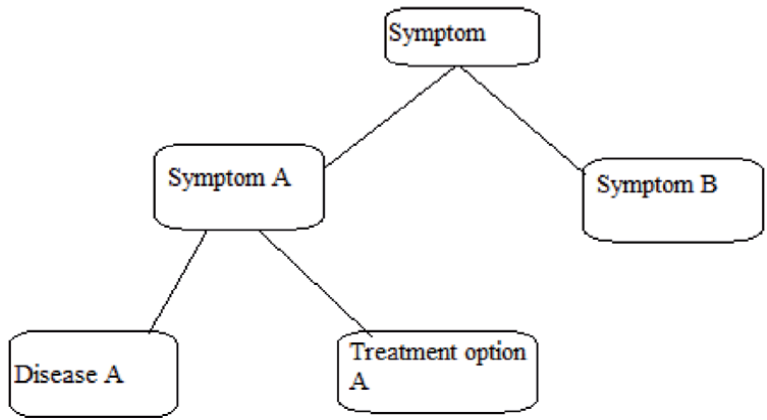

**Figure 3.**
*Framework of decision tree used by Aurora Health Care [19].*

The input is provided through answers to a set of questions regarding the symptoms presented. The bot adapts to the answers and outputs possible causes and treatment plan suggestions. The algorithm enables the creation of a command for making a decision on whether the patient may need further clinical care by the patient clicking a section that reserves his or her place in a line at an Aurora urgent care center. The conceptual framework of the chatbot is illustrated in **Figure 3**.

## 6. Data management

Testing data quality using predictive analytics algorithm takes place through the creation of a computer model for validity and reliability of data. The resulting computer model is usually a Pearson correlation that explains the relationship between response and design variables [20]. In measurement of reliability, the objective is to measure the extent to which the measured value is likely to change when the research is replicated. Some computer algorithms measure reliability by performing random and systematic error analyses. Eco-Absorber is a panel acoustics commercializing company that uses reliability and validity algorithms to get accurate outcomes of its surveys [21]. The outcomes are used to determine the suitability of the survey findings to recommend a change in practice that addresses the 4Ps of marketing in an effective manner.

## 7. Program management implications of PAAs

A number of considerations must be made when applying PAAs in program management. Good prediction can be achieved only if there are good data such as past records, which can be used to predict future outcomes of a process or an activity. For instance, prediction of sales of an organization in the next six months is subject to the availability of historical data that, when analyzed, provide a better understanding of the trend of changes in sales [22]. Before data analysis is conducted, they must be organized to reduce redundancy and unnecessary fields must be discarded. In order to deploy the insights from predictive analysis into the systems, it is recommended that software applications should be used to integrate them into predicting performances of businesses [23]. Some of the software that can be used includes API calls, predictive markup language, and web services. The reliability of PAAs algorithms is subject to the use of original data that have been prepared effectively through calculation of aggregate fields, identifying missing data, and merging a number of sources. Each component of data analysis should be analyzed independently. In case of advanced requirements, more advanced algorithms may be required [24].

## 8. Stages of PAA development

This section explains a more streamlined and contextual version of cross industry standard process for data mining (CRISP-DM). It is a neutral framework that addresses data analytics from two perspectives: application and technical. It is commonly used in predictive data analytics. As we focus on these details, it needs to be pointed out here that conducting (PDA) should never be diploid simply for the sake of expressing curiosity or flaunting one's knowledge of an existing problem-solving strategy. PDA is meant to solve problems. And in order to solve these problems, significant efforts are required to justify its application. One important component of such an exercise is the identification of a relevant management challenge. Hard questions need to be asked. What specifically is the issue? What are some of the interventions that have been made? How have the intervention outcomes improved or addressed the problem? And how have these interventions contributed in mitigating these problems. A combination of these questions will help significantly in redirecting and focusing intervention strategies.

### 8.1 Problem statement

In this stage, the business problem that needs to be addressed should be identified. The objective can be to perform a forecast of the future needs or to establish the likelihood of occurrence of a particular defect. The resulting predictive algorithm should be one that promotes the attainment of the goals and objectives that have been identified [13]. Problem statement identification also involves the definition of performance metrics that a business needs to achieve. A plan should be devised that enables the measurement of the metrics when data are input into the algorithm.

### 8.2 Intervention strategies

The intervention strategy involves making a decision about the right software or application to use in creating algorithms for resolving a particular operational procedure in a business. The intervention strategy may be to design an algorithm that enables understanding of the breakage of devices being manufactured, the likelihood of reduction in the number of purchases, or overall change in customer satisfaction.

### 8.3 Processes

The process of algorithm development will be determined by the goals to be achieved and the data to be analyzed. Algorithm development is achieved by the use of machine learning and data mining methods composed of relevant analytic platforms. The process of developing an algorithm can take different shapes according to the purpose to be achieved [25] of the commonly used methods in creating algorithms are the creation of computer programs that enable processing of data input to perform a number of tasks such as regression analyses or estimation of variances. The relationships between an organization's data sets can be amassed by the use of unsupervised clustering algorithms. The processes to be followed during the design of algorithms can be illustrated using flow charts [26]. These are charts composed of activities to be performed, decisions to be made, the arrows which show the direction of a program, and conditions that must be satisfied before a program progresses to the next stage.

## 8.4 Algorithm design

During algorithm design, the designer creates mathematical processes that can be used to solve problems. The concept used to develop algorithms is coding engineering. Algorithm design and implementation are achieved by the use of design patterns or template patterns and involve the use of data structures to create programs and subprograms that can be used to derive the mathematical output from a particular data input [27]. In order to develop an algorithm, mainframe programming languages that are recommended include ALGOL, FORTRAN, PL/I, and SNOBOL. The developer of an algorithm can create hand-written processes and a set of mechanical activities to be performed by hand before creating a corresponding algorithm using a computer program.

## 8.5 Program development

During the program development stage, a code is written in the form of pseudo-code and logic requirements to be followed in a particular programming language. Various coding language choices can be made in relation to a programming task depending on its characteristics and usability [18]. A relevant coding language is selected and syntax rules are followed with little deviation to improve the accuracy of the program.

## 8.6 Pilot testing

In this stage, the written program undergoes a debugging stage in which the programmer identifies errors in the program. The identified errors can be syntactic or logic. In addition, the programmer explores other areas that are likely to make the program not run in a proper manner or not run completely [21]. The pilot testing stage is usually lengthy and tedious and often constitutes more than 50% of the program development process. However, when there is greater attention to program design and coding, it is possible to reduce the amount of time spent in the debugging stage. Syntax errors result in difficulty of executing a program and constitute simple errors such as misspelling or failure to comply with the syntax rules to be followed in a particular programming language [12].

## 8.7 Pre-implementation testing

In this testing, test data is added to the program to determine its usability in providing the required outputs. Agile testing can also be performed by following the principle of testing from the customer's perspectives [23]. This testing should be performed by the quality assurance (QA) team. User acceptance testing (UAT) is performed on the program to determine whether it is usable in the intended system when released. This is due to the fact that changes in software characteristics undergo changes as it is developed. The resulting changes can be misunderstood in a fashion that is not according to the objectives of users. When UAT is completed, if all requirements are met, the program is moved to production and made available to the users.

## 8.8 Final implementation

The final implementation stage is where a program is used to conduct an analysis of a particular data to provide an output that can be used to predict future activities of an organization [14]. In the implementation stage, the data mined from an organization's database are input into the written computer program, processed (machine learning) and the resulting output is recorded and analyzed to enable prediction of a future characteristic of a program.

## 8.9 Lessons learned

The programmer conducts an assessment of a written program to establish whether the expected output has been achieved. A program that results in a desired output such as the number of customers who purchase products in a particular time period and considered useful should be retained by the organization.

## 8.10 Challenges

A major challenge that is likely to be encountered during any programming activity is that some programmers may not use algorithms that produce the expected output. Some problems are difficult to solve because they do not have parallel codes that can be used to write their corresponding programs. Some parallel algorithms have complex features that make execution of programs difficult. Debugging is an important skill but most people do not have the ability to identify and correct errors due to the frustrations and difficulties encountered during this process. The design phase of a computer program can be challenging in terms of the need to think about the program requirements that need to be put together in a manner that would facilitate future updates. When program design is not effective, the resulting program can be difficult to modify in the future.

## 9. Life case studies of the use of PAAs in institutions

In an attempt to simplify the conceptual complexities of PAAs, a select number of themes are included with life case studies. It is my hope that such an approach will enable readers to better internalize some of what has been accomplished and relate these accomplishments to their respective and unique themes.

## 9.1 Health

Data analytics are used extensively to predict the resource requirements for hospitals. At Texas Hospital, predictive analytics algorithms have been used to enable reduction of its 30-day rate of readmission due to heart failure [25]. The data used to conduct the analysis are the admission number of patients who are readmitted and those having heart failure in the past months. The most commonly used method is a computer program that can be written using Java, JavaScript, or C in which an algorithm is created to establish a regression equation that can be used to predict future readmission rates. The independent variable is the number of patients with heart failures while the dependent variable is the number of readmissions in the past months. The resulting output provides a value of regression equation that can be used in combination with the current number of heart failure patients to predict future readmission rates.

At the Harris Methodist Hospital outside Dallas, predictive analytics algorithms are used to conduct scans on medical records to establish the most suitable care that can result in an improvement in patient outcomes. The algorithm accounts for a number of data characteristics such as blood pressure and the amount of glucose in blood to act as an identifier of patients who are at risk of experiencing heart failure [28]. The algorithm creates a 30-day risk score representing the likely heart failure incidence. This enables physicians to focus on patients who need to be provided with intensive care. The commonly used programming languages are Python and PHP. The risk score is determined by creating an algorithm that measures the p-value using a computer program. A particular level of significance is used to determine whether there is a likelihood of heart failure. The input variables are the amount of glucose in blood and blood pressure. The output of the analytic program is the level of significance, which may be 0.05 or any set value by the hospital. Patients whose values fall within the significance value are at risk of heart failure and effectiveness of treatment measures should be improved in promoting their health [29]. An algorithm is created that measures multiple regressions in which two independent variables are used; amount of glucose in blood and blood pressure. The resulting regression equation in a computer program contains the sections for input of the independent variables. The program is run and a regression value provided is used to predict the possibility of heart failure in a patient.

## 9.2 Problem statement

It has been necessary to determine methods of identifying patients who are at risk of heart failure with less human involvement. The existence of machine languages such as Java, Javascript, and Python has provided the opportunity for practitioners at Harris Methodist Hospital in Dallas to develop a machine learning algorithm that enables distinction of patients at risk of heart failure in order to provide them with more intensive treatment.

## 9.3 Intervention strategy

The intervention includes the creation of a computer program based on machine learning languages in which the practitioners record patients' data and calculate the relationship between the values of blood glucose level and blood pressure to heart failure. This is where a notification is provided to the practitioners when blood pressure or blood glucose levels reaches a particular value.

## 9.4 Process

The process involved the installation of the machine learning languages into the systems at Harris Methodist Hospital, coding and testing of programs using sample patient values, training the employees to use the program, and its commission for use in identifying patients at risk of heart failure.

## 9.5 Algorithm design

The design of the algorithm was achieved by complying with the framework shown in **Figure 4**.

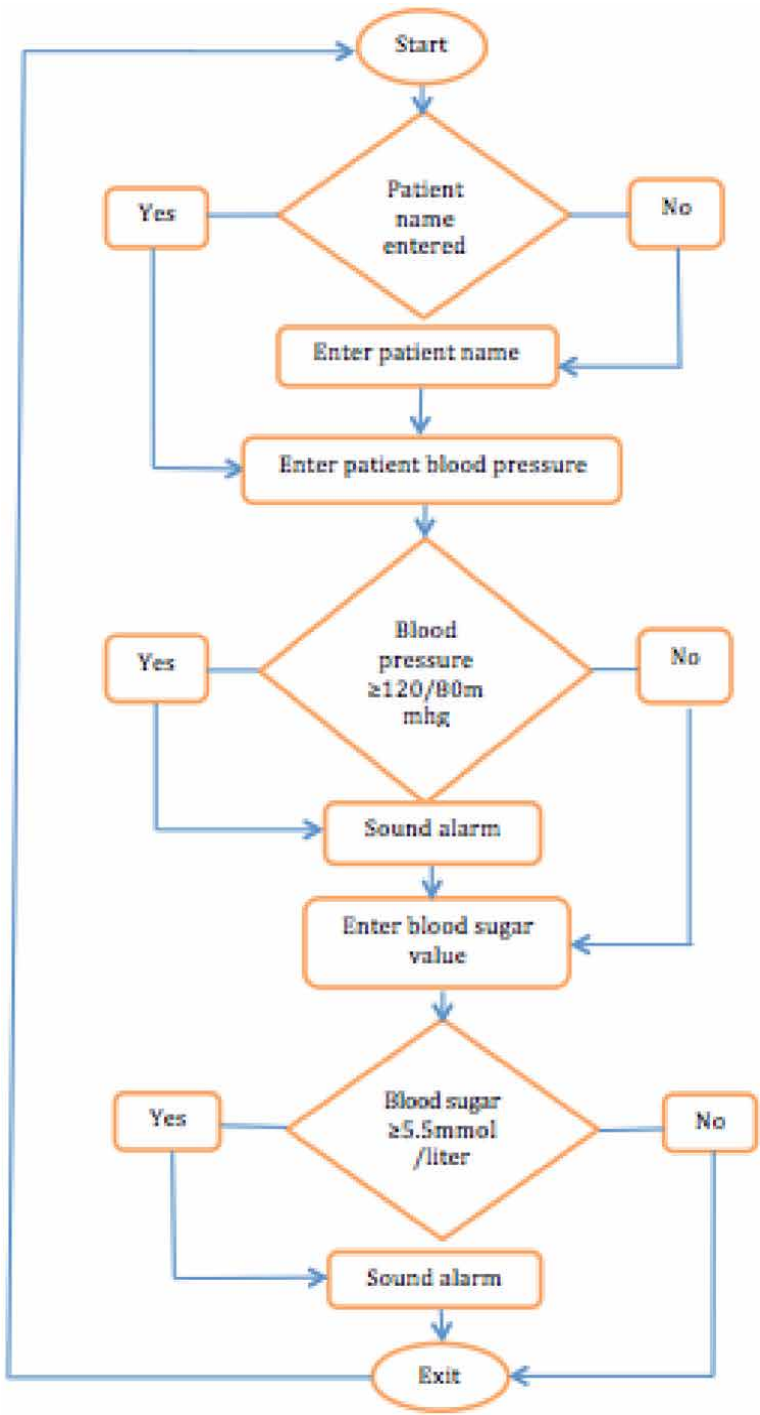

**Figure 4.**
*Algorithm framework for testing patients at risk of heart failure [21].*

### 9.6 Pre-implementation testing

Before the actual implementation of the algorithm, it is tested by adding the value of blood pressures and blood glucose levels of patients to determine whether it is able to sound an alarm when the values are higher than the maximum amounts.

The program is also debugged by removing syntax errors and misspelled words in order to improve its readability.

## 9.7 Final implementation

The final implementation is the integration of the machine learning language in the diagnosis of patients who are at risk of heart failure. The implementation involves authorizing the use of the software in the organization as well as training the personnel involved in patient care to examine patients who are at risk of heart failure.

## 9.8 Lessons learned

Machine learning algorithms can be created to enable healthcare professionals to make accurate decisions during the diagnosis of patients such as identifying those who are at risk of heart failure. The effectiveness of the program is determined by the nature of the machine language used, the competence of the personnel, and the dedication of the staff involved in monitoring blood sugar levels and blood pressure as determinants of heart failure.

## 9.9 Challenges

The major challenges that are likely to be encountered in the use of the program are the lack of staff motivation, difficulty in debugging due to failure to locate errors in coding, failure of organizations to allocate enough resources, and the practice of using machine learning language to diagnose patients for risks of heart failure.

## 9.10 Education

Many learning institutions have used predictive analytics to predict future performances by applying past performance scores of students in their institutions. At Southern Methodist University, an associate provost has contributed to student data management practices by applying predictive analytics algorithms that combine the grades attained by students in the past years to predict their performances in the future [11].

The analysis performed involves entering the raw data into the package and following the procedure of regression analysis. The preliminary result of the regression is a regression value that is related to the current performance of the student and is a factor that enables prediction of future performance. The final outcome is a standardized coefficient that acts as a predictor of the performance of a student in future tests based on the present performance.

## 9.11 Problem statement

The need to achieve an accurate prediction of the future performance of students at the Southern Methodist University (based on their present performances) is unquestionable. The use of a machine learning (ML) program is regarded as the most suitable approach for achieving this objective.

## 9.12 Intervention strategy

The intervention strategy that has been recommended is the use of an ML algorithm that calculates the regression value for the students' scores, which can be used to predict their performances in the next academic periods. The recommended statistical package is GNU PSPP, which has features that enable calculation of statistical measures such as simple linear regression, multiple linear regression, cluster analysis, and reliability analysis [30].

**9.13 Process**

The process involved was the installation of the GNU PSP application into the computer system followed by the design of the machine codes that return particular values of performance when information is input. The computer program will be composed of input points and the points of making decisions regarding the required outputs.

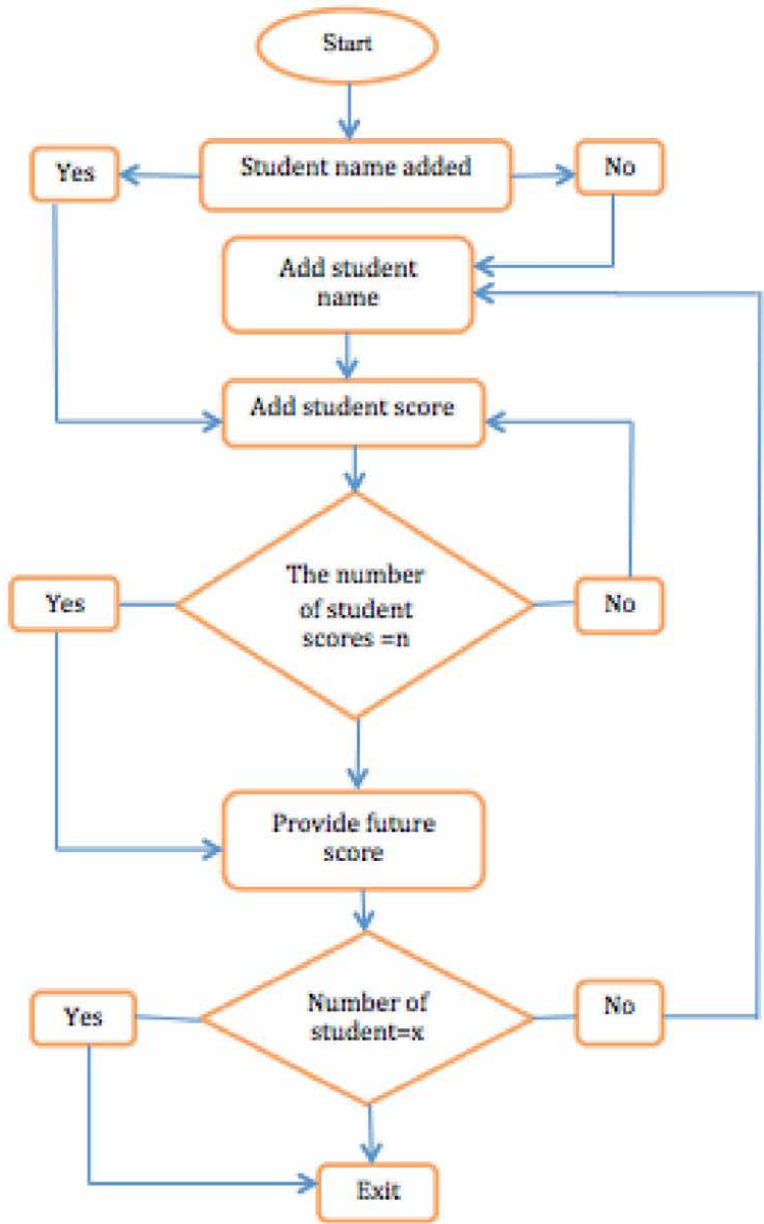

**Figure 5.**
*Design of the algorithm for prediction of future student's performances [10].*

### 9.14 Algorithm design

The design of the algorithm will take place immediately after the installation of the GNU PSP computer application. The design involves the use of computer decision frameworks such as the flowchart shown in **Figure 5**.

### 9.15 Pre-implementation testing

During the pre-implementation stage, the program is tested to determine whether there are any errors. Debugging is done to correct syntax errors and factors contributing to the failure of the program are examined. The ability of the program to be used in a particular system is tested.

### 9.16 Final implementation

The program is authorized for use in predicting the future academic performance of students in an institution, in which it is destined to be used [31]. The staff is trained to apply the program during the entry of students' previous performances. They also trained on the skills of interpreting the results of the program.

### 9.17 Lessons learned

The lessons learned from the program are that it is possible to design an effective program if the desired outcome is established. The programmer also needs to have the relevant knowledge including the steps for writing a machine code containing greater details. When a program for predicting future performances is created, it provides an approximate future performance of a student so that potential low performances can be mitigated.

### 9.18 Challenges

The challenges that are likely to be encountered during the design of the computer program are the omission of particular procedures that enable analysis of the inputs to provide the accurate prediction of future outcomes. A challenge is also likely to occur in the debugging stage when the source of the error cannot be located.

### 9.19 Agriculture

AgDNA intends to solve the issue of excess nitrogen by implementing the PAAs concept, in which nitrogen requirements are optimally matched with site-specific conditions in the field, thus reducing the likelihood of the occurrence of high amounts of nitrogen in the atmosphere. The company has integrated next-generation cloud computing technology and techniques for big data analysis, soil characteristics analysis, and climate data as information that enables understanding the nature of a farming field and its suitability for crop production [32]. These inputs are then combined using the most recent precision nitrogen management (PNM) frameworks to provide a prediction of the required amounts of nitrogen. The methodology used is the creation of a computer program in which the characteristics of the soil are compared to the amount of nitrogen in order to determine whether there is significance in the relationship. The statistical measure used in the analysis is the p-value, which measures the level of significance of the relationship between various soil characteristics and the amount of nitrogen. The software used in the computation of the relationship is JavaScript, which is cloud-computing software that enables the creation of programs for regression analyses. The analysis involves the input of the amount of nitrogen and the corresponding soil characteristics such as soil type, color, moisture content, and soil texture. The preliminary results are the p-values in which the values greater than the set significance value are soil types that have higher amounts of nitrogen that need to be regulated [33].

## 9.20 Business: marketing

Business application of PAAs occurs at the New York Times (NYT) as a means of improving its business and operational model. Predictive analytics models have been created that enable subscription to the organization's social media sites and other business decisions. According to a report by Chris Wilgins in a Predictive Analytics conference, predictive analytics is used to influence customers [10]. The NYT uses natural language processing as a means of increasing reader engagement so that the most beneficial types of articles can be sold. The software used is C program, in which an algorithm is developed that enables recognition of words such as adjectives used by customers to demonstrate their satisfaction. The software also has a subprogram, which enables the creation of a decision tree that matches the natural languages used by customers to make a particular decision. The preliminary result of the program is a tree diagram, which matches the natural language used by customers and the decisions that need to be taken to promote the sales of the NYT products.

## 9.21 Business: transportation

Virgin Atlantic uses predictive analytics algorithms to determine the prices of tickets according to the likelihood of travel demands by customers [6]. The statistical packages used are either MATLAB or SPSS, which have features that enable the calculation of statistical measures such as regression analysis, multiple regression analyses, correlation analyses, and the T-test. The methodology used is the input of the raw data such as prices of tickets and the corresponding number of customers who board flights in a specified period such as a month or a year. The statistical measures conducted include regression analysis and significance analyses. The preliminary regression value is used as a measure of the relationship between independent variables (price) and the dependent variable (number of customers). A final prediction of future demand in ticket sales is established by the use of the regression coefficient to predict the likely number of customers.

## 9.22 Sports

A commonly used predictive analytic model in sports is Sports Performance Platform (SPP) from Microsoft, which incorporates an ML and AI in the creation of algorithms used to make decisions regarding the performance of athletes. This application provides solutions for the locker room, performance lab, and has an algorithm that enables prevention of injuries, making decisions pertaining to games, and changing training schemes to improve the performances of athletes [15]. An example of a sports club that uses PAAs is Midtjylland, a Danish club that was on the brink of bankruptcy but improved to nearly winning a championship title. The club made changes to the steering approach by implementing analytical procedures in which experts conducted an analysis of each player twice a month to obtain information that addressed the player's training needs. The experts also provided the coach with information such as when to change the game plan in accordance with the in-game statistics. Information from analytical models was used to recommend new players [34]. The programming software used for the analysis of the players was SPP. The methodology used was the creation of an algorithm that enabled input of player behaviors such as the number of passes, distances covered, number of touches of the ball, and the resulting team performance such as the number of wins, draws, and losses. The algorithm creation methodology also involved the creation of a code that measured the regression between the variables. The preliminary results were the raw player data in the computer program and the team's performance in the past matches. The final outcome was the regression value, which showed the relationship between each player's characteristics and the team's performance. This value is important in making decisions such as whether to substitute a player in order to improve the performance of the club.

### 9.23 Social media

Social networking companies such as Facebook have been using predictive analytics algorithms that enable updates regarding a brand to be available to the user after a product has been "liked". Therefore, users are able to see posts, which improve their engagement rates with their individual networks such as posts that their friends have engaged with [16]. The programming language used is JavaScript due to its cloud computing feature and the ability to make changes to an already existing algorithm. The methodology used is the creation of an algorithm that enables the site to link a liked product to a user's page. The process includes the statistical analysis of a decision tree in which the website is automatically coded to link a "liked" product to the user's page. The final outcome is a user experience in which when a person likes a product, the updates regarding the product appear on their page in the future. This implies that Facebook will promote the ability of marketers to promote social engagement with customers.

### 9.24 Manufacturing

In manufacturing companies, machine-learning algorithms have been used to understand the machine problems that are likely to be encountered in order to apply preventive practices to keep the supply chain operational. At Georgia Institute of Technology, machine-learning algorithms provide the opportunity to promote forecasting the likelihood of machine failures, thus, enabling the technicians to perform maintenance practices [3]. The machine learning language used is a C program with capabilities for creating codes that enable calculation of statistical tests such as regression analyses, linear regression, and multiple regressions. The methodology used is the creation of a computer algorithm in which past intervals of failures is added. The data are the failure times (the dependent variable) and the time interval (independent variable). A sub-program is created that enables the calculation of simple regression analysis, which establishes the relationship between machine failure times and the time interval. The preliminary results are the input values of failures of the machines against time interval. The outcome of the analysis is a regression coefficient, which can be multiplied by the current failure frequency to determine the next likelihood of the machine's failure. This ML algorithm has been applied in the performance of regular maintenance tasks on lathes, grinders, saws, and gears (**Figure 6**).

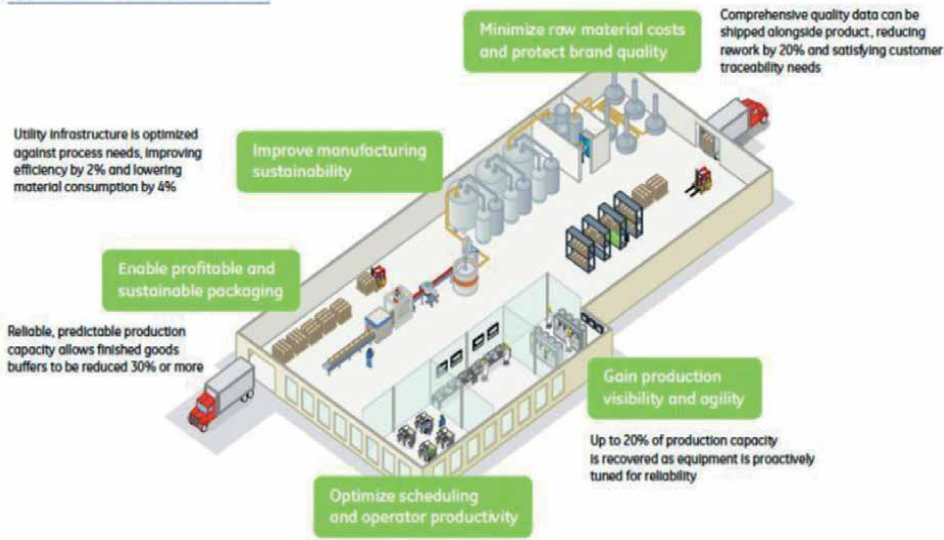

**Figure 6.**
*Summary of the improvements made at General Electric [3].*

### 9.25 Government institutions

In the United Kingdom (UK), the Ministry of Defense uses machine learning algorithms to explore and organize public documents. This is achieved by creating algorithms that enable the identification of documents depending on their subjects and conducts the analysis of information for the purpose of finding patterns and anomalies in data systems [25]. The algorithms are also implemented in the detection of fraudulent activities, transactions, or activities of any public official for personal gain. The algorithms have been effective in the detection of activities such as money laundering, the creation of counterfeit trade items or the duplication of business documents. The processes include the installation of machine learning languages into the systems of the organizations, the creation of computer programs, testing, and implementation [35]. The inputs are information regarding future activities such as the attempt to change the content of documents in order to achieve personal objectives or defraud the government. The program is capable of providing information about the perpetrators of the acts and insights on characteristics that can be used to trace them.

### 9.26 Crowdsourcing

Bugcrowd Inc. uses crowdsourcing, in cooperation with Fortune 500 companies such as MasterCard Incorporation, to identify vulnerabilities that may be used by hackers to infringe on their infrastructure. This is achieved by the use of a machine learning language called a bug bounty program, which enables the engagement of the cybersecurity community, providing them with monetary rewards for their contribution to the resolution of the vulnerabilities [2]. A major advantage associated with the company is the lack of a requirement for evaluation of claims of cyber threats using the crowd-sourced information to determine the areas of security where greater attention should be placed. Crowdsourcing also involves the use of application programming interfaces (APIs), a tool for software development that integrates the sourced data into the current workflows or reports of business security analyses. The process involves the selection of a suitable programming language such as Python and installing it in the organization's system [5]. Professionals in machine code development develop machine codes that enable the recording of information from a number of sources. The output is a list of sources of information containing cybersecurity information that is usable for improving the organization's databases.

## 10. International development programs that use PAAs

From a geopolitical perspective, I have also included case studies on themes that are universally applicable with specific emphasis on select themes that significantly contribute in making the world a better place and hence promoting a better quality of life.

### 10.1 Natural disaster programs

The concept of predictive analytic algorithms has been implemented in the analysis of big data regarding past natural disasters and used to predict future incidences [23]. An example of an incident that provided data for fighting natural disasters is the earthquake that occurred in Haiti in 2010. Crowdsourcing has been used to obtain real-time images of disasters such as earthquakes while big data approaches in artificial intelligence (AI) have been used to determine meanings in messages such as SMS that were generated during the occurrence of natural disasters.

The processes involved the installation of machine learning language followed by the creation of an algorithm that enables the performance of mathematical analyses such as regression analysis and providing the output that can be interpreted to estimate the likelihood of occurrence of a similar incident such as another earthquake in the future [33]. The analytical procedures performed involve the input of information pertaining to disasters such as the magnitude of an earthquake, time of occurrence, and region into the machine language. The machine language performs an analysis of mathematical processes such as linear regression and multiple regressions to provide statistical coefficients that can be used to predict future disasters.

### 10.2 Poverty eradication program

Predictive analytics have been used by the World Bank (WB) in poverty eradication initiatives such as the collection of information of affected areas, the analysis of the number of people who need relief services, and the relationship between their status with infectious diseases. This is in accordance with the WB objective of eradicating poverty by the year 2050. Countries conduct household surveys and provide WB with information used to classify the population according to the level of poverty [25].

The processes involve the creation of a machine language that enables input of raw data such as the economic statuses of families. Data from statistical offices in various countries are input into the machine learning language that has been designed in a customized fashion to enable the stratification of families according to their gender, age, income levels, geographical location, race, or culture. The program has commands that enable the quick calculation of statistical measures such as linear regression or multiple regressions to provide coefficients that enable the prediction of poverty levels in the future [2]. The machine learning language has also been designed in a manner that enables the transfer of data from mobile phones to the program for analysis. This program has been implemented to measure the economic status of people in Togo, Tanzania, and Tajikistan to provide outputs that enable prediction of poverty status in the future. A similar program has been used by the WB in the measurement of the movements of nomadic people in Somalia to predict future migration patterns.

## 11. Programming software

### 11.1 Turn-key programming model

A turn-key program (TKP) is one that is developed according to specifications because the owner has specified all the functional requirements. A TKP has the primary advantage of enabling the user to establish a program budget, inputs, and outputs in a scheduled manner. Turn-key programs do not provide easy flexibility in the management of changes and other features requested by the programmer.

### 11.2 In house programming model

In in-house programming, a program is developed by the IT department of the company rather than an outside company [32]. An example of in-house programming is Google's software development, which is done using its machines that are located in various parts of the computer network system.

### 11.3 Outsourcing programming model

Outsourcing programming is the process in which a computer program is written by a third party and generally external institutions on a consulting basis. It is a more advantageous method of programming because an organization reduces the cost of undertaking a particular project. It is also a means of ensuring time-saving in the development of computer programs because it tends to be less time-consuming when a number of experts are assigned to complete program development. The risks and challenges involved in outsourcing are confidentiality, limited supervision, possible tardiness and service-provider loyalty.

## 12. Programming languages, architecture development, platform, interfaces

### 12.1 Java

Java is a major programming language used in building server-side programs for video games and apps in mobile phones. It is also popular in the creation of programs for operation on Android-based platforms. Java incorporates both compilation and interpretation techniques [3]. Java compiler is used to convert a source code into bytes. Java Virtual Machine (JVM) performs an interpretation of the bytecode and the creation of a code that can be executed when the program is run. Java is highly recommended during the creation of web server programs, web commerce applications such as electronic trading systems, scientific applications, and enterprise databases (**Figure 7**).

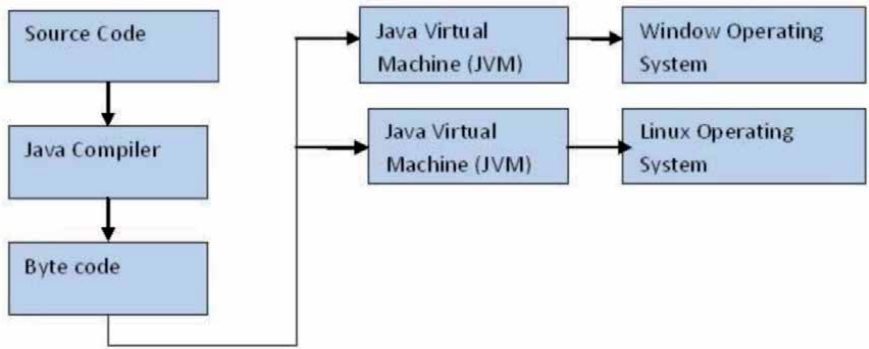

**Figure 7.**
*A mapping of Java programming language architecture [3].*

## 12.2 Python

Python is an object-oriented programming language that is popular due to its simple and readable syntax. It is easy to learn and uses simple language for program coding. For instance, if the computer is required to write something, the command "print" is used. Python makes use of the concept of dynamic typing, reference counting, and detection of garbage in order to facilitate memory management [11]. It uses similar expressions to other programming languages such as C and Java (**Figure 8**).

## 12.3 C language

C is a compiler program that can be used to translate functions, declarations, and definitions into files that are executable. It has a simpler command procedure and performs less programming tasks compared with other languages used in programming such as Python or Java. Executable files are created by the compiler translating source code into executable codes independently. It does not remember the defined variables while performing file processing [2]. This implies that a variable cannot be used if it has undergone previous declaration in the same file. C is similar to Java in functions such as loops and conditionals, but the former is simpler in other aspects, such as the structure of data definitions (**Figure 9**).

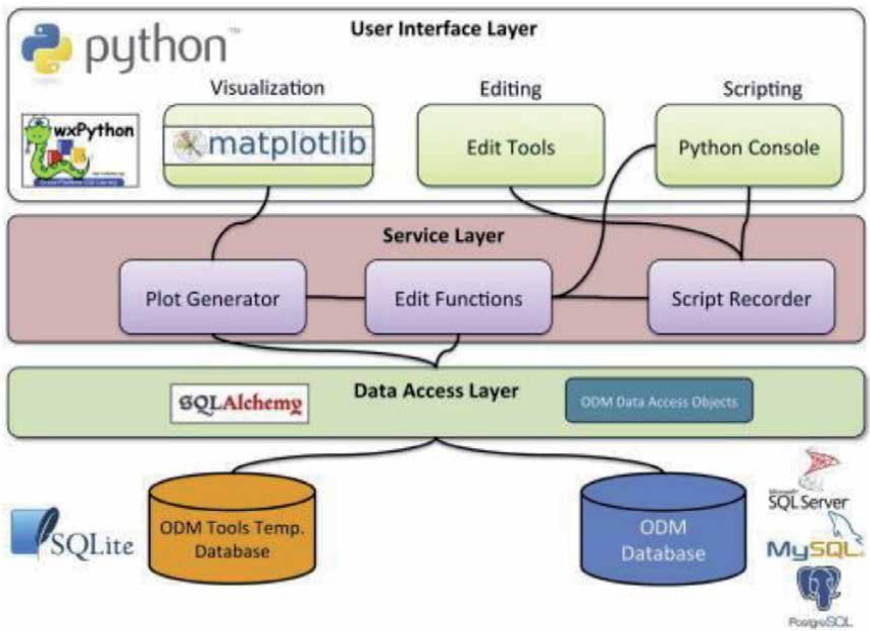

**Figure 8.**
*Architecture of Python programming language [11].*

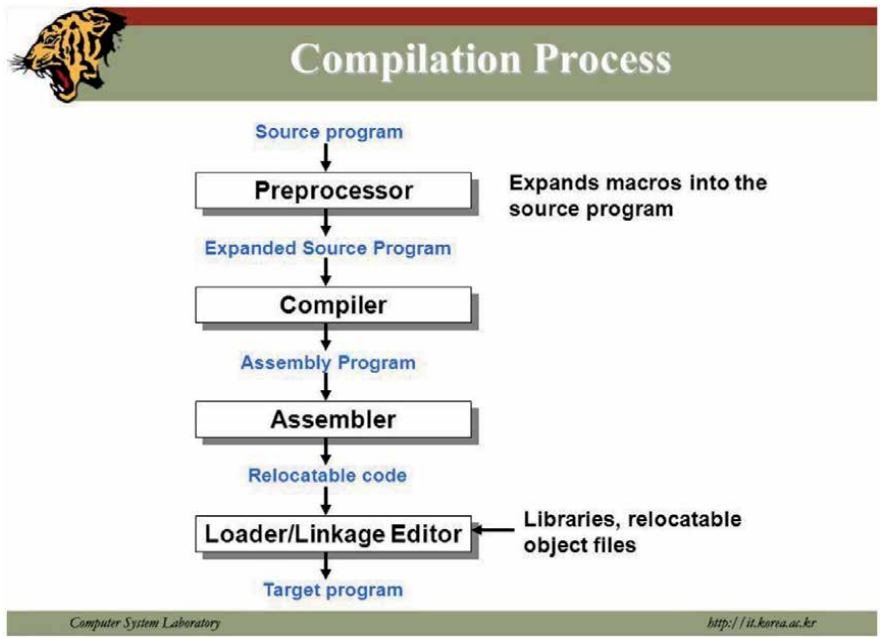

**Figure 9.**
*Compiler architecture of a C program [2].*

## 13. Algorithm development: examples of algorithms

### 13.1 Brute force algorithms

Brute force algorithms enable enumeration of all integers from 1 to n and establish whether each number is divisible by n to obtain a whole number. With this type of algorithm, direct computation is performed based on a problem statement to be resolved and the corresponding concepts [7]. The search phase for the text can be done randomly. It is an algorithm that is commonly used in the solution of problems such as sorting, searching, and binomial expansion.

### 13.2 Simple recursive algorithm

A recursive (self-executing) algorithm is one that uses smaller input values and applies simple operations to them in order to obtain the result. It applies the principle of solving a problem by dividing it into smaller versions, which can then be solved by the use of recursive algorithms. If a function is represented recursively, the corresponding recursive algorithm for the computation of its members is a mirror of the definition.

### 13.3 Backtracking algorithms

A backtracking algorithm is an algorithm that is used to find solutions to computational problems such as conditional problems. The process of programming starts with a particular move out of a number of alternatives [36]. If it is possible to reach a solution using the selected move, the solution is printed; otherwise, the program backtracks and selects another move to try.

### 13.4 Randomized algorithms

Randomized algorithms use the concept of randomness to determine the task to be performed anywhere in the algorithm. Their preferred use is for the analysis of expectation of worst cases, in which all likely values of the random variables are considered and the corresponding time by a possible value is evaluated.

### 13.5 Dynamic programming algorithms

Dynamic programming is the process where algorithms are created for breaking down a problem into a number of sub-programs. These problems are solved just once and the result is stored so that when a similar problem occurs in the future, a solution is looked up amongst the stored solutions [7]. This basically involves creating a program that memorizes the results of a particular state and using it to solve a sub-problem.

## 14. Highlights

This chapter has reviewed and analyzed contemporary documentation pertaining to the use of PAAs, the processes involved in their development, their application in the computation of mathematical procedures, such as linear regression and multiple regression, and prediction of future outcomes. The stages in which PAAs undergo until the outcome is achieved include problem statement, intervention strategy formulation, processes, algorithm design, program development, pilot testing, pre-implementation testing, the analysis of lessons learned, and examination of the challenges encountered.

The concept of PAAs has been used in most machine-learning languages to develop computer programs that provide an output, which enables understanding future events in healthcare, education, manufacturing, governance, and natural calamities such as earthquakes or poverty levels. In healthcare practice, it has been possible to develop a PAA that uses blood sugar levels and blood pressure to predict the patients who are at risk of heart failure so that intervention measures can be implemented. In educational institutions, PAAs have been developed that enable the input of the student's performance in the present period to predict future performances in various fields of specialization. In agriculture, big data PAAs have been used to formulate soil characteristics in the future based on the current characteristics such as soil moisture content, the amount of nitrogen in the soil, and the amount of salts. The output has been used, for example, as a guide on the measures that can be taken to reduce the amount of nitrogen in the soil. Other areas where PAAs have been used are player performance prediction in sports, sales predictions in businesses, predictions of unauthorized acts in government departments, and crowdsourcing to promote organizational cybersecurity.

## 15. Summary

The euphoria created by the advent and exponential evolution of predictive analytics seems to have left many stakeholders in awe. From every level of business to different institutional categories, the best and optimal performance seems to be in sight with no establishment being left behind.

While the positive outcomes achieved so far continue to escalate, institutions at large need to take one step backwards to do some stocktaking. This process involves asking critical and provocative questions, including: Are we doing the right thing? How evidence-based are our strategies? Are they sustainable? How reliable are our data sets? Is client data adequately protected from potential cybercriminals? Have all the ethical concerns been adequately addressed? What is the gold standard?

If PAAs' dynamics are any indication, the learning curve is bound to be long, steep, and daunting. One major reason for this possibility is the growing complexities of managing data and the institutions involved in processing them. There is also the challenge of establishing a diverse team of experts involved in developing problem solutions. Members of such a complementary group serve as an invaluable backbone to any potential success. The problems are complex, ranging from good quality data to the nuances that accompany risks and assumptions of selecting and applying the appropriate algorithms.

As already indicated elsewhere in this chapter, good quality data is *sine qua non* to any successful analysis (quantitative and qualitative). Mark Twain's characterization of lies, "lies, damned lies and statistics," should always serve as a compelling reminder that the information generated from data through the machine learning (ML) process is only as useful as the quality of data used. Having and using the appropriate and reliable piece of information is a catalyst for making informed decisions. PAAs are no exception! ML processes continue to gauge significant amounts of data. This data is transformed through the ML process to predictive outcomes (information) used in making informed decisions. ML's propensities to process big data sets have made cloud computing an inevitable requirement. The arrival of quantum computers (QC) has made the transformation process faster, reliable, and more efficient. These QCs, which have miniaturized the binary digit (bit), have moved computing to a higher level. According to an IBM definition, "Quantum computers, on the other hand, are based on qubits, which operate according to two key principles of quantum physics: superposition and entanglement. Superposition means that each qubit can represent both a 1 and a 0 at the same time." Access to good quality data is one way of optimizing the utilization of these QCs.

In one of my series of lectures given to graduate students at the University of the West Indies in Kingston, Jamaica, a student wanted to know why program managers firmly believe that in any strategic framework — "logframe" for example — outputs (and their indicators) always contribute to outcomes, especially given the potential for misleading and unreliable results reported at the output level.

In my response, I agreed with the student while elaborating on the data collection and reporting vulnerabilities, especially in environments where very little appreciation is given to data that are subsequently converted to information. I explained the trade-offs that managers and other stakeholders are faced with. I described what it takes to address issues like these, including conducting a control study. I further shared an anecdote with the group; an experience I had conducting a program evaluation for a UN agency. In this case, the agency had spent 4.5 million dollars over a three-year period on national capacity strengthening. The participants, who were medical health workers, were trained both nationally and internationally. This was identified as one of the output indicators that contributed to a corresponding relevant indicator — improved quality of health services — at the outcome result level. During the evaluation assignment, I followed up (something that was never done after training), and as it turned out, most of those who benefitted from the training had moved on; some changed ministries, others had left the country, and some had even changed professions! Obviously, any planning decisions made using that training report would undoubtedly be erroneous, misleading, and deceptive at best.

It is quite conceivable that the evolving, inclusive, and streamlining dynamic of PAAs will continue to have positive and unquestionable consequences on how programs are managed. The myriad implications are unfathomable with synergies that collectively yield both intended and unintended outcomes. If current thematic applications are any indications, introducing analytics in any intervention will continue to be a win-win initiative.

While different institutions condition their interventions towards their respective strategies, the ultimate outcome is improved productivity and optimization of resource (human, financial, and material) utilization. There is also the human (quality of life) dimension that can revolutionize, reverse, and mitigate certain nuances that affect our wellbeing. For example, academic institutions now apply some models for improving student performance. By using historical data these institutions are able to identify vulnerable students, counsel them on an individual basis, and enable them to set more achievable objectives based on their academic performance with respect to the average group standing. The ultimate outcomes demonstrate counterfactuals that are obvious. And the results have been quite impressive. Some students in some cases have even encouraged themselves to become their own agents of change.

There is also gradually and increasingly, an inclusive element of analytics that continues to encourage and involve members of different community populations: crowdsourcing. This strategy has mushroomed and generated an astounding dynamic amongst communities. It remains to be seen to what extent the strategy will contribute to improving people's quality of life.

In general, business institutions are ahead of the curve with marketing as one of the trailblazers. The competition remains extensive and brutal.

## Abbreviations

| | |
|---|---|
| ML | Machine learning |
| PAA | Predictive analytics algorithms |
| SP | Service provider |
| QOD | Quality of data |
| GIGO | garbage in garbage out |
| CU | Columbia University |
| SPSS | Statistical Package for Social Sciences |
| BSC | Balanced scorecard |
| LCU | Louisiana University College |
| PCA | Principal component analysis |
| AI | Artificial intelligence |
| QA | Quality assurance |
| UAT | User acceptance testing |
| PNM | Precision nitrogen management |
| NYT | New York Times |
| SPP | Sports Performance Platform |
| UK | United Kingdom |
| API | Application programming interface |

| WB | World Bank |
|---|---|
| TKP | Turn-key program |
| JVM | Java virtual machine |
| QC | Quantum computer |
| MANOVA | Multivariate analysis of covariance |
| CRISP-DM | Cross industry standard process for data mining |

## Author details

Bongs Lainjo
Cybermatic International, Montréal, QC, Canada

*Address all correspondence to: bsuiru@icloud.com

## IntechOpen

*Edited by Jaydip Sen*

Recent times are witnessing rapid development in machine learning algorithm systems, especially in reinforcement learning, natural language processing, computer and robot vision, image processing, speech, and emotional processing and understanding. In tune with the increasing importance and relevance of machine learning models, algorithms, and their applications, and with the emergence of more innovative uses–cases of deep learning and artificial intelligence, the current volume presents a few innovative research works and their applications in real-world, such as stock trading, medical and healthcare systems, and software automation. The chapters in the book illustrate how machine learning and deep learning algorithms and models are designed, optimized, and deployed. The volume will be useful for advanced graduate and doctoral students, researchers, faculty members of universities, practicing data scientists and data engineers, professionals, and consultants working on the broad areas of machine learning, deep learning, and artificial intelligence.

*Andries Engelbrecht, Artificial Intelligence Series Editor*

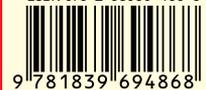